\documentclass{article}

\usepackage[final]{corl_2020} 

\title{Hierarchical Robot Navigation \\ in Novel Environments using Rough 2-D Maps}

\author{
  Chengguang ~Xu \hspace{1cm} Christopher ~Amato \hspace{1cm}    Lawson L.S. ~Wong\\
  Khoury College of Computer Sciences\\
  Northeastern University \\
  United States\\
  \texttt{\{xu.cheng, c.amato\}@northeastern.edu, lsw@ccs.neu.edu} \\
}

\usepackage{graphicx}
\usepackage{algpseudocode}
\usepackage{algorithm}
\usepackage{subcaption}
\usepackage{booktabs}
\usepackage{wrapfig}

\usepackage{amsmath}
\usepackage{amssymb}
\usepackage{natbib}
\usepackage{enumitem}

\begin{document}
\maketitle


\begin{abstract}
        In robot navigation, generalizing quickly to unseen environments is essential.  Hierarchical methods inspired by human navigation have been proposed, typically consisting of a high-level landmark proposer and a low-level controller. However, these methods either require precise high-level information to be given in advance, or need to construct such guidance from extensive interaction with the environment. In this work, we propose an approach that leverages a rough 2-D map of the environment to navigate in novel environments without requiring further learning. In particular, we introduce a dynamic topological map that can be initialized from the rough 2-D map along with a high-level planning approach for proposing reachable 2-D map patches of the intermediate landmarks between the start and goal locations. To use proposed 2-D patches, we train a deep generative model to generate intermediate landmarks in observation space which are used as subgoals by low-level goal-conditioned reinforcement learning. Importantly, because the low-level controller is only trained with local behaviors (e.g. go across the intersection, turn left at a corner) on existing environments, this framework allows us to generalize to novel environments given only a rough 2-D map, without requiring further learning. Experimental results demonstrate the effectiveness of the proposed framework in both seen and novel environments.
\end{abstract}


\section{Introduction}
	
As robots become ubiquitous in our society, they will frequently encounter novel scenarios, making fast and efficient generalization critical.
For robot navigation, although SLAM-based approaches
already achieve impressive performance (e.g., \citep{taketomi2017visualSLAM,aulinas2008SLAM,mur2015orb-slam}), such methods need to build accurate occupancy maps of environments before navigation. This is computationally expensive and time intensive; a robot going somewhere new to run an errand should not need to first build a map of every new place it encounters along the way. We need navigation approaches that can not only navigate the robot in known environments, but also generalize efficiently to novel environments.

\begin{figure}[t]
    \centering
    \includegraphics[width=12cm, height=4cm]{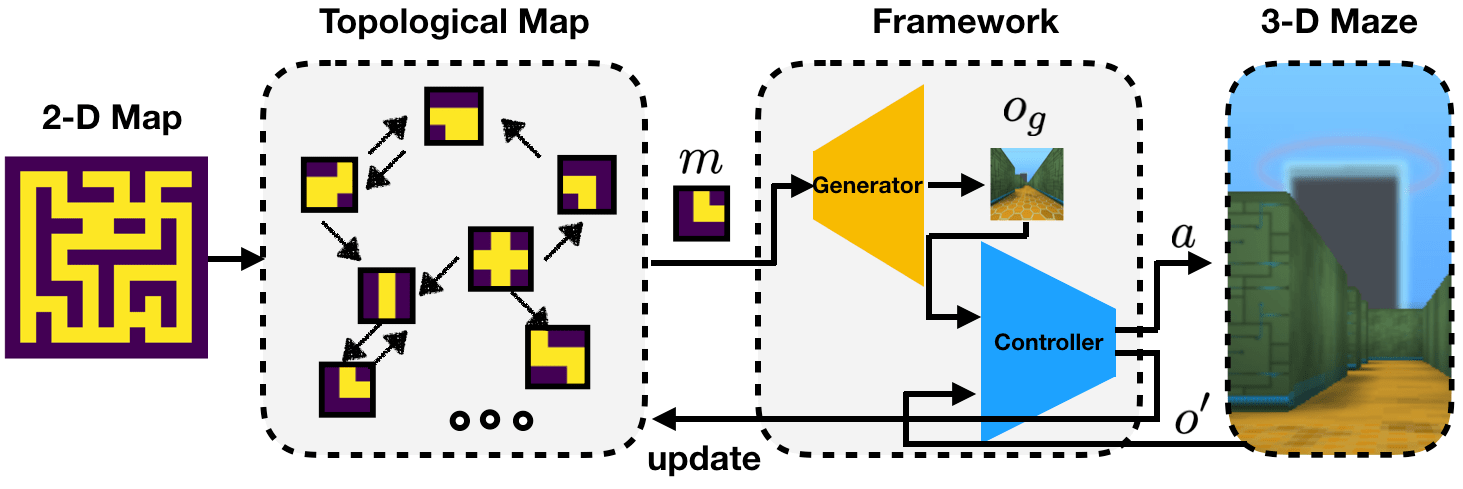}
    \caption{General framework. Given the rough 2-D map of an unseen maze, a dynamic topological map is first initialized from it. During navigation, the dynamic topological map proposes a local map patch $m$ of the next landmark to reach. Next, the generative model uses $m$ to generate a first-person observation $o_g$ corresponding to the landmark. The local controller then uses $o_g$ as a subgoal and executes the local navigation. If the move is successful, the local map patch of next landmark is generated; otherwise, the dynamic map is updated, and another path is planned and executed.}
    \label{fig:framework}
\end{figure}

Humans are very capable of navigating in novel environments. 
Studies have shown that humans do not strongly depend on metric information to navigate \citep{foo2005huamnPsychology,wang2002humanSpatioRep}.
Instead, given a rough 2-D map (e.g., the 2-D map shown in Figure~\ref{fig:framework}) of a novel environment, humans typically plan several landmarks between the start and goal locations and then reach them sequentially. Following this insight, previous hierarchical work has generally fallen into two categories, either assuming precise high-level information is given (\citet{brunner2018MapReader}), or learning the high-level structure directly from the environment (\citet{eysenbach2019SoRB}, \citet{huang2019mapplanner}). With provided precise high-level information, methods in the first category achieve good generalization performance. However, obtaining such precise high-level information is usually expensive. With the map purely constructed from experience, methods in the second category do not require precise high-level information. However, the learned topological map only reflects the experienced environment and makes it hard to generalize to novel environments.

Our proposed framework is a blend of these two categories, retaining the efficient generalization performance of the first category, while extending it to handle rough or even imprecise high-level information. As illustrated in Figure~\ref{fig:framework}, we propose a framework with following contributions:
\begin{itemize}[leftmargin=*]
    \item We introduce a \textbf{dynamic topological map} that can be initialized from a \textbf{rough 2-D map}. The topological map is updated during navigation to reflect both inaccuracies in the provided high-level rough 2-D map and imperfect local behaviors of the low-level controller.
    \item We propose a hierarchical navigation framework that uses a deep generative model to generate \textbf{unseen observations} corresponding to future landmarks. These observations are used as subgoals by a \textbf{reusable low-level controller}.
    In our case, we use a conditional VAE~\citep{sohn2015CVAE-1} for observation generation and a goal-conditioned double DQN~\citep{mnih2015DQN,schaul2015UvFA} for local navigation.
\end{itemize}
We test our framework in DeepMind Lab~\citep{beattie2016deepmind}, a complex 3-D maze environment. High-level information is provided by rough 2-D maps that only capture the global layout (e.g. the 2-D map in Figure~\ref{fig:framework}), but not the first-person visual appearance, and may even be incorrect.
Compared with baseline methods and the state-of-the-art Map Planner~\citep{huang2019mapplanner}, experimental results demonstrate that our framework can achieve better navigation performance in both seen and unseen mazes. Furthermore, our approach is robust to small amounts of error in the rough 2-D map, and ablation experiments demonstrate the utility of both the rough 2-D map and the constructed dynamic topological map.

\section{Background}
\label{sec:back ground}
We formulate the problem of visual navigation as a partially observable Markov decision process (POMDP)~\citep{Kaelbling1998}, represented by a tuple $\langle\mathcal{S},\mathcal{A},\mathcal{T},\mathcal{R},\Omega,\mathcal{O}\rangle$, where $\mathcal{S}$, $\mathcal{A}$, and $\Omega$ are finite sets of states, actions, and observations respectively, $\mathcal{T}$ and $\mathcal{O}$ specify transition and observation probabilities, and $\mathcal{R}$ is the reward function. In visual navigation, states $s$ consist of $2$-D location and orientation $(x,y,\theta)$, whereas observations $o$ are RGB images corresponding to the panoramic view at the given state $s$. To specify point-goal navigation with arbitrary distance, we define $\rho_0$ and $\rho_g$ as the distributions for start states $
s_0 \in \mathcal{S}$ and goal states $g \in \mathcal{G}$, where $\mathcal{S}$ and $\mathcal{G}$ is identical. Furthermore, we define the reward function as $r(s, a, g) = 0$ if $s = g$ and $r(s, a, g) = -1$ otherwise. Given the start $s \sim \rho_0$ and goal $g \sim \rho_g$ locations, the objective is to find a policy $\pi: \mathcal{O}_{\mathcal{S}}^{*} \times \mathcal{O}_{\mathcal{G}} \rightarrow \mathcal{A}$ that maximizes the expected accumulative rewards for all the start and goal pairs $\mathbb{E}_{s_{0} \sim \rho_0, g \sim \rho_{g}} \left[ V(s_{0}, g)\right]$, where $V(s_{0}, g) = \mathbb{E} \left[ \sum_{t=0}^{T-1}r(s_t, a_t, g) | s_{0}, g \right ]$.

For our hierarchical approach solving point-goal navigation, we assume that the agent is additionally provided with a 2-D binary occupancy image $M$ as the \emph{high-level} guidance, where a cell has value $1$ if it is occupied and $0$ if it is free. Although the dimensions of the image and its cell values do not necessarily correspond to the 3-D environment where the agent is, we will generally assume they are closely related. Furthermore, we discretize the image $M$ into a $n \times n$ grid world where each cell $s_i$ defines a corresponding local map patch $m_i$, a fixed-size sub-image cropped from $M$ and centered at $s_i$. These map patches will be used to provide \emph{high-level} navigation guidance (i.e. $m$ in Figure~\ref{fig:framework}) in the proposed framework.

For the \emph{low-level} navigation controller, we formulate it as goal-conditioned reinforcement learning (RL) problem. In RL~\citep{Sutton2018}, the agent learns to make decisions in the environment, modeled as a Markov decision process (MDP)~\citep{Bellman1957}, through interaction. In goal-conditioned RL, the reward function $\mathcal{R}: \mathcal{S} \times \mathcal{A} \times \mathcal{G} \rightarrow \mathbb{R}$ and the policy $\pi: \mathcal{S} \times \mathcal{G} \rightarrow \mathcal{A}$ will be conditioned on a particular goal $g \in \mathcal{G}$, whereas others remain the same. Since the low-level controller in our approach is only trained with local behaviors with short horizons, we formulate each local behavior as a MDP, where both states and goals are still represented by RGB images corresponding to the panoramic view of the states and goals. Partial observability is ignored. The goal is to find the local policy $\pi: \mathcal{O_{\mathcal{S}}} \times {O_{\mathcal{G}}} \rightarrow \mathcal{A}$ that maximizes the expected accumulative rewards for all start and goal locations.

\section{Related work}

\textbf{Planning and learning with 2-D maps} Our method requires a rough 2-D map to be given as initial high-level information. Prior work also considers using such 2-D maps. \citet{brunner2018MapReader} use an accurate global 2-D map and plans actions by localizing the agent on the map. \citet{gao2017intention} also proposes a deep neural network for navigation with a local 2-D map provided as additional signal. \citet{liu2020hallucinative} requires a global context (e.g., an accurate global layout) of the environment to guide the local controller in novel domains. All of the above methods require accurate high-level information; however, they all illustrate promising zero-shot generalization performance in novel tasks, suggesting that high-level maps and contexts are useful for navigation.
 
\textbf{Hierarchical navigation} We adopt a general hierarchical framework that combines a high-level planner and a low-level controller based on goal-conditioned RL methods. \citet{nasiriany2019goal-conditioned-planning} proposes LEAP, where the high-level landmark proposer outputs the sequence of landmarks by optimizing over a feasibility vector, and the local controller is a goal-conditioned temporal-difference model \citep{pong2018TDMs}. Similar to \citet{savinov2018semiTopologicalMem} that learns the global layout of the environment, \citet{eysenbach2019SoRB} proposes SoRB, combining planning with RL. SoRB constructs the high-level topological graph from training experience and proposes the landmarks by searching on the graph. A DDPG \citep{casas2017DDPG} or DQN \citep{mnih2015DQN} controller is trained depending on the action spaces. Similarly, Map Planner~\citep{huang2019mapplanner} also combines a graph-based high-level planner and a goal-conditioned UVFA~\citep{schaul2015UvFA} controller. Hierarchical methods above achieve a robust high-level planning and an efficient low-level policy learning; however, because of the high-level information built from experienced environments, it is hard for them to generalize to novel environments. 

\textbf{Goal-conditioned RL} We build our local low-level controller based on deep goal-conditioned RL methods. \citet{kaelbling1993learning} is early work that defines goal-conditioned value functions and reward functions. \citet{schaul2015UvFA} proposes learning universal goal-conditioned value functions using deep neural networks. \citet{NIPS2017HER} proposes a goal relabeling strategy that dramatically improves the learning efficiency of goal-conditioned values and policies. However, as \citet{huang2019mapplanner} argues, purely goal-conditioned model-free RL methods (e.g., HER \cite{NIPS2017HER} and UVFA \cite{schaul2015UvFA}) are efficient in solving tasks with short horizons, but are inefficient for long-horizon tasks due to difficulties in exploration and local optima. This suggests that pairing these locally robust methods with high-level information (e.g., from rough 2-D maps) is a promising direction.

\section{Approach}
\label{sec:method}
Our navigation framework consists of three components (shown in Figure~\ref{fig:framework}):
\begin{enumerate}[leftmargin=*]
    \item A dynamic topological map that can be initialized from a rough 2-D map and proposes the local 2-D map patches of the future landmarks as navigation sub-goals.
    \item A generative model that generates observations corresponding to 2-D map patches of landmarks.
    \item A goal-conditioned controller that can execute local navigation to reach these observations.
\end{enumerate}
We will discuss components ($2$) and ($3$) first, then return to the dynamic topological map ($1$).

\begin{figure}[ht]
    \centering
    \includegraphics[width=10cm, height=4cm]{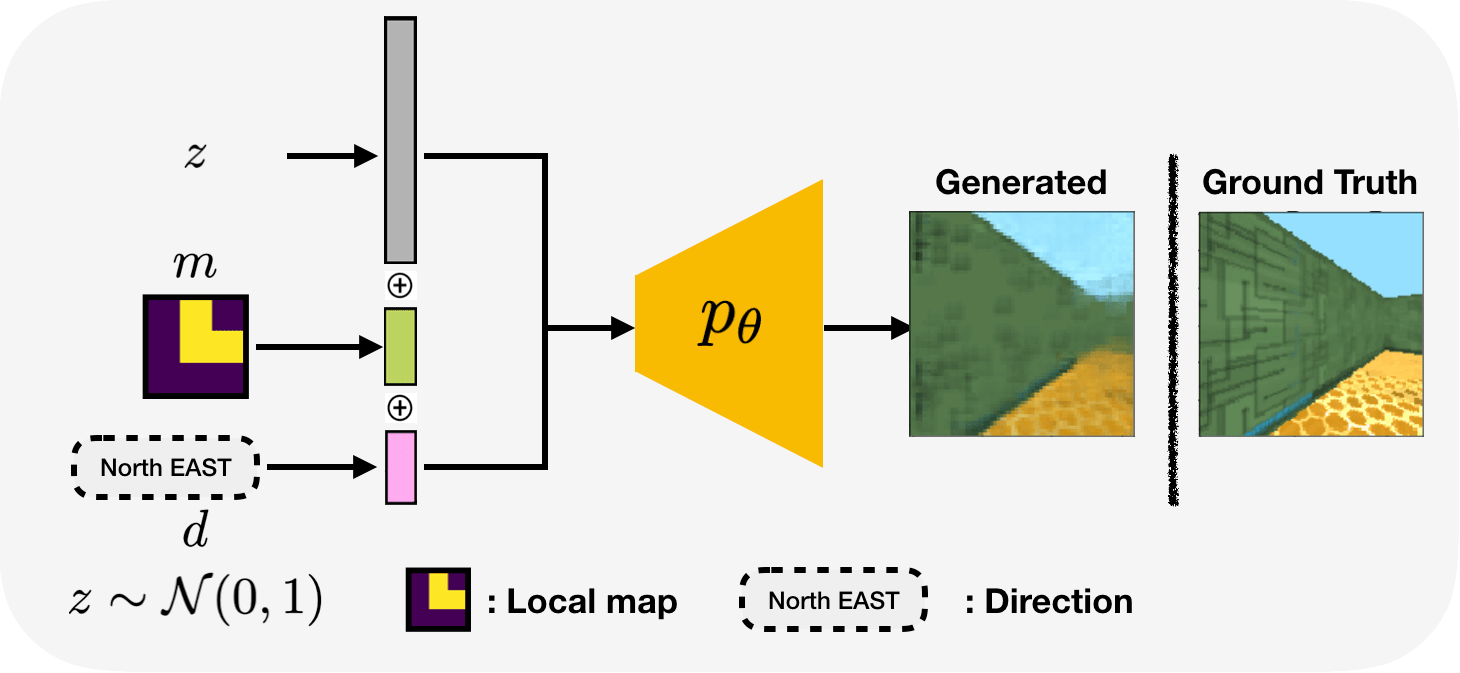}
    \caption{Process of generating landmark observation in a particular direction.}
    \label{fig:observation_generator}
\end{figure}

\subsection{Generating observation subgoals corresponding to landmarks}
\label{sec:cvae}
The landmark observation generator infers the appearance of intermediate landmarks during navigation using local 2-D map patches. For example, a cross shape on a 2-D map should infer an intersection. Similarly, in visual navigation, we want the observation generator to generate possible observations for any particular position based on the provided 2-D rough map. The generated observation will be used by the goal-conditioned controller as the next navigation subgoal.

We build the landmark observation generator using a conditional variational autoencoder (CVAE) \cite{sohn2015CVAE-1, walker2016CVAE-2}. The basic VAE architecture in our generator is adopted from \citet{ha2018world}. However, different from the original VAE that takes images as the only input, the CVAE model in our generator has two extra inputs as the conditional information (see Figure~\ref{fig:observation_generator}): the local map patch cropped from the rough 2-D map, and a direction indicator. Intuitively, our generator should generate the observation in a particular orientation based on the given local map and the direction.

Formally, let $X$ be the observation to be generated, and $c = (m, d)$ be the conditional feature that contains both local map patch $m$ and the direction $d$. The encoder of the CVAE is $q_\phi$ and the decoder is $p_\theta$. The latent variable is denoted by $Z$, where $Z \sim \mathcal{N}(0, I)$. The latent variable $Z$ is conditioned on $c$ by concatenating the feature vectors together. We train the CVAE by maximizing the following variational lower bound:

\begin{equation}\label{eq:CVAE}
    \mathcal{L}_{\text{CVAE}} = - D_{\text{KL}} \left[ q_\phi(Z|X,c) \,\middle\|\, p(Z|c) \right] + \mathbb{E} \left[ \log p_\theta(X|Z, c) \right]
\end{equation}

To collect training data, we use a random policy to collect images from the 3-D environment and label them with the corresponding local map patches and directions. After training, we use the decoder $p_\theta$ of the learned CVAE to generate observations. Given an arbitrary position on the rough 2-D map, the inputs to our generator are the cropped local map patch $m$ (i.e., $3 \times 3$ binary image), a target direction $d$ (one-hot encoding of the $8$ cardinal and ordinal directions), and a latent feature $z$ sampled from a $64$-D Gaussian distribution $\mathcal{N}(0, I)$. These three components are concatenated and given to the learned decoder $p_\theta$ to generate observations $X$ ($32 \times 32$ RGB images) in the 3-D maze. Figure~\ref{fig:observation_generator} shows the procedure of generating the landmark observations. See Appendix B for the CVAE architecture and training details, and Appendix C for more generation results.

\subsection{Navigating to observation subgoals using a local goal-conditioned controller}
\label{sec:ddqn}
The goal-conditioned controller performs local navigation to reach a landmark corresponding to a generated image-based subgoal. We learn local controllers instead of global ones to support generalization -- local behaviors can be reused in both seen or unseen mazes. Additionally, local behaviors are easy to learn because of their short horizon length and low execution complexity.

We used goal-conditioned RL methods to learn the local goal-conditioned controller. The particular choice of goal-conditioned controller is flexible; any value-based or actor-critic method can be used. Since our action space is discrete, we learn a goal-conditioned double DQN \citep{mnih2015DQN,van2016double-DQN}. Formally, for reaching the given goal $g$, the reward function is defined as $r(o, a, g) = -1$ if $o \neq g$ and $r(o, a, g) = 0$ otherwise. Recall that all states, including $g$, are images. The objective to find a goal-conditioned policy $\pi$ that maximizes the expected return, averaged over a distribution of local start and goal location pairs, respectively $o \sim \rho_0$ and $g \sim \rho_g$. The Q-value for taking action $a$ in state $o$ is defined as $Q(o, a, g)$, which is approximated by a function approximator $Q_\psi(o, a, g)$. In deep double Q-learning~\citep{van2016double-DQN}, in order to learn $Q_\psi$, we minimize the mean squared-error loss:
\begin{equation}\label{eq:DDQN}
\mathcal{L}_\psi = \mathbb{E}_{\langle o, a, r, o', g\rangle \sim \mathcal{D}}[(y - Q_\psi(o, a, g))^2], \text{ where } y = r + \gamma Q_{\psi^-}(o', \arg \max_{a'}Q_\psi(o', a', g), g)
\end{equation}
In the above, $\mathcal{D}$ is the replay buffer, and $\psi, \psi^-$ parameterize the two Q-networks used in double DQN. After training, when presented with a new image-based subgoal (e.g., from the generator in Section~\ref{sec:cvae}), the agent follows the greedy local goal-conditioned policy $\pi_\psi(a|o, g) = \arg \max_{a} Q_\psi(o, a, g)$. See Appendix D for the double-DQN architecture and training details.

Additionally, we propose a variant of the goal-conditioned controller that automatically decides whether a landmark is reached. Previous work typically assume that the environment provides this signal (in our experiments, this is the ``oracle'' scenario); however, this is a strong assumption. Since the navigation agent is generating its own subgoals, we cannot expect the environment to determine whether subgoals have been reached yet. To detect this, we add an extra fully-connected output head that predicts the probability that the current position corresponds to the landmark subgoal.
This is a binary classification problem and is trained by adding an extra cross-entropy loss to equation \ref{eq:DDQN}.

\subsection{Planning high-level subgoals using the dynamic topological map}
\label{sec:dtm}
In principle, the goal-conditioned controller should be sufficient for navigating to the goal state $s_g$. However, as we show in an ablation experiment in Section~\ref{sec:ablation_study}, this does not allow us to generalize to new environments effectively. Instead, we will use the provided rough 2-D map to instantiate a dynamic topological map, plan a sequence of subgoals using the topological map, and update the topological map when the provided rough map and/or local controller is imperfect.

\subsubsection{Building a dynamic topological map}
Given start and goal positions on the rough 2-D map, the easiest way to use the rough map is to apply a grid-search method such as A$^*$ on the 2-D map to obtain a high-level trajectory. Then, the observation generator can be used to generate the observation for every position along the trajectory. Finally, given the sequence of landmark observations, the local goal-conditioned controller is used to navigate to every landmark sequentially. This strategy is an open-loop control method and suffers from lack of robustness, as will be demonstrated in an ablation experiment in Section~\ref{sec:ablation_study}.

Instead of using the 2-D map directly, we build a directed topological graph $\mathcal{G}$ that can help the agent understand its learned behaviors. Specifically, we define $\mathcal{G}$ as follows:
\begin{equation}\label{eq:TBM}
    \mathcal{G} = (\mathcal{V}, \mathcal{E}) \quad \textrm{where} \quad \mathcal{V} = \{m_i \}_{i=1}^{N} \quad \mathcal{E} = \{e_{i, j}\}_{i,j=1}^{N,N}
\end{equation}
A node in the graph $m_i$ corresponds to a feasible local $3 \times 3$ map patch in the provided 2-D map $M$; a patch is feasible if the center cell $s_i$ is empty space ($0$). An edge $e_{ij}$ denotes the feasibility of traversal between nodes; its value is $1$ if the local controller can navigate the agent from $s_i$ to $s_j$ and $0$ if the behavior is not learned. Since the local controller may not behave symmetrically, we maintain separate edges for $e_{ij}$ and $e_{ji}$.
The graph is instantiated by defining nodes for all feasible local map patches in $M$, and bidirectional edges between neighboring feasible map patches (i.e., assume all empty space specified in $M$ is traversable in any direction). This is clearly optimistic, since the rough 2-D map may be incorrect and the low-level controller may fail on some edges. Our planning algorithm described next will update the topological map by simply removing edges where failures are encountered, and replan with the updated topological map.

\subsubsection{Planning with and updating the dynamic topological map (Algorithm~\ref{alg:plan-with-DTM})}
Given the start and goal positions ($s_0, s_g$) in the rough 2-D map, as well as the topological graph $\mathcal{G}$, we first apply Dijkstra's algorithm to search for the shortest path between $s_0$ and $s_g$. Then, instead of following the path in an open-loop fashion, the topological graph only outputs the first local map $m$ as the next landmark. The generator $p_\theta$ uses $m$ to generate the corresponding observation $o$, which is then given to the local controller $\pi_\psi$ as a subgoal to navigate to. If the landmark is reached, the graph will keep the traversed edge, and the next landmark is returned. Otherwise, the edge will be deleted from the graph, and a new high-level path is planned. This closed-loop strategy allows the agent to always use the latest graph, avoiding unreachable landmarks and recovering from failure. See Algorithm~\ref{alg:plan-with-DTM} in Appendix A for the complete pseudocode. 
\label{aproach_train}

\section{Experiments}
\label{sec:result}
We first introduce the DeepMind Lab 3-D maze environment and the approaches we compare our method against. Next, we demonstrate that our method achieves better navigation performance both in seen mazes and zero-shot generalization in unseen mazes (Sections~\ref{subsec:multi_goal_seen} and \ref{subsec:generalize_unseen}). We also show that our method can still solve some navigation tasks even when the provided 2-D rough map is partially wrong (Section~\ref{subsec:imprecise_map}). Finally, we include an ablation study (Section~\ref{sec:ablation_study}).

\textbf{Environment} DeepMind Lab~\citep{beattie2016deepmind} is a benchmark environment for visual navigation in complex 3-D mazes. The agent receives first-person RGB images as observations; note that this is a panoramic observation in all $8$ directions (see Appendix C for examples). Since our focus is on generalization and not on long-horizon RL, we simplify the action space by providing the agent with $4$ discrete actions (up/down/left/right), which moves the agent by one tile in the corresponding direction, where one tile corresponds to a $100 \times 100$ block in the original 3-D maze. Despite this simplification, navigation in the modified 3-D mazes is still non-trivial due to partial observability.  To further demonstrate the utility of our method in complex control settings, we also include results (Section~\ref{sec:new results}) of a hard version of Deepmind Lab where the discrete actions only change the accelerations in the corresponding directions. We generated random 3-D mazes of $5$ different sizes, with $20$ mazes for each size.

\textbf{Baselines and Methods} The first baseline is a \textbf{random} policy. The second baseline is \textbf{HER}~\citep{NIPS2017HER}, a purely goal-conditioned model-free RL method. The third method is \textbf{Map Planner}~\citep{huang2019mapplanner}, a state-of-the-art hierarchical method combining a learned high-level topological map and a low-level goal-conditioned controller based on UVFA~\citep{schaul2015UvFA}. We compare against two variants of our method: \textbf{our-oracle} obtains the landmark-reaching signal from the environment, whereas \textbf{our-pred} must predict this signal in the controller (see Section~\ref{sec:ddqn} for details). The input at each time step to all agents except Map Planner is the panoramic observation; Map planner receives the true state as input, which provides it an advantage. See Appendix A for training and testing details of our method and baselines. 

\textbf{Evaluation} Navigation to random goals from random positions in random maps is a multi-task problem; we report the average success rate $\rho$ of each method in a fixed set of sampled tasks to evaluate performance. Since task performance correlates strongly with navigation distance, we report $\rho$ separately for different start-goal lengths. In particular, for each distance in $\{1, 5, 10, \ldots, 50\}$, we randomly sample $50$ start-goal pairs from the evaluation maze(s). All methods are trained and evaluated on the same set of sampled start-goal pairs and mazes. The reported success rate for distance $d$ is defined as $\rho^{d} = \frac{n_{\text{success}}^{d}}{n_{\text{total}}^{d}}$, where $n_{\text{success}}^{d}$ is number of the successful navigation tasks and $n_{\text{total}}^{d}$ is the total number of navigation tasks for distance $d$. All methods are given a budget of $100$ time steps.

\subsection{Multi-goal navigation tasks in seen mazes}
\label{subsec:multi_goal_seen}
We first consider navigation in seen mazes, i.e., the same maze is used during both training and evaluation, but different start-goal pairs may occur. In each run, for each size, the maze is randomly sampled from the $20$ mazes. Except for the random policy, we train $5$ runs with different random seeds for each the method. We report the mean and standard error of the results from the $5$ runs. 

\begin{figure}[ht]
\begin{subfigure}{0.3\textwidth}
    \centering
    \includegraphics[scale=0.11]{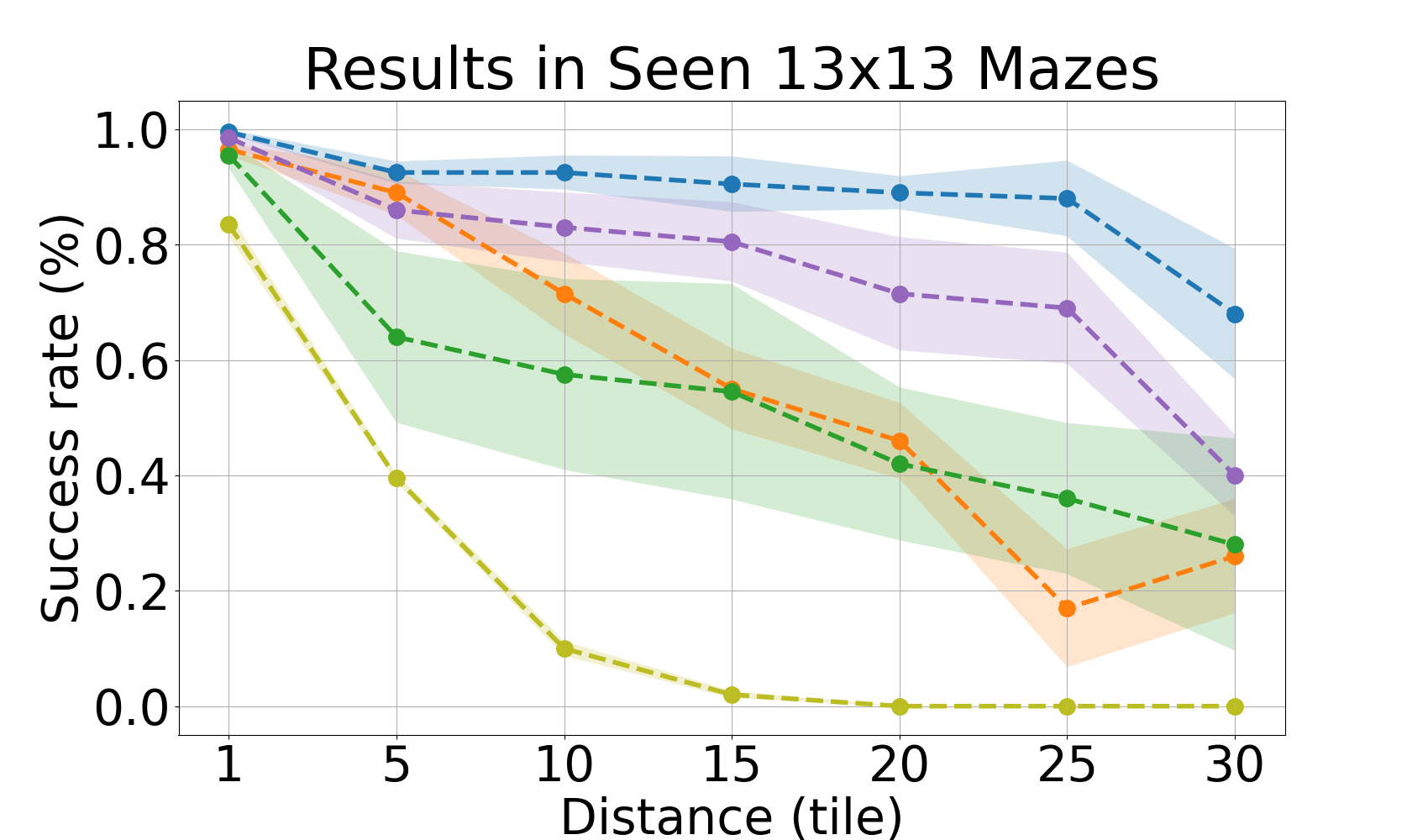}
    \label{fig:scene_1_13_main}
\end{subfigure}
\begin{subfigure}{0.3\textwidth}
    \centering
    \includegraphics[scale=0.11]{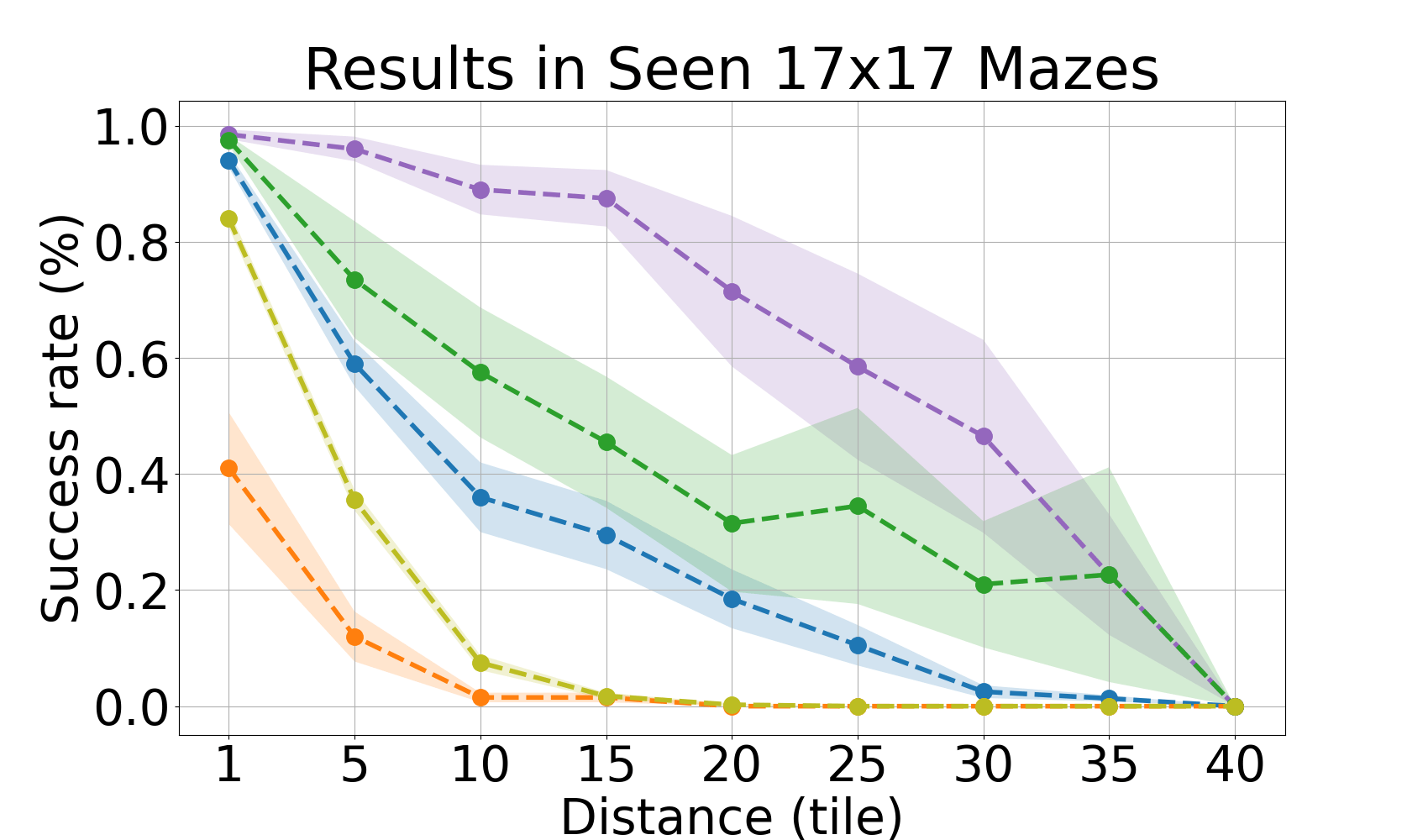}
    \label{fig:scene_1_17_main}
\end{subfigure}
\begin{subfigure}{0.3\textwidth}
    \centering
    \includegraphics[scale=0.11]{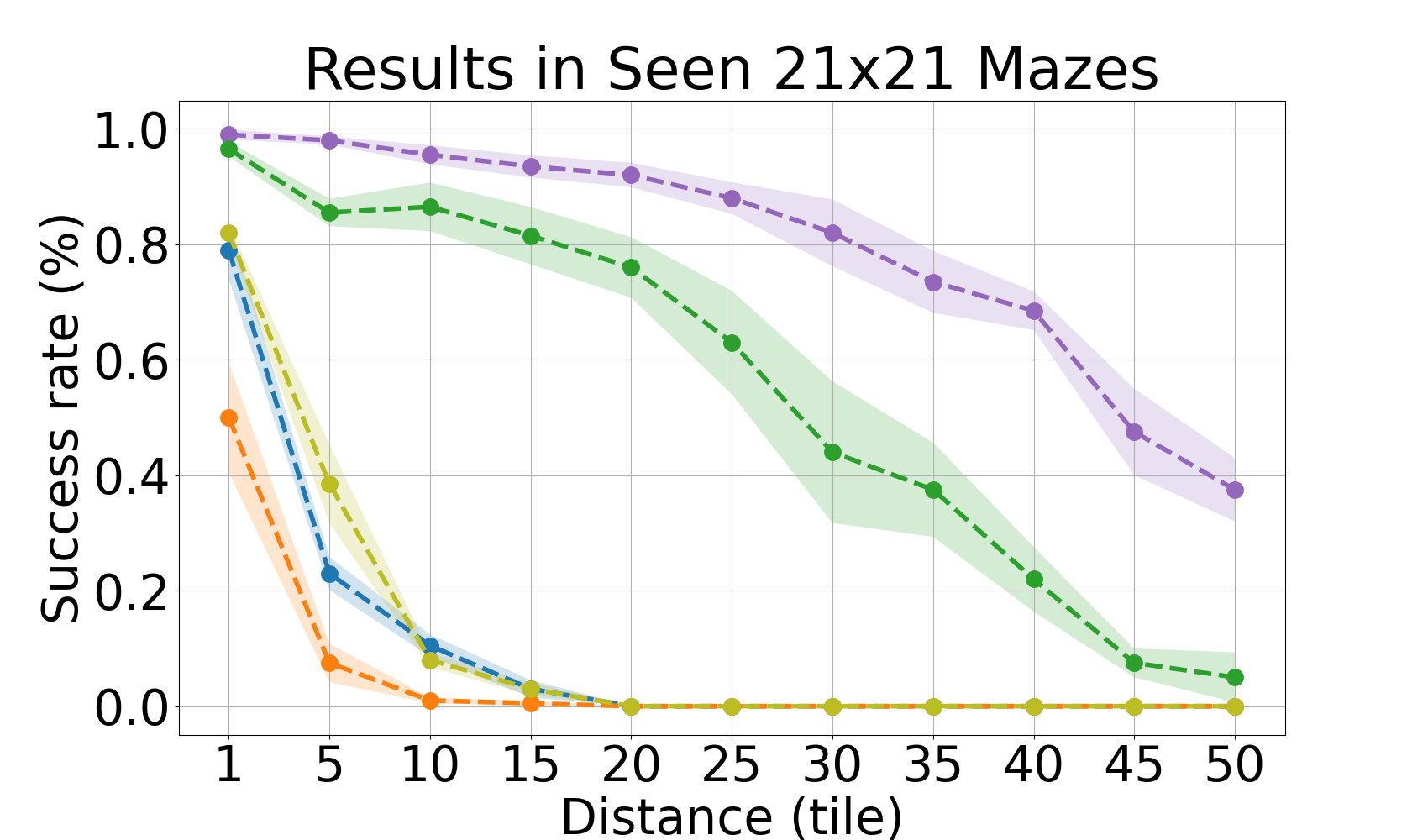}
    \label{fig:scene_1_21_main}
\end{subfigure}

\begin{subfigure}{0.3\textwidth}
    \centering
    \includegraphics[scale=0.11]{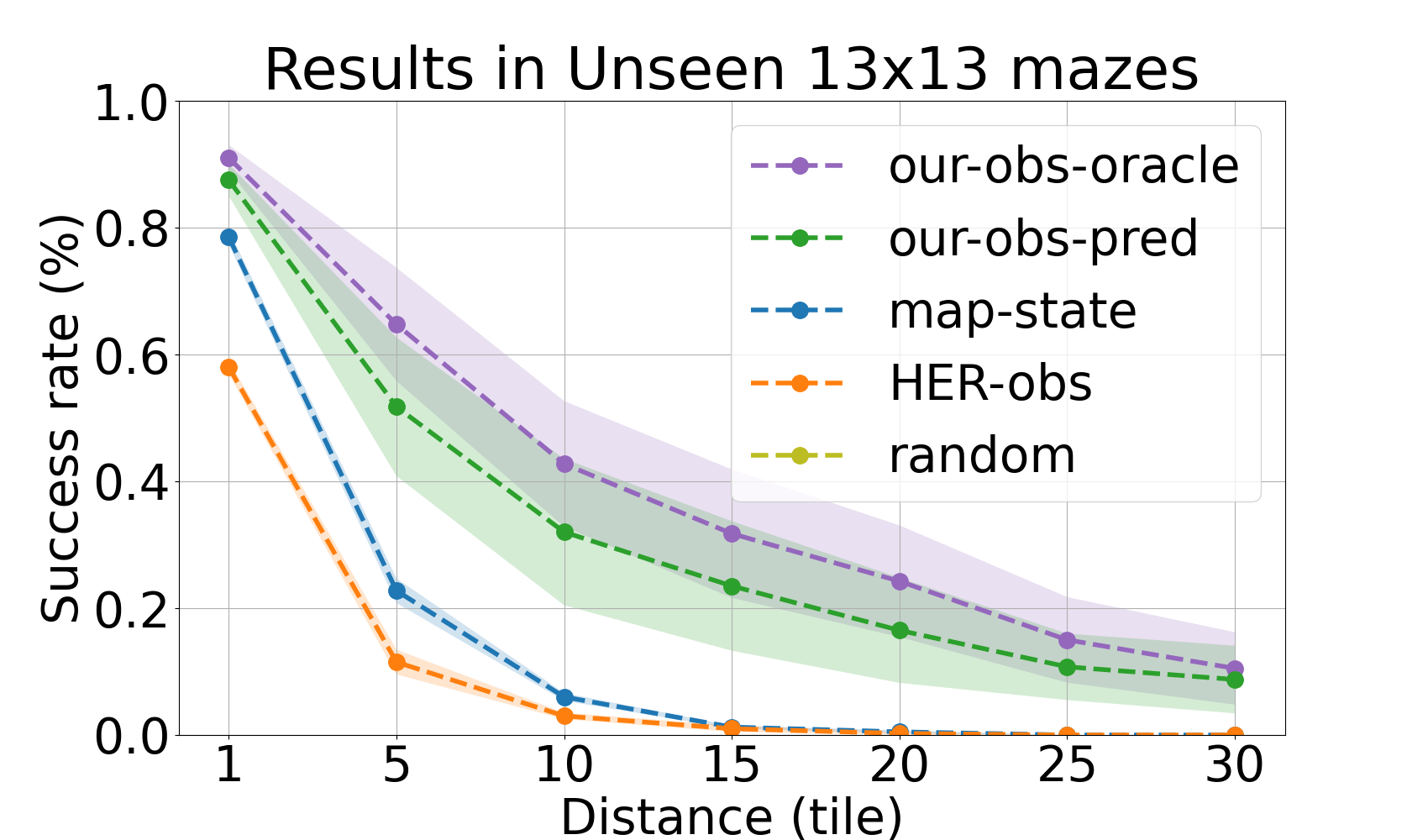}
    \label{fig:scene_2_13_main}
\end{subfigure}
\begin{subfigure}{0.3\textwidth}
    \centering
    \includegraphics[scale=0.11]{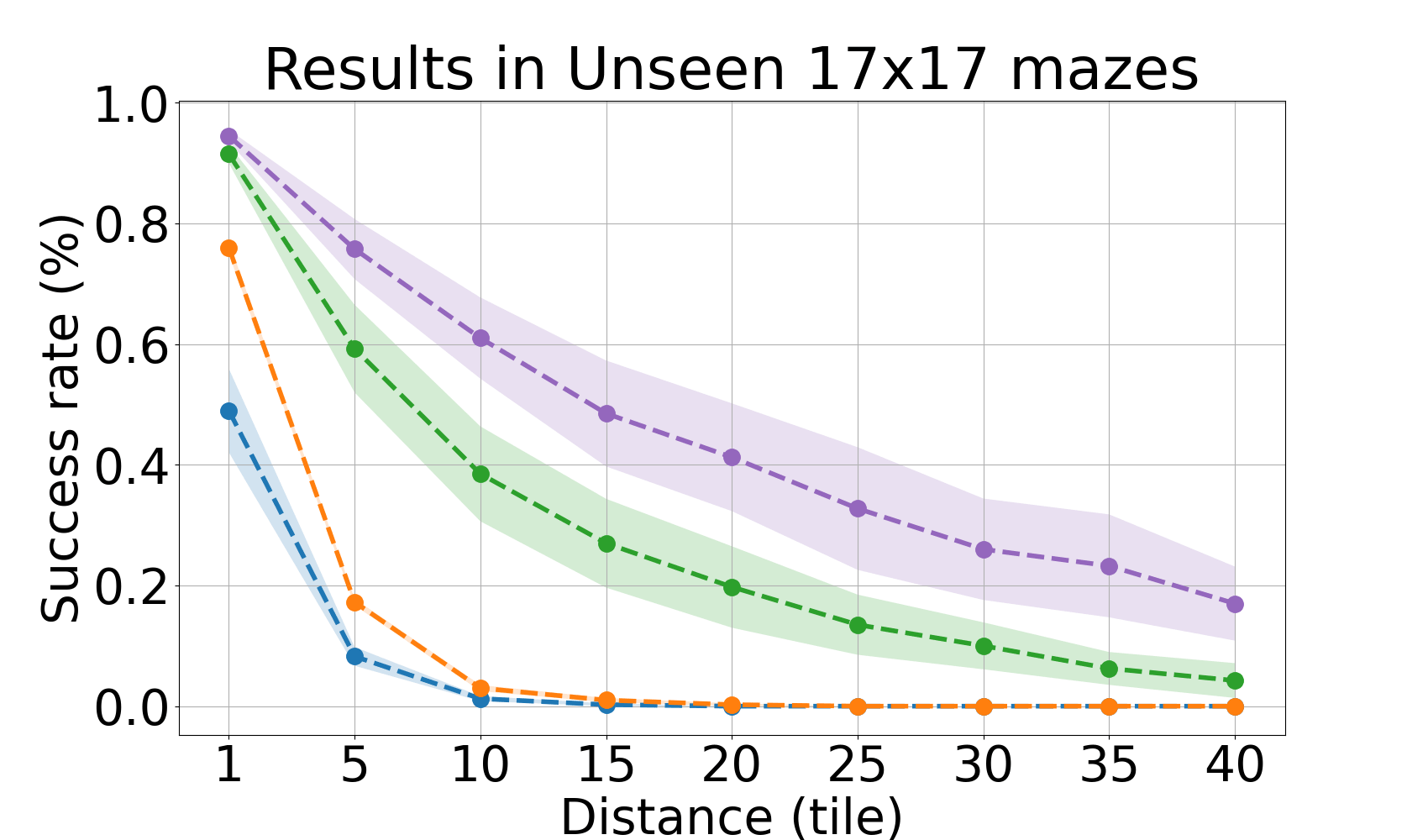}
    \label{fig:scene_2_17_main}
\end{subfigure}
\begin{subfigure}{0.3\textwidth}
    \centering
    \includegraphics[scale=0.11]{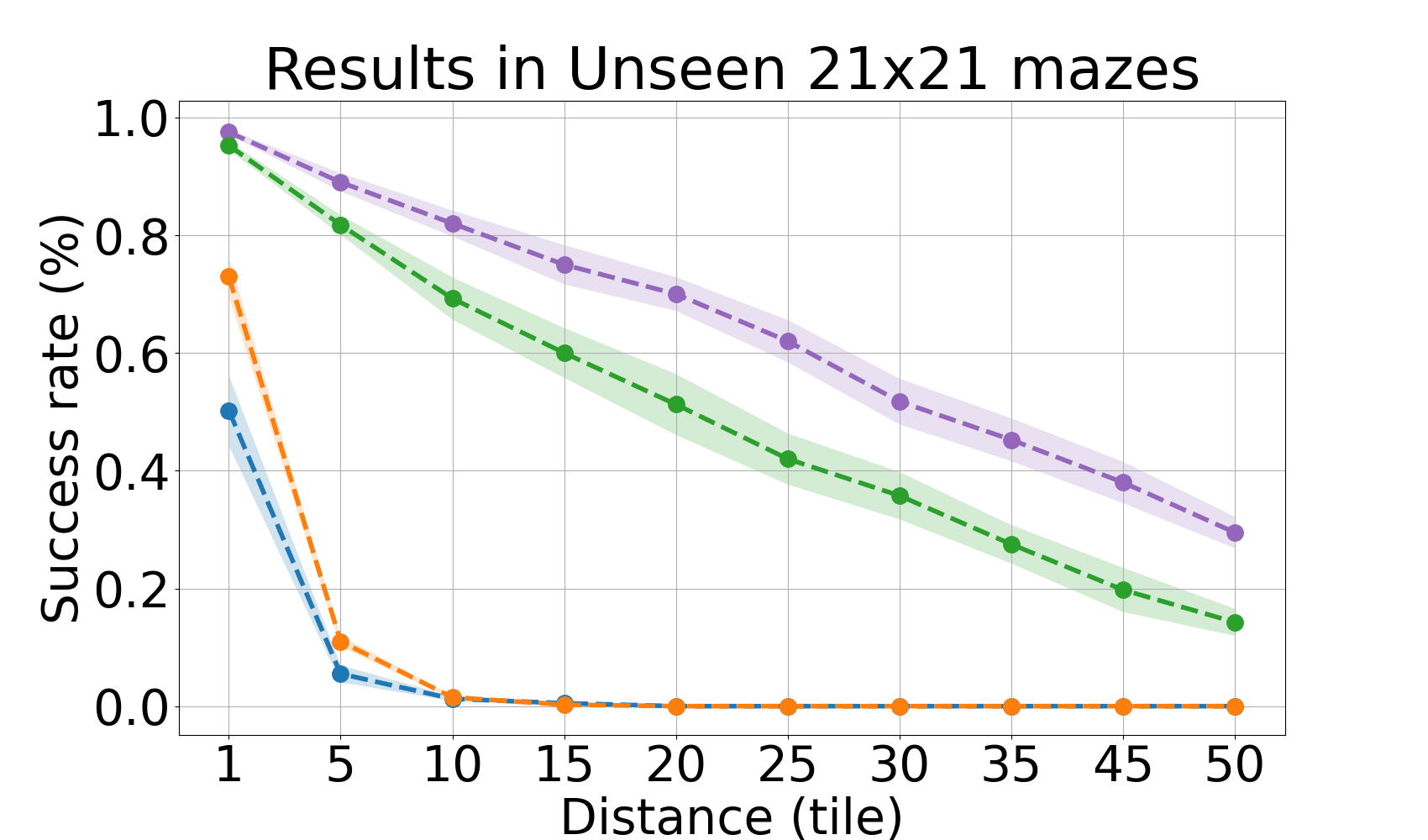}
    \label{fig:scene_2_21_main}
\end{subfigure}
    \caption{Average success rate for each distance in seen and unseen mazes in the \textbf{discrete} Deepmind Lab. The first row shows the results for seen mazes. The second row shows the results for unseen mazes.}
    \label{fig:results_main}
\end{figure}

Figure \ref{fig:results_main} shows that in all maze sizes, the performance of all methods decrease exponentially when the distance increases. However, our method shows a significantly slower decay, especially in larger mazes. The sharp decrease of the random policy (e.g., $\rho^{10} \leq 10\%$) shows that longer-distance navigation is non-trivial. HER performs reasonably well for short distances, but performance drops significantly when the maze size increases or the distance increases. Additionally, in larger mazes, when the policy of HER is greedy, it actually performs worse than the random policy.

Our method outperforms all the baselines, except in $13 \times 13$ where Map Planner performs the best (but recall that Map Planner uses the true state while our methods use observations). Map Planner performs better in small mazes because it is easier to explore and cover all states. However, when the maze size increases, fully covering the maze with a learned topological map becomes harder. Although the local UVFA controller in Map Planner is reliable, the cumulative error is still problematic for longer distances. In contrast, our method performs consistently for the same navigation distance across different maze sizes.

\begin{figure}[t]
\begin{subfigure}{0.3\textwidth}
    \centering
    \includegraphics[scale=0.1]{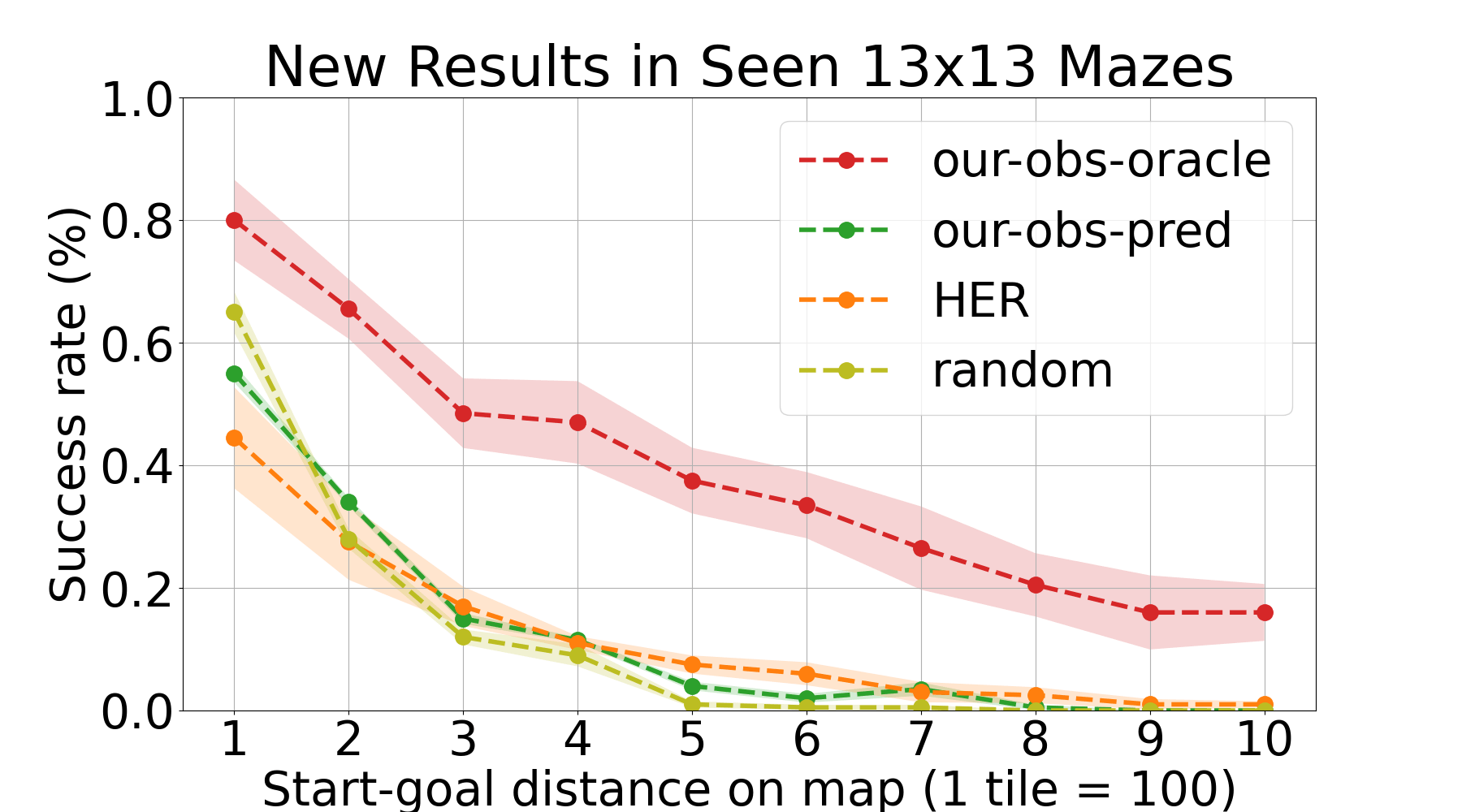}
    \label{fig:new_scene_1_13_main}
\end{subfigure}
\begin{subfigure}{0.3\textwidth}
    \centering
    \includegraphics[scale=0.1]{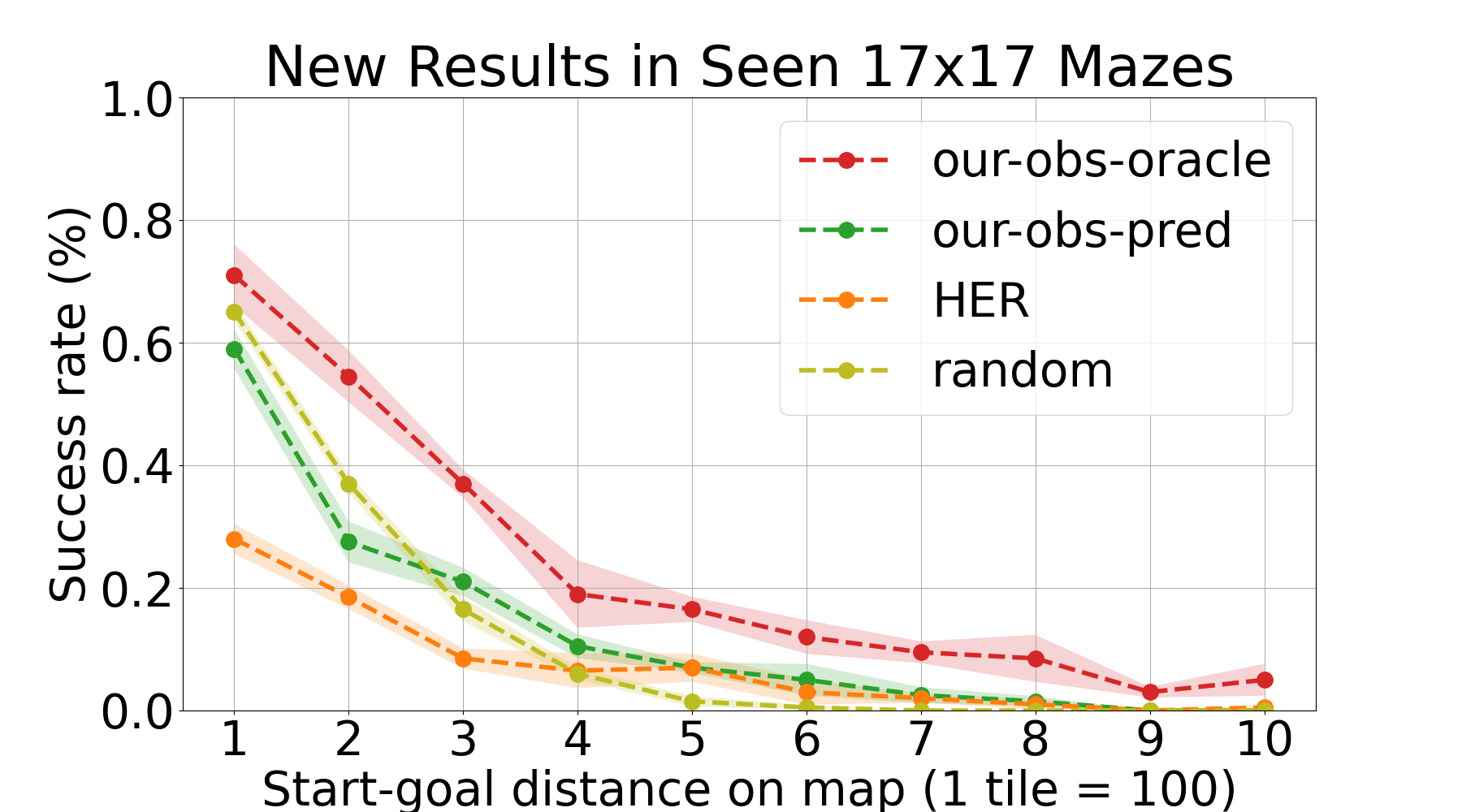}
    \label{fig:new_scene_1_17_main}
\end{subfigure}
\begin{subfigure}{0.3\textwidth}
    \centering
    \includegraphics[scale=0.1]{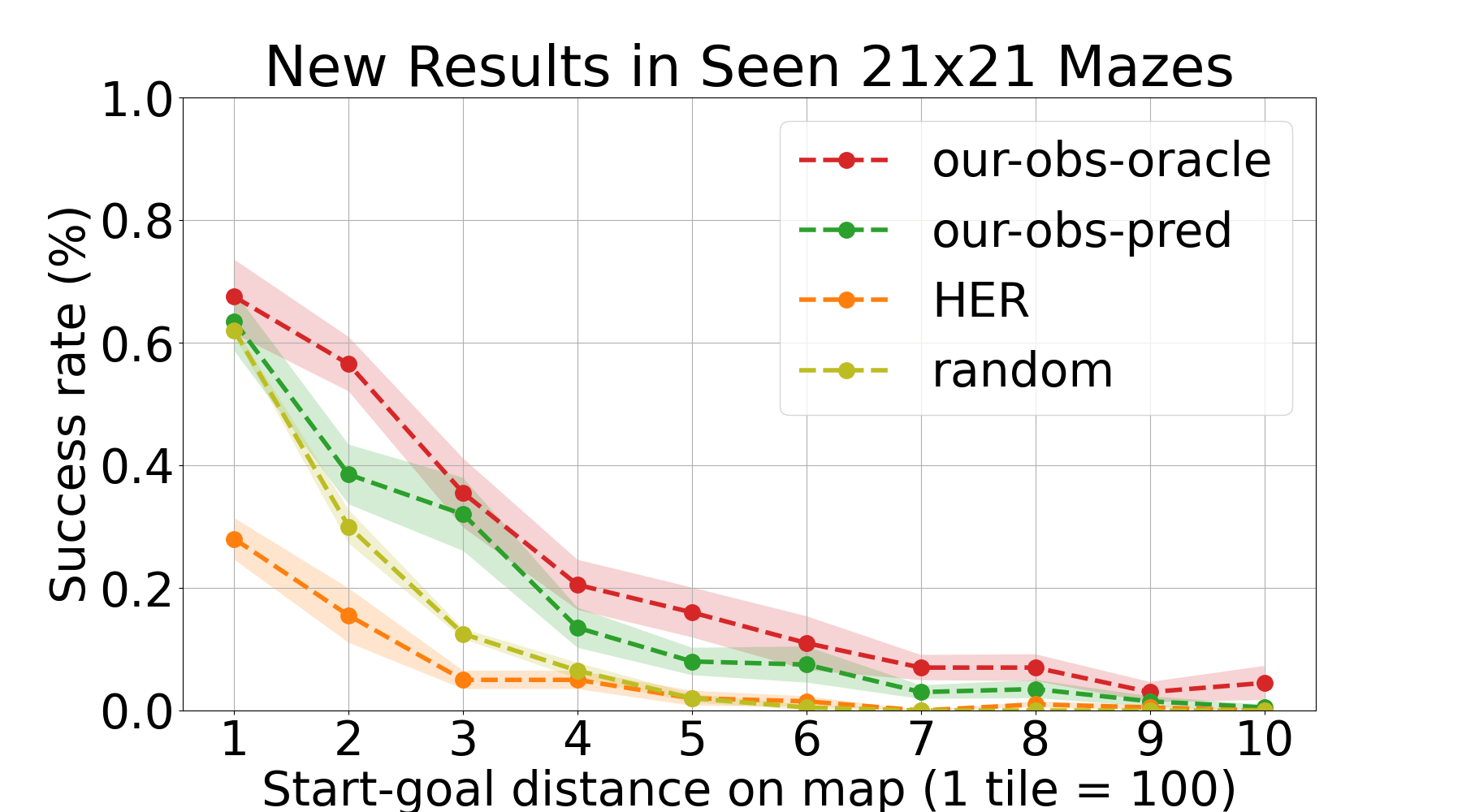}
    \label{fig:new_scene_1_21_main}
\end{subfigure}

\begin{subfigure}{0.3\textwidth}
    \centering
    \includegraphics[scale=0.1]{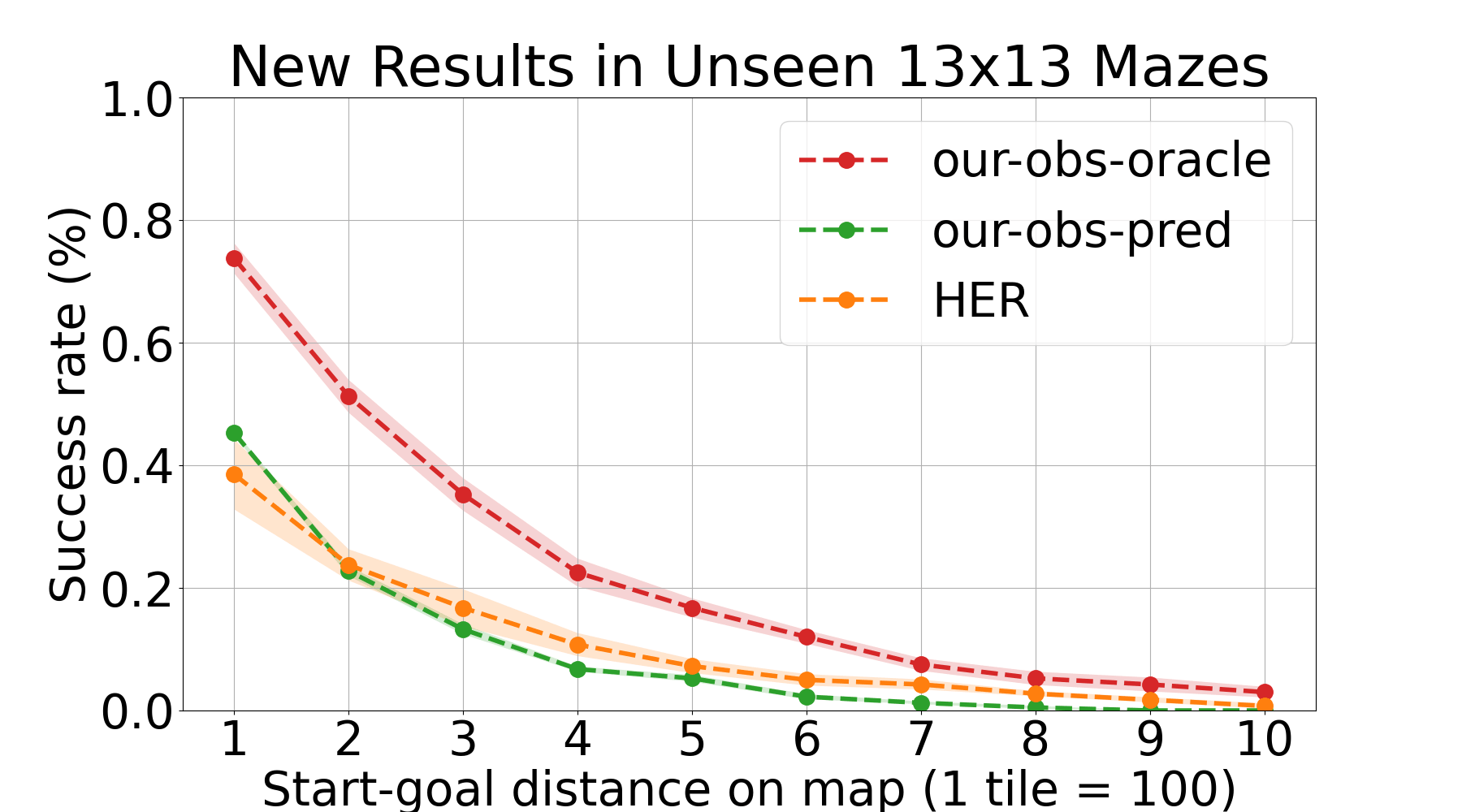}
    \label{fig:new_scene_2_13_main}
\end{subfigure}
\begin{subfigure}{0.3\textwidth}
    \centering
    \includegraphics[scale=0.1]{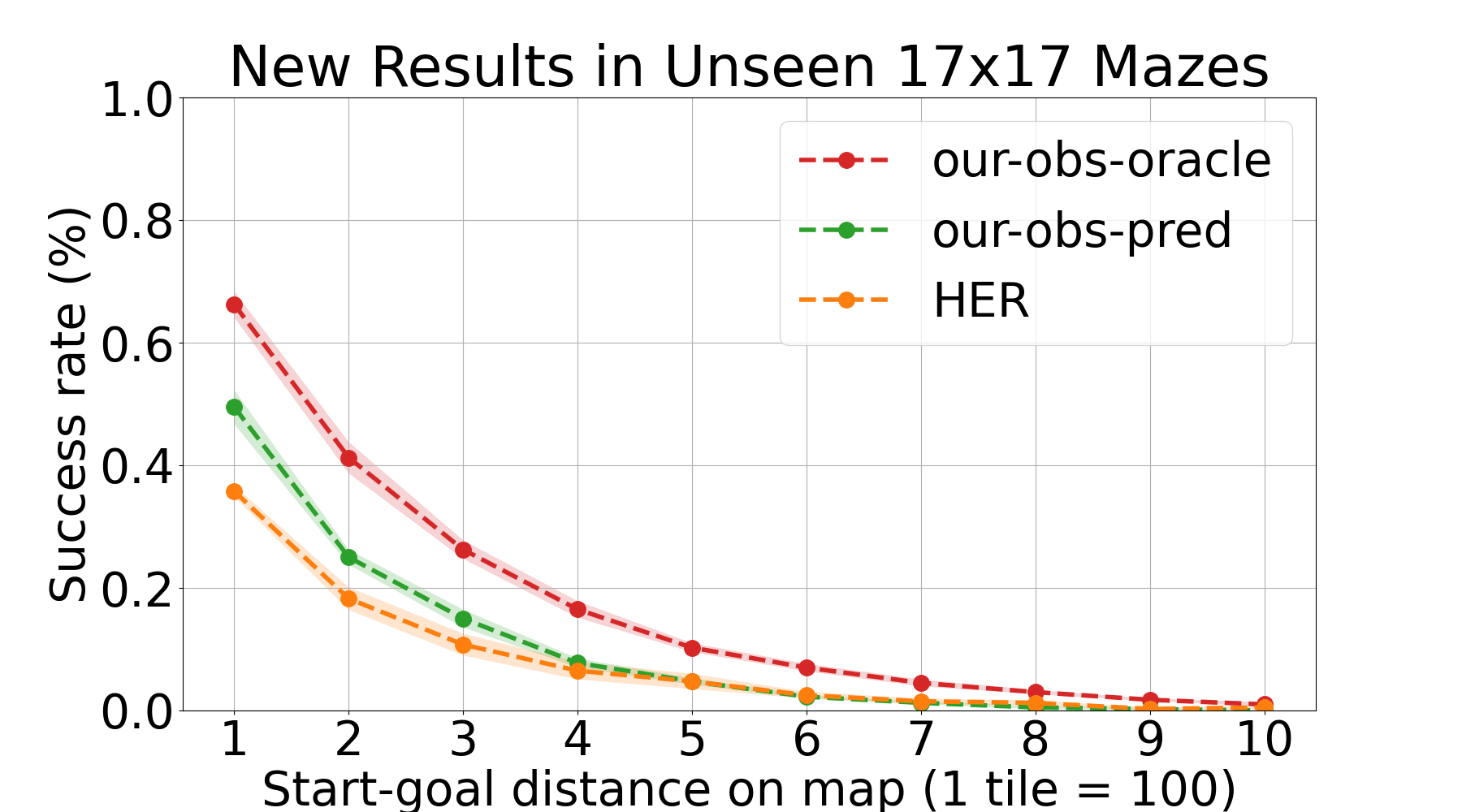}
    \label{fig:new_scene_2_17_main}
\end{subfigure}
\begin{subfigure}{0.3\textwidth}
    \centering
    \includegraphics[scale=0.1]{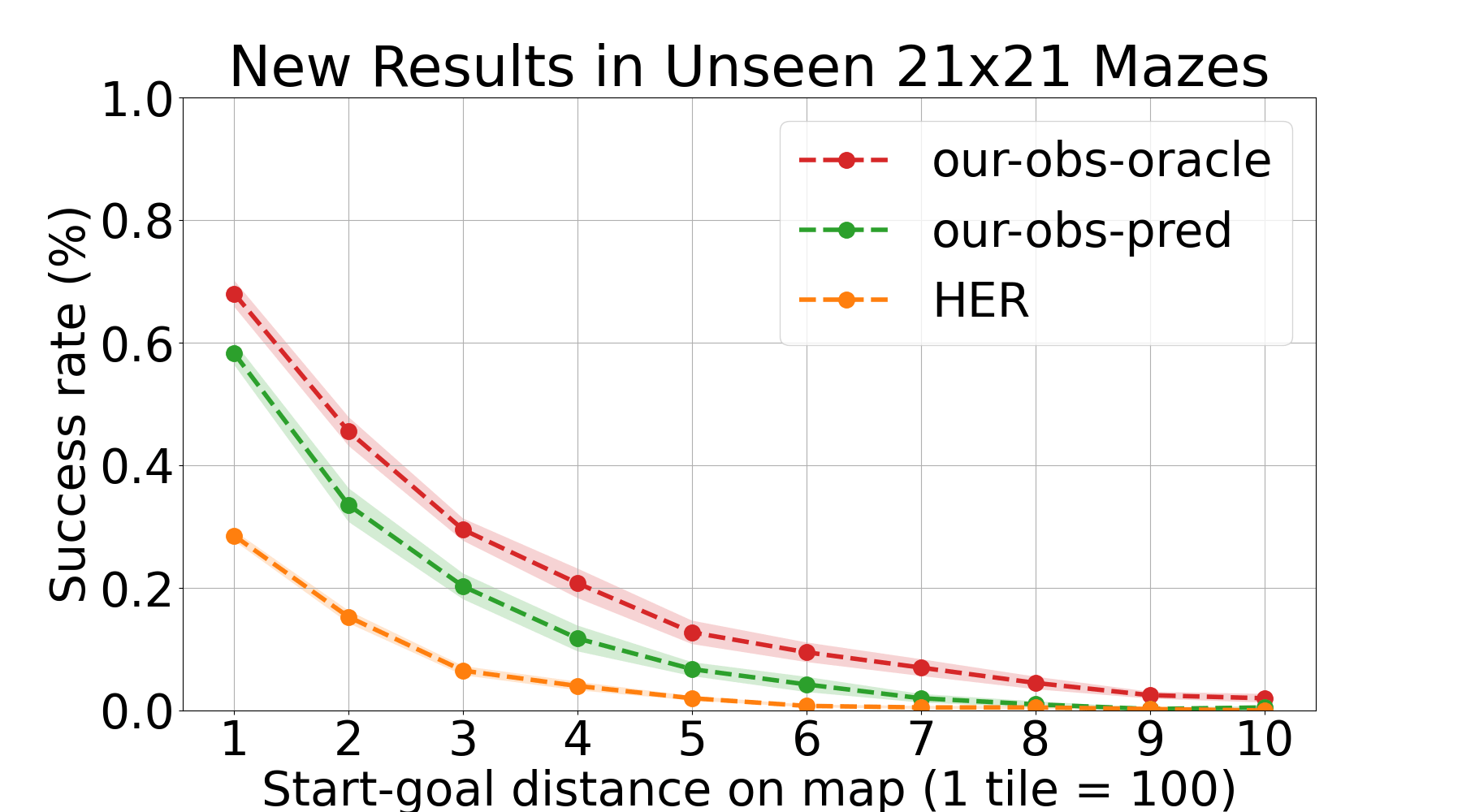}
    \label{fig:new_scene_2_21_main}
\end{subfigure}
    \caption{Average success rate for each distance in seen and unseen mazes in the \textbf{continuous} Deepmind Lab. The first row shows the results for seen mazes. The second row shows the results for unseen mazes.}
    \label{fig:new_results_main}
\end{figure}

\subsection{Generalization in unseen mazes}
\label{subsec:generalize_unseen}
We further test for zero-shot generalization performance, without allowing further fine-tuning on unseen mazes. All methods are trained in just one maze and evaluated on the remaining $19$ unseen mazes. Since Map Planner builds its topological map based on the replay buffer, we provide the buffer with $50$ randomly sampled true states from the novel mazes. Figure~\ref{fig:results_main} shows that our method significantly outperforms the other methods across all maze sizes. Even though Map Planner is provided states from the unseen maze to build its topological map, it fails to generalize to distances $>10$, and even performs slightly worse than HER. The high success rates for HER ($78\%$) and our methods ($90\%$) for short distances ($\leq$ 3) demonstrate that local behaviors are quite reliable and can generalize well to novel mazes when paired with a high-level dynamic topological map.

\subsection{Imprecise rough 2-D maps}
\label{subsec:imprecise_map}
\begin{figure}[t]
    \centering
    \begin{minipage}[b]{0.25\textwidth}
        \centering
        \includegraphics[width=.7\textwidth]{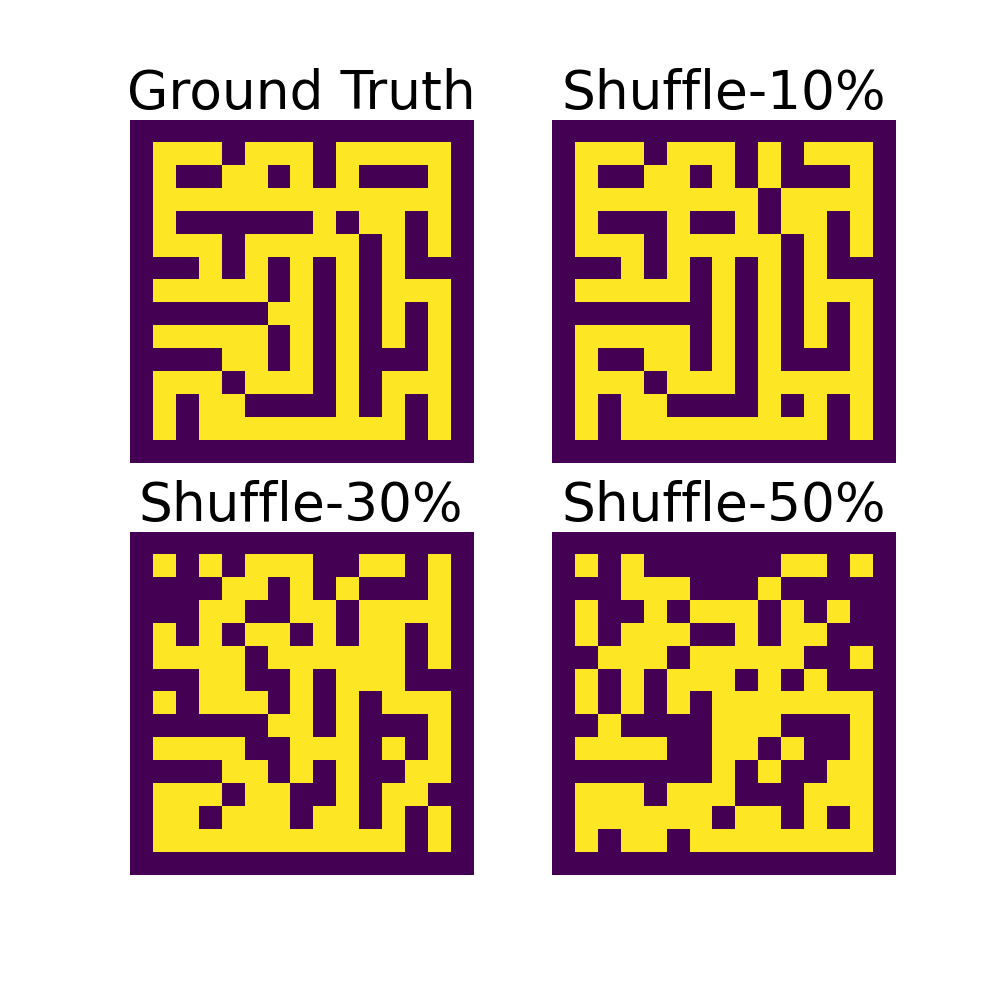}
        \caption{Flipped maps}
        \label{fig:imprecise_map}
    \end{minipage}
    \hfill
    \begin{minipage}[b]{0.35\textwidth}
        \includegraphics[width=\textwidth]{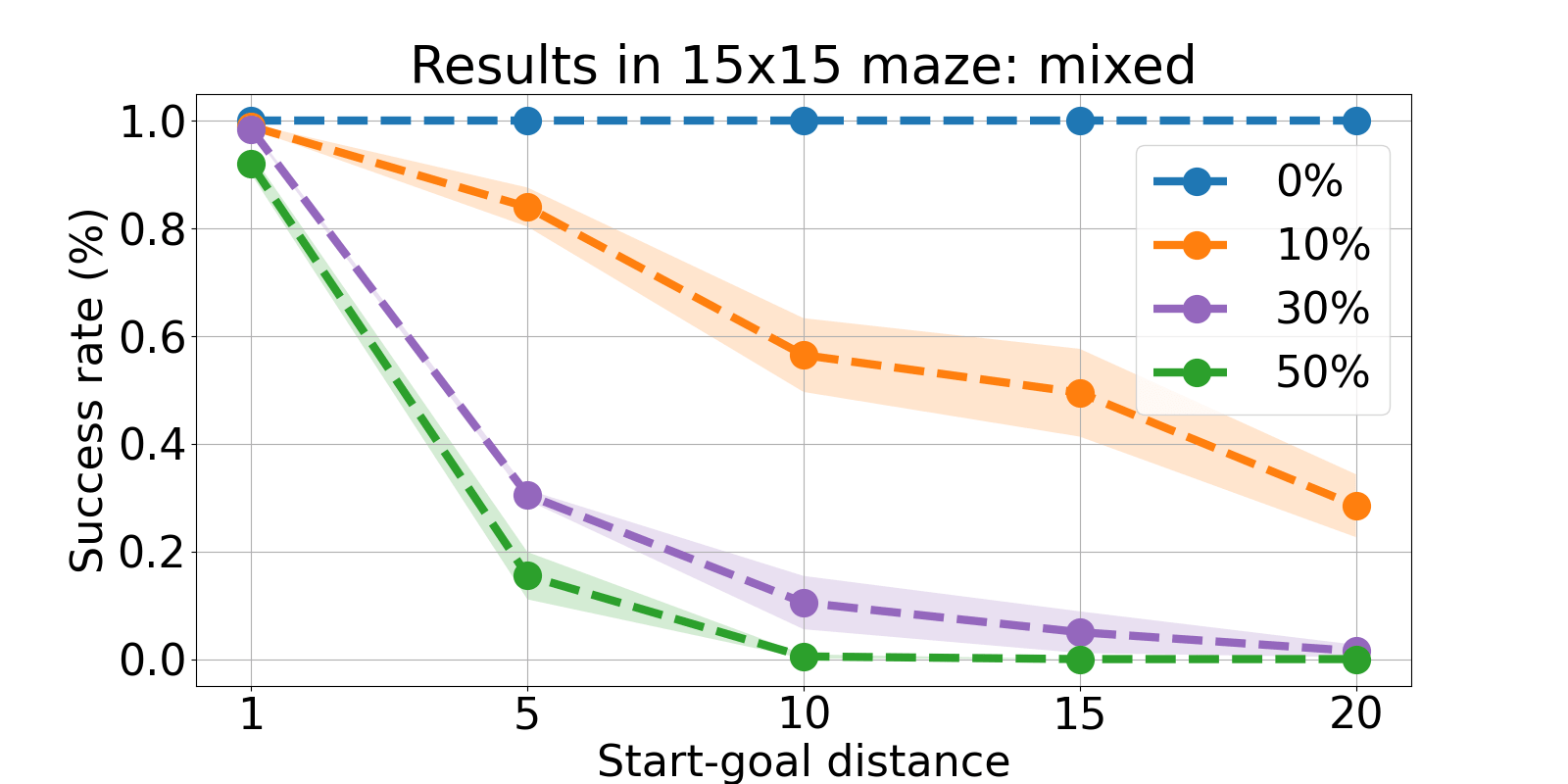}
        \caption{Imprecise maps}
        \label{fig:res_use_imprecise_map}
    \end{minipage}
    \hfill
    \begin{minipage}[b]{0.35\textwidth}
        \includegraphics[width=\textwidth]{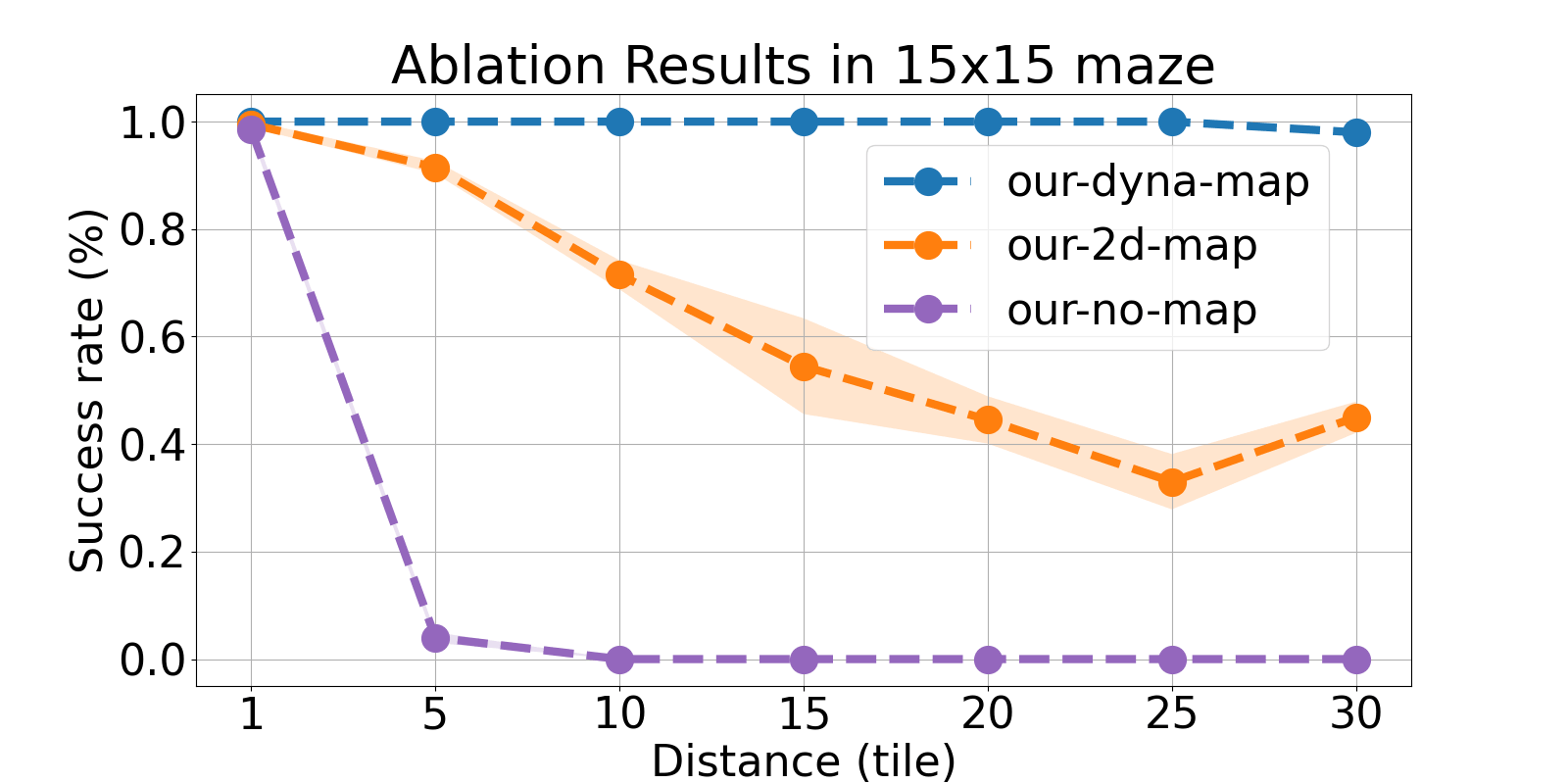}
        \caption{Ablation study}
        \label{fig:ablation_15}
    \end{minipage}
\end{figure}

In this section, we show the results of a more challenging task, where the rough 2-D map only partially captures the correct layout. In order to obtain such partially correct 2-D maps, we randomly sample a proportion of positions in the map and randomly flip their value with probability $0.5$ (i.e., change free to occupied, and vice versa). We test the sampling proportion from $10\%$ to $50\%$. The results in Figure~\ref{fig:res_use_imprecise_map} on $15 \times 15$ seen mazes shows that our framework can handle partially wrong 2-D map to a limited extent. With $10\%$ flipped positions, our framework still achieves $\geq 60\%$ navigation success rate for distance $\leq 15$; however, for longer distances and higher error rates, the incorrect 2-D map causes wrong goal observations to be generated, or all paths are incorrectly assumed to be blocked.
Our framework does not not work well when the map contains $\geq 30\%$ error rates, but as illustrated in Figure \ref{fig:imprecise_map}, the rough 2-D maps in these cases differ significantly from the truth, so we should not expect our method to perform well.

\subsection{Ablation experiments}
\label{sec:ablation_study}
We compare against two ablated versions of our method, demonstrating the importance of both using the rough 2-D map and the dynamic topological map. We run our methods on $15 \times 15$ seen mazes. In Figure~\ref{fig:ablation_15}, our full method (\textbf{our-dyna-map}) achieves robust and consistent navigation success rate across all distances. When the dynamic topological map is not used (i.e., open-loop path-following, \textbf{our-2d-map}), performance drops significantly, although some long-distance navigation tasks are still possible. Finally, when the rough 2-D map is not used and only the local goal-conditioned controller is followed (\textbf{our-no-map}), navigation for distances $\geq 5$ almost always fails. Combined with the previous experiment, we clearly see that both using the rough 2-D map, and keeping track of incorrect information and controller failures is critical.

\subsection{Navigation in seen and unseen mazes with realistic control actions}
\label{sec:new results}
We further test our method on a harder version of Deepmind Lab where the actions are accelerations in corresponding directions. This version has more realistic control actions and contains more partial observability. Figure~\ref{fig:new_results_main} shows that HER performs even worse (i.e., only $40\%$ for map distance $ = 1$) in both scenarios, suggesting that learning a global navigation policy is difficult. However, our method performs much better for short-distance navigation in both scenarios, suggesting learning a reusable local policy is much easier. The performance of our method also drops in both scenarios. Because the local behaviors in continuous space are infinite, learning a good local controller is much more difficult. However, our method still outperforms the baselines in the two scenarios, demonstrating the utility of our method in the domain with realistic control actions. 

In Appendix A, we report additional results including results for the other two maze sizes (i.e. 15x15, 19x19), generalization from smaller to larger mazes, and results for training using more mazes. We also visualize the final topological map, the replanning strategy, as well as the generated panoramic observation to aid understanding.

\section{Conclusion}
\label{sec:conclusion}
In this work, we presented an hierarchical framework for robot navigation in novel environments using rough 2-D maps. The rough 2-D map is used to instantiate a dynamic topological map, which is used to plan a high-level trajectory, and can be further updated to handle imperfect high-level information and low-level controllers. A landmark generator generates observations for high-level states, which is then used as subgoals by a local image-based goal-conditioned controller. By combining robust local navigation with high-level information, our approach achieves superior generalization performance for navigating in novel environments. In the future, we would like to scale up our approach to handle continuous action spaces and realistic visual observations, as well as extend our approach to discover new local behaviors that are deemed impossible by the high-level information.
\clearpage
\acknowledgments{We would like to thank the reviewers for their valuable feedback. We also thank colleagues from GRAIL, LLPR, and HHL labs at Northeastern University for the insightful discussions and suggestions. This research was partially funded by NSF grant 1734497 and an Amazon Research Award.}
\bibliography{chengguang_xu20}  

\newpage
\clearpage
\section{Appendix A: Additional results}
In this section, we provide additional results to aid further understanding of the proposed method. 
\subsection{Planning algorithm}
Algorithm~\ref{alg:plan-with-DTM} shows the complete pseudocode of the  planning algorithm. Given the trained observation generator $p_\theta$ and the local controller $\pi_\psi$, a dynamic topological map $\mathcal{G}$ is first initialized from a 2-D rough map of a novel maze. After randomly sampling a start $s_0$ and goal $s_g$ locations, we set a fixed budge (i.e., $T$ time steps in total) for the navigation. When the navigation starts, our planner will propose the local map patch $m$ of the next landmark. Next, the observation generator will generate the observation subgoal $o$ corresponding to the landmark using $m$. Next, the local controller will use the subgoal $o$ and executes the navigation. When the subgoal navigation terminates in $k$ time steps, the planner will check whether the subgoal is reached based on either environment feedback (i.e., oracle) or agent prediction (i.e., prediction). If it succeeds, then, the next landmark will be proposed; otherwise, the planner will update itself and replan another path. 
\begin{algorithm}[ht]
    \caption{Planning with the dynamic topological map}
    \label{alg:plan-with-DTM}
    \begin{algorithmic}[1]
    \State{Train the conditional VAE: observation generator $p_\theta$}
    \State{Train the goal-conditioned double DQN: local controller $\pi_\psi$}
    \State{$\mathcal{G} \leftarrow$ initGraph($M$)}
    \State{$s_0 \sim \rho_0, s_g \sim \rho_g$} \Comment{Sample start and goal positions in rough 2-D map (or given)}
    \State{$t = 0$} \Comment{Step counter}
    \While {$t < T$ \textbf{or} $s_g$ not reached} \Comment{Navigate with budget $T$}
        \State{$m \leftarrow$ Dijkstra$(s_t, s_g, \mathcal{G})$} \Comment{Find the next landmark $m$}
        \State{$o \leftarrow p_\theta(m)$} \Comment{Generate the corresponding observation $o$}
        \State{$o_{t+k} \leftarrow \pi_\psi(o)$} \Comment{Controller navigates to $o$ in $k$ time steps}
        \If{landmark reached ($o_{t+k}$ corresponds to $m$)} \Comment{Predicted or using oracle}
            \State{$t = t + k, s_t = m.s$} \Comment{Set state to reached landmark (center cell of $m$)}
        \Else
            \State{$o_{t+\hat{k}} \leftarrow \pi_\psi(o_t)$} \Comment{Landmark not reached, return to previous landmark}
            \State{$\mathcal{G} \leftarrow$ Update($\mathcal{G}$)} \Comment{Update the topological map}
            \State{$t = t + k + \hat{k}, s_t = s_{t-k-\hat{k}}$}  \Comment{Set state to previous landmark}
        \EndIf 
    \EndWhile
    \end{algorithmic}
\end{algorithm}

\subsection{Remaining results for maze size 15x15 and 19x19}
\begin{figure}[ht]
\begin{subfigure}{0.5\textwidth}
    \centering
    \includegraphics[scale=0.15]{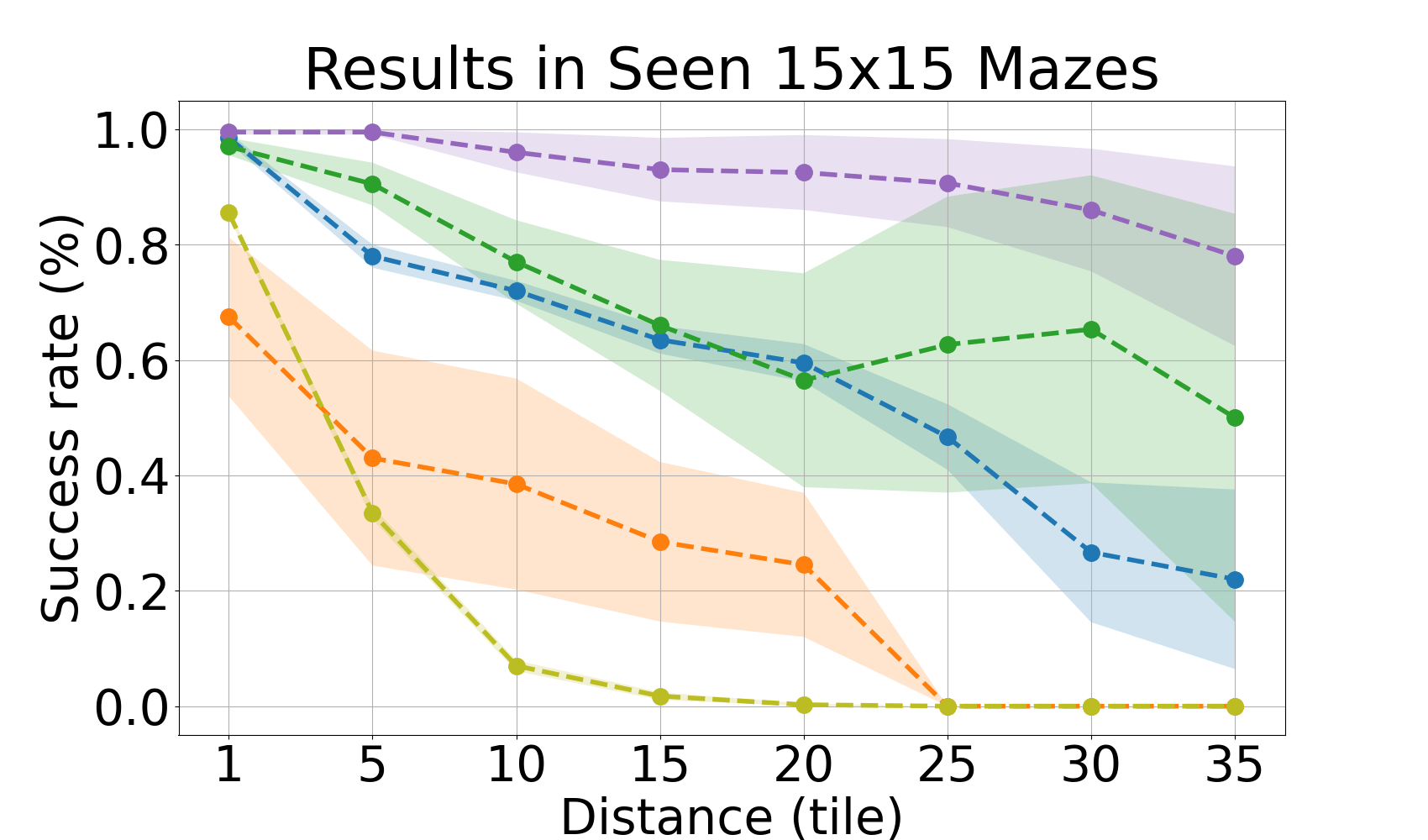}
    \label{fig:scene_1_15_sup}
\end{subfigure}
\begin{subfigure}{0.5\textwidth}
    \centering
    \includegraphics[scale=0.15]{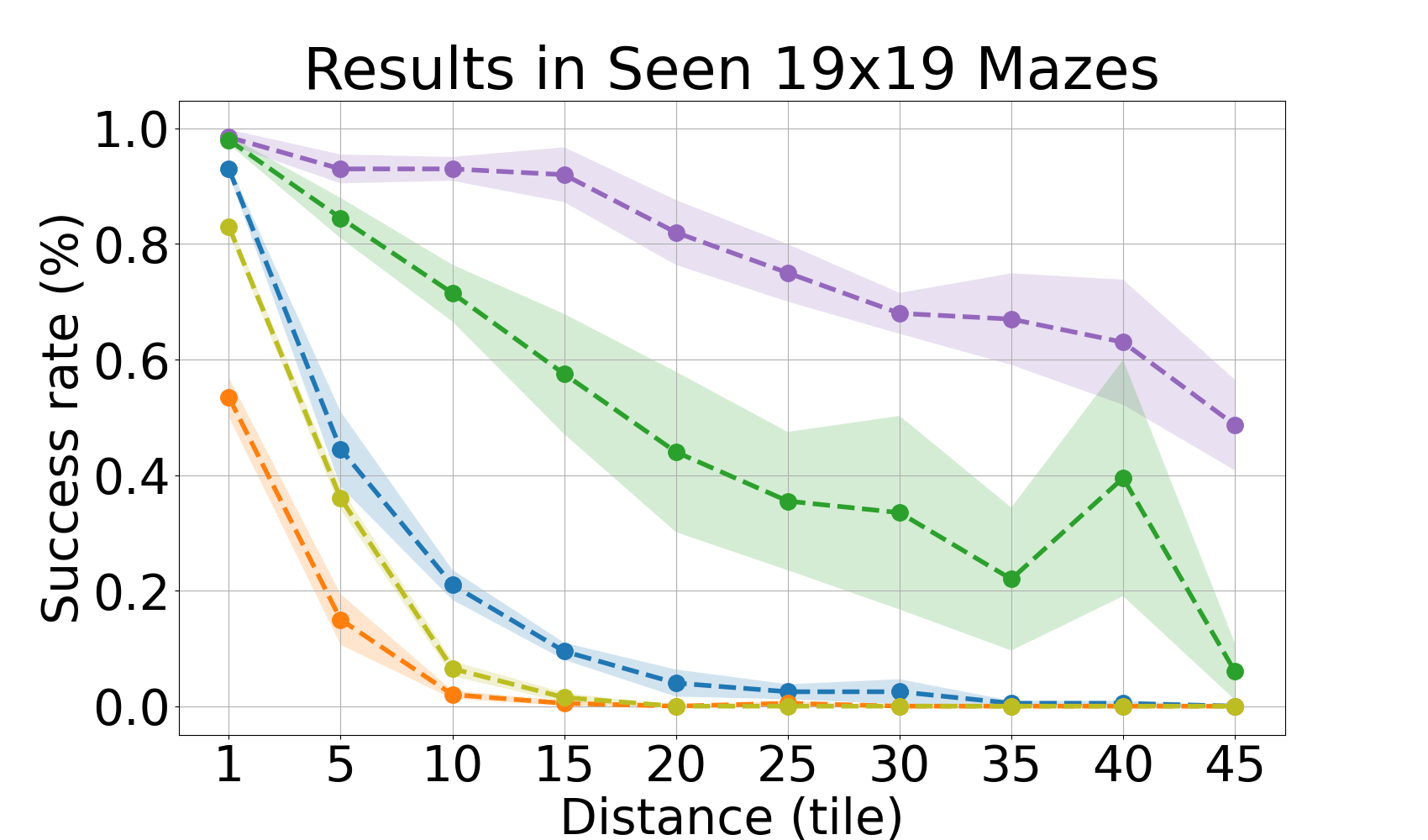}
    \label{fig:scene_1_19_sup}
\end{subfigure}

\begin{subfigure}{0.5\textwidth}
    \centering
    \includegraphics[scale=0.15]{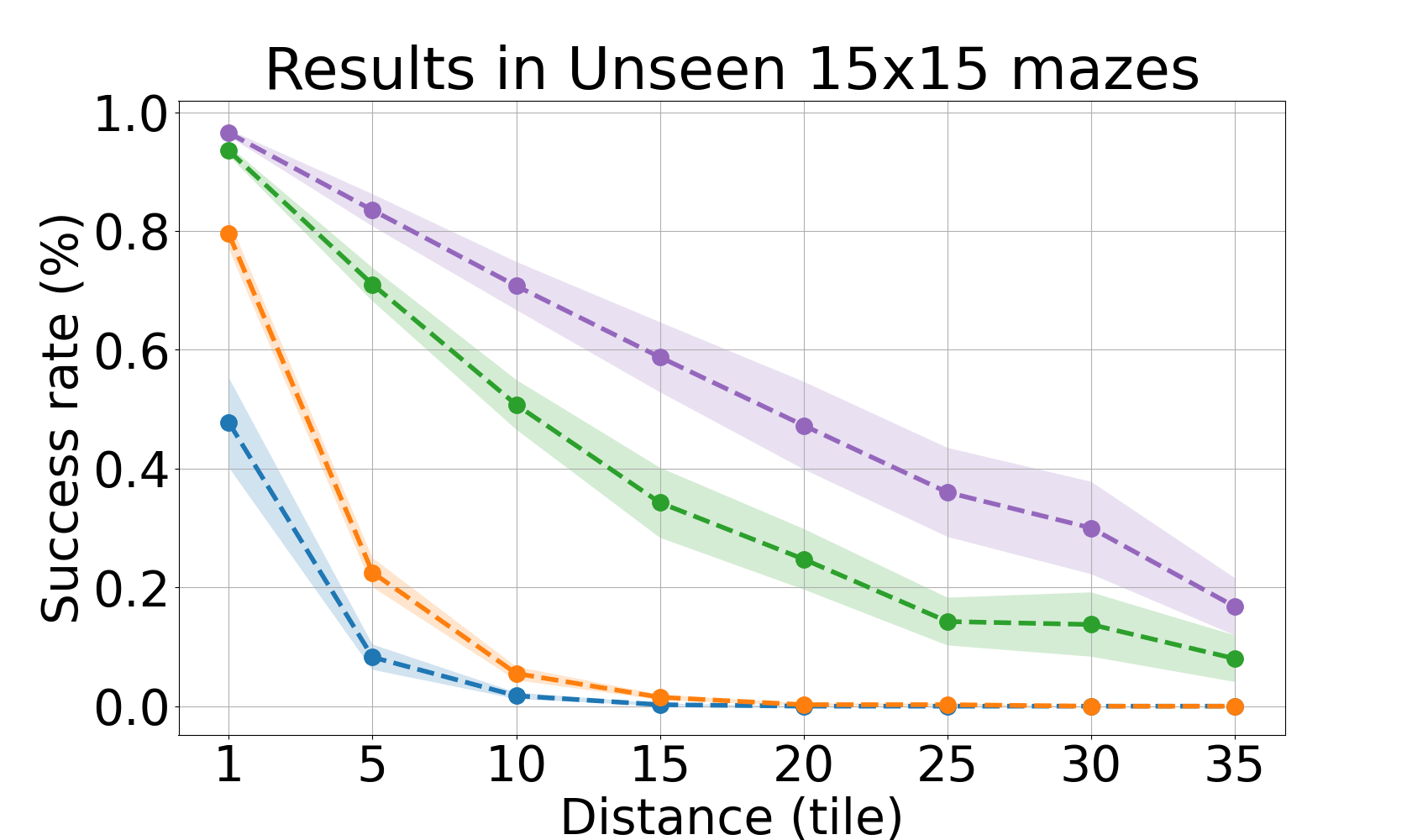}
    \label{fig:scene_2_15_sup}
\end{subfigure}
\begin{subfigure}{0.5\textwidth}
    \centering
    \includegraphics[scale=0.15]{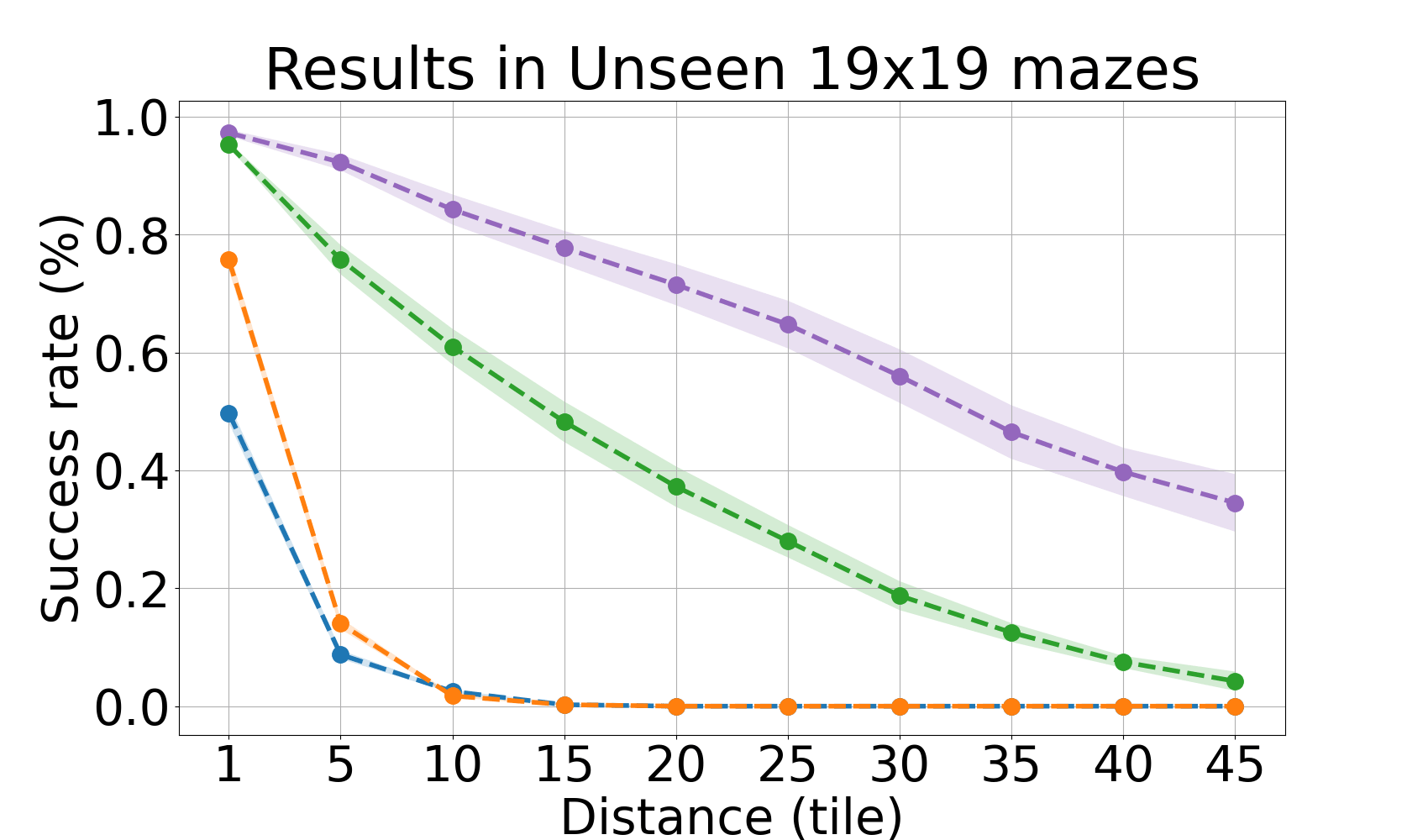}
    \label{fig:scene_2_19_sup}
\end{subfigure}
    \caption{Average success rate for each distance in seen and unseen mazes in the \textbf{discrete} Deepmind Lab. The first row shows the results for seen mazes. The second row shows the results for unseen mazes. Please refer to Figure~\ref{fig:results_main} for legend.}
    \label{fig:results_sup}
\end{figure}

\begin{figure}[ht]
\begin{subfigure}{0.5\textwidth}
    \centering
    \includegraphics[scale=0.15]{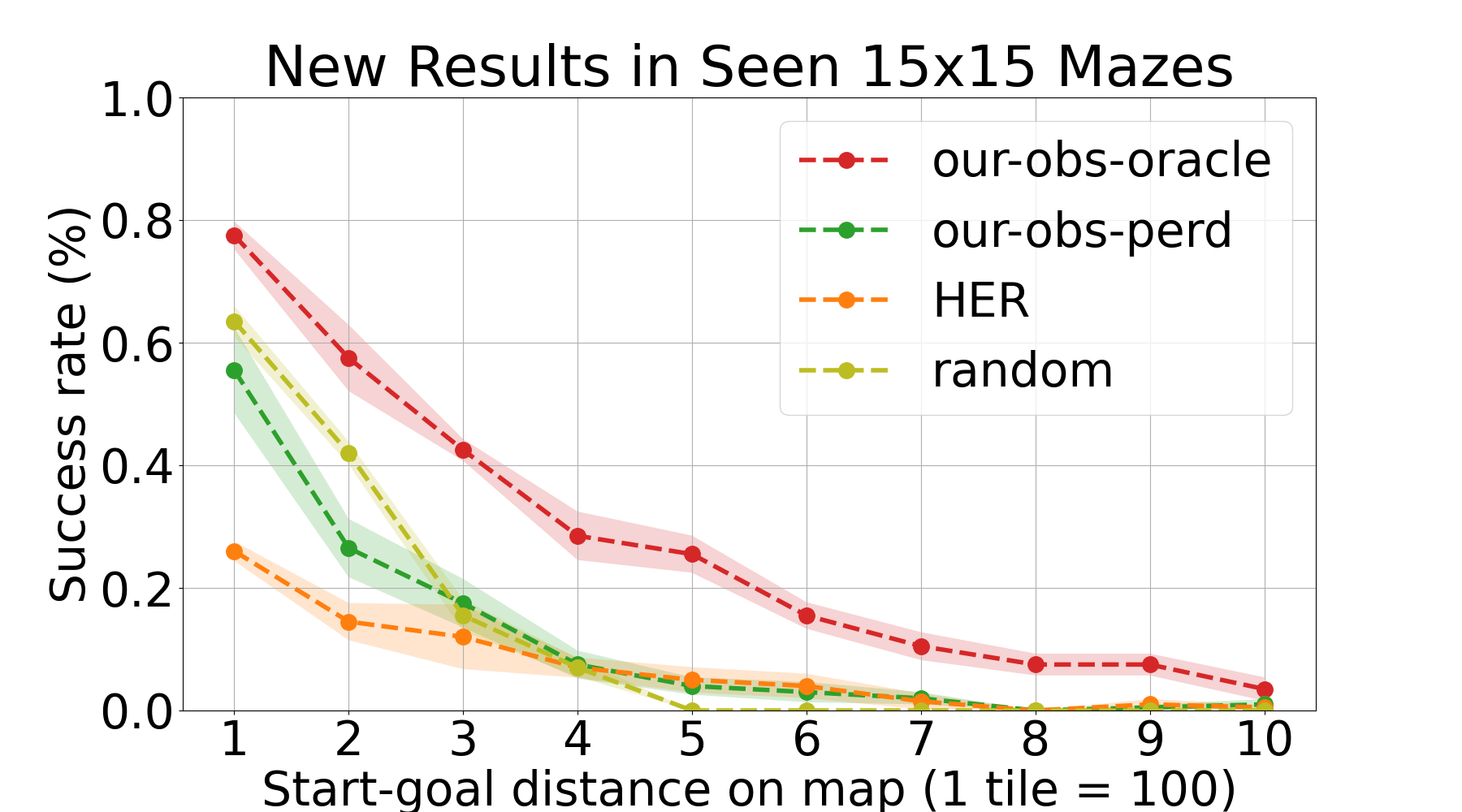}
    \label{fig:new_scene_1_15_sup}
\end{subfigure}
\begin{subfigure}{0.5\textwidth}
    \centering
    \includegraphics[scale=0.15]{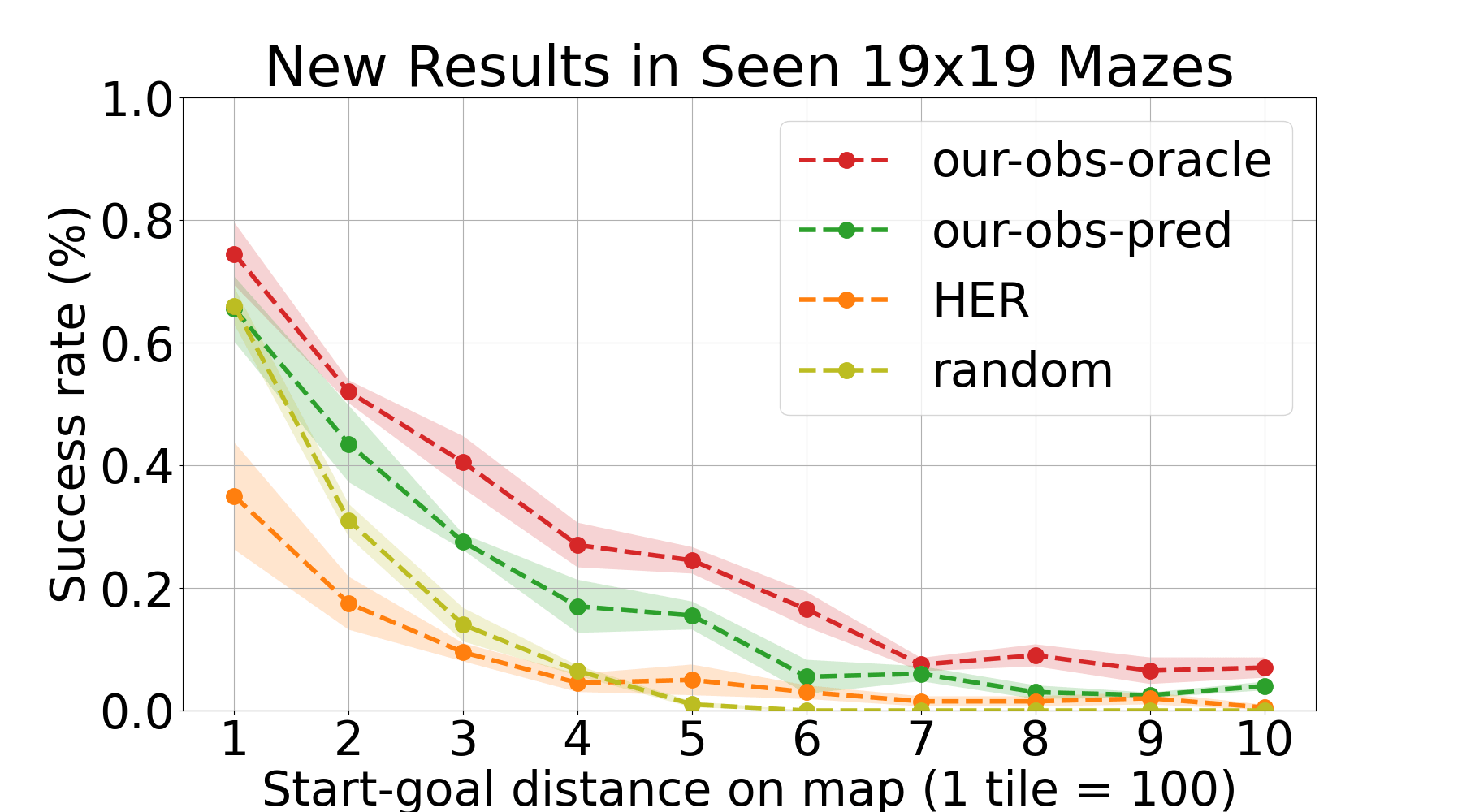}
    \label{fig:new_scene_1_19_sup}
\end{subfigure}

\begin{subfigure}{0.5\textwidth}
    \centering
    \includegraphics[scale=0.15]{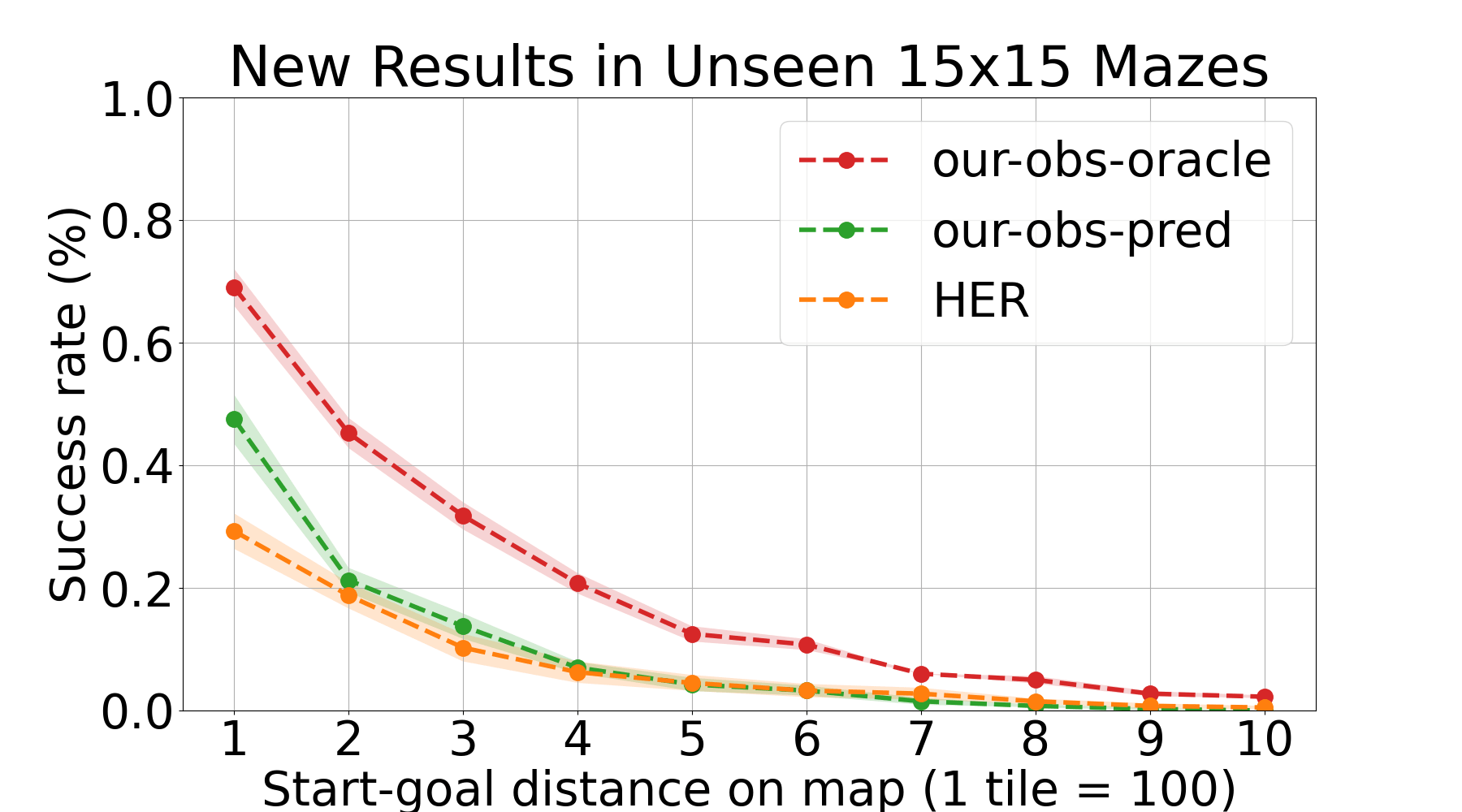}
    \label{fig:new_scene_2_15_sup}
\end{subfigure}
\begin{subfigure}{0.5\textwidth}
    \centering
    \includegraphics[scale=0.15]{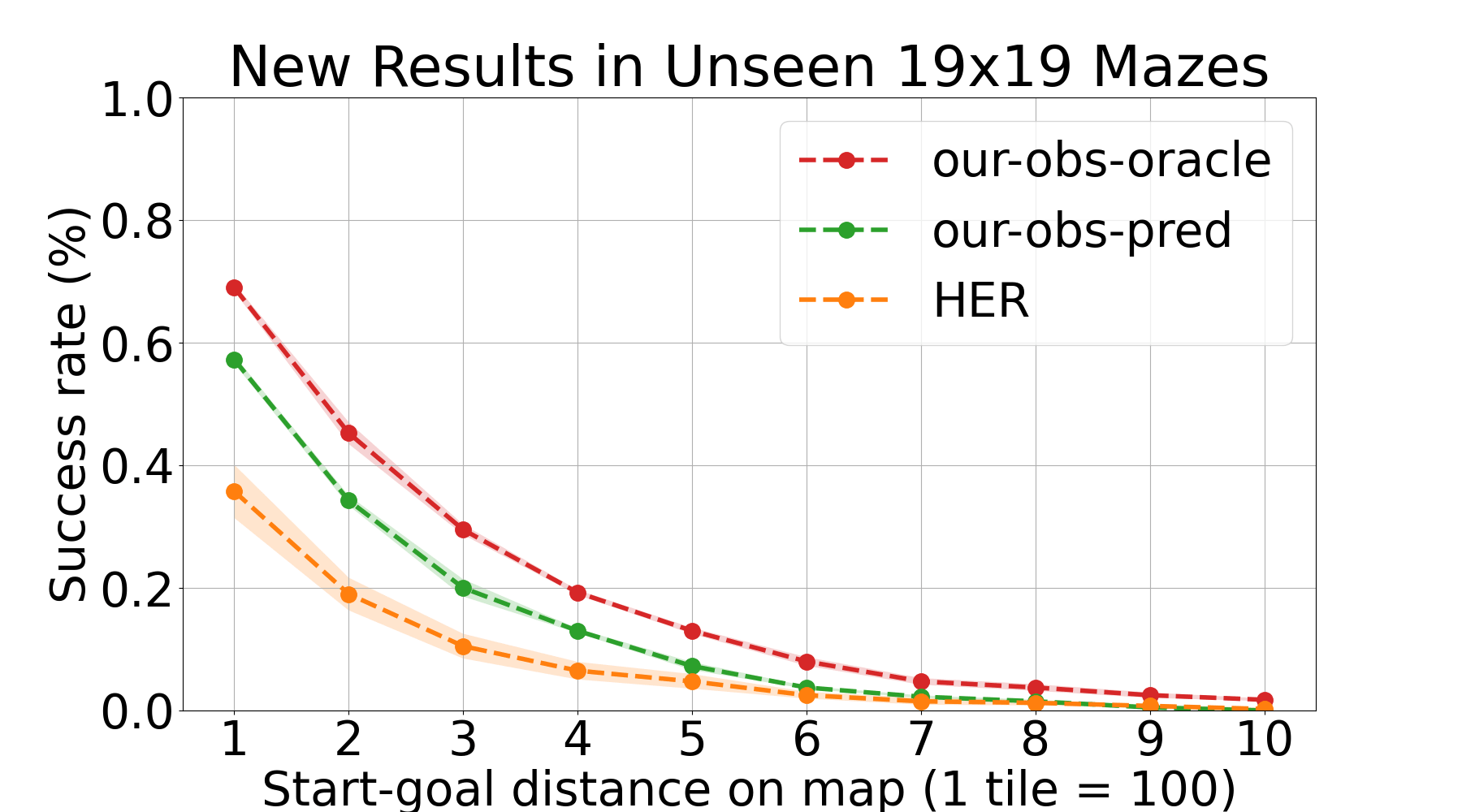}
    \label{fig:new_scene_2_19_sup}
\end{subfigure}
    \caption{Average success rate for each distance in seen and unseen mazes in the \textbf{continuous} Deepmind Lab. The first row shows the results for seen mazes. The second row shows the results for unseen mazes.}
    \label{fig:new_results_sup}
\end{figure}

In this section, Figure~\ref{fig:results_sup} and Figure~\ref{fig:new_results_sup} the results for the remaining two sizes (i.e., 15x15 and 19x19) in the two scenarios.

\subsection{Domain}
\begin{figure}[h]
    \centering
    \includegraphics[scale=0.25]{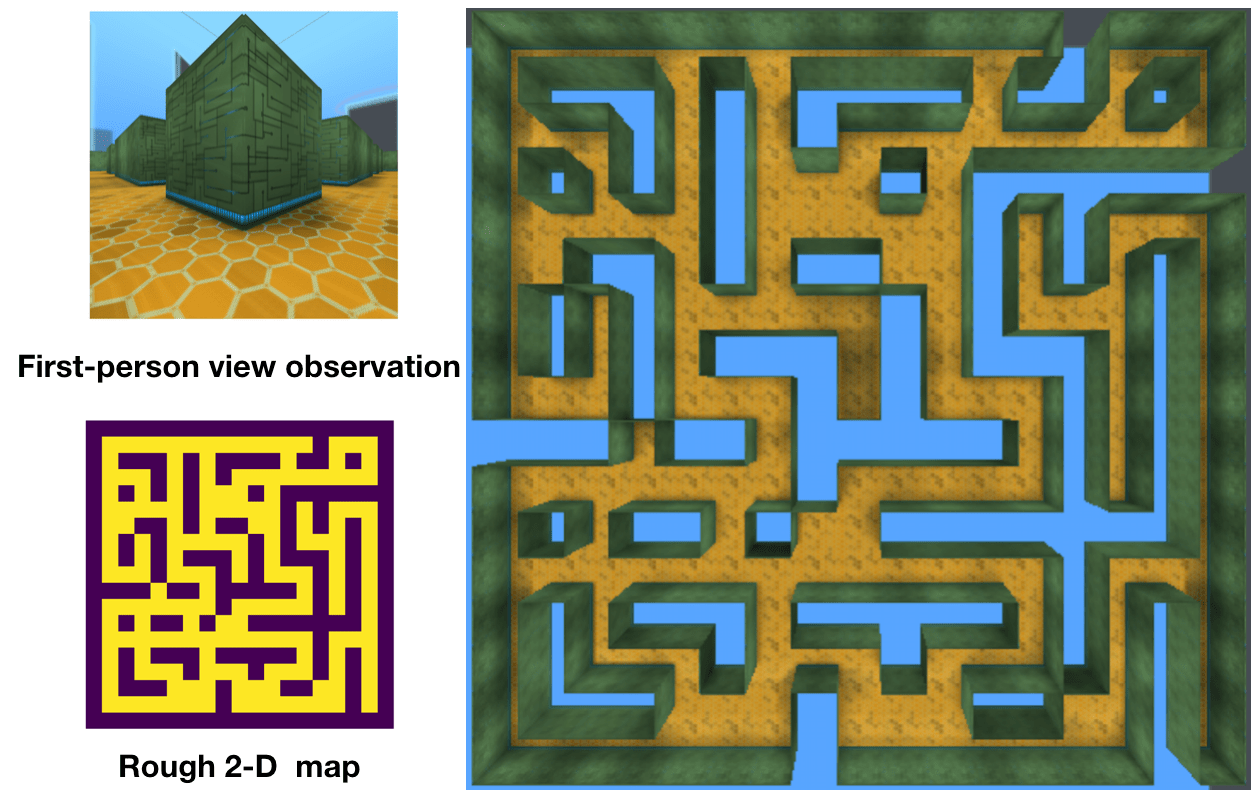}
    \caption{A DeepMind Lab 3D 19x19 Maze. The top left image shows an example of the first-person view observation; 
    The bottom left image shows an example of the rough 2-D map; The image on the right shows a top-down 
    view of the 3D maze, where the yellow part is the corridor area and the green part is the wall area. The blue part is the background. Note that the top-down view on the right is for illustration purposes only. The agent does not have access to it either during training or testing. 
    }
    \label{fig:deepmind_maze}
\end{figure}

We use DeepMind Lab \cite{beattie2016deepmind} as the benchmark. To clarify the data type (i.e. the observation image and the rough 2-D map) and the 3D maze, Figure \ref{fig:deepmind_maze} shows an instance.

\subsection{Generalization to larger mazes}
The generalization to unseen mazes is the primary goal of our method. In section \ref{subsec:generalize_unseen}, we demonstrate that our method can generalize to unseen mazes with the same size. Furthermore, in this section, we show the results of a more challenging generalization task where we train our models in just one maze of size 
15x15 but test on larger mazes.

\begin{figure}[h]
\begin{subfigure}[b]{0.2\textwidth}
    \centering
    \includegraphics[scale=0.25]{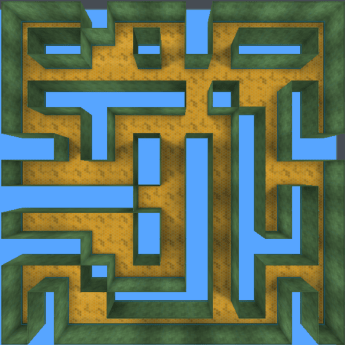}
    \caption{Train: 15x15 maze}
    \label{fig:train_15}
\end{subfigure}
\begin{subfigure}[b]{0.24\textwidth}
    \centering
    \includegraphics[scale=0.3]{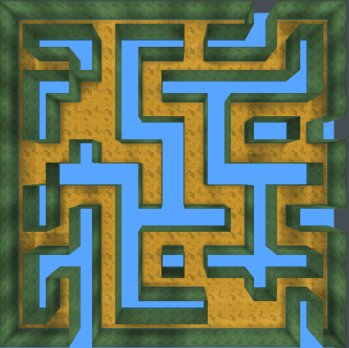}
    \caption{Test: 17x17 maze}
    \label{fig:test_17}
\end{subfigure}
\begin{subfigure}[b]{0.26\textwidth}
    \centering
    \includegraphics[scale=0.35]{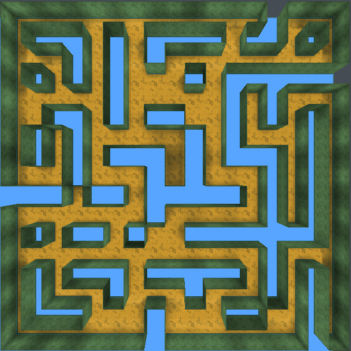}
    \caption{Test: 19x19 maze}
    \label{fig:test_19}
\end{subfigure}
\begin{subfigure}[b]{0.28\textwidth}
    \centering
    \includegraphics[scale=0.4]{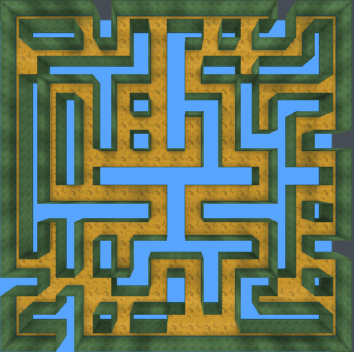}
    \caption{Test: 21x21 maze}
    \label{fig:test_21}
\end{subfigure}
    \caption{Instances of training and testing mazes. (a) shows the only training maze of size 15x15. (b), (c), and (d) show the first maze in the 5 testing mazes of size 17x17, 19x19, and 21x21, respectively. The testing mazes are significantly different from the training maze in both layout and size. The salient difference makes the generalization tasks much harder.}
    \label{fig:generalize_larger}
\end{figure}

We test on 3 larger mazes of size 17x17, 19x19, and 21x21. Figure \ref{fig:generalize_larger} shows the significant difference between the training and testing mazes in both top-down layouts and sizes. 
For each size, we test 5 mazes and report the average 
success rate and the standard error. Table \ref{tab:lager_mazes} shows the results. Note that we report the average success rate for a fixed set of sampled tasks. In other words, we randomly sample 50 tasks for every distance in $\{1, 5, 10, ..., 35\}$, and report the average success rate for every distance. This strategy gives a comprehensive evaluation of random distance navigation performance. Besides, it can also avoid the situation when we use uniform sampling among all distances, where most sampled tasks are short. 

\begin{table}[h]
\centering
\resizebox{12cm}{!}{
\begin{tabular}{*9c}
\toprule
Distance & 1 & 5 & 10  & 15 & 20 & 25 & 30 & 35 \\
\midrule
17x17  & 95.1$\pm$1.2 & 77.3$\pm$2.5 & 61.0$\pm$3.9 & 50.9$\pm$4.7 & 45.0$\pm$5.5 & 38.9$\pm$6.8 & 27.6$\pm$6.2 & 12.0$\pm$5.6\\
19x19  & 94.5$\pm$0.8 & 75.8$\pm$1.9 & 59.3$\pm$3.1 & 48.3$\pm$3.6 & 32.1$\pm$4.7 & 24.4$\pm$4.3 & 19.0$\pm$4.2 & 16.8$\pm$3.9 \\
21x21  & 94.2$\pm$0.7  & 77.1$\pm$2.1 & 60.6$\pm$2.2 & 49.5$\pm$3.4 & 36.3$\pm$3.7 & 32.4$\pm$4.4 & 25.0$\pm$3.1 & 14.6$\pm$2.4\\
\bottomrule
\end{tabular}
}
\vspace{1mm}
\caption{\small Results of generalization to larger mazes
}
\vspace{-2.5mm}
\label{tab:lager_mazes}
\end{table}

Overall
, our method achieves $77.2\%$ success rate for distance $\leq 10$, $43.7\%$ for $10 < $ distance $\leq 20$, and $35.1\%$ for $20 < $ distance $\leq 35$. Our method shows
good generalization performance for 
short distance navigation in larger mazes. However, when the distance increases, the performance decreases due to the existence of some novel local behaviors that are not contained in the maze of size 15x15. In other words, the local goal-conditioned controller learns the local navigation behaviors purely from one 15x15 maze, which might not be
enough for navigation in larger mazes. Although the performance decreases in longer distance navigation, the distance-wise success rates are consistent in different maze sizes, which justifies the stable performance of our method.

\subsection{Training with more mazes}
In previous experiments, we only train the local goal-conditioned controller in one maze. However, intuitively, training with more mazes enables the agent to observe more local behaviors and might commonly improve the performance. Nevertheless, it is also possible that the agent would get confused during the learning because of the learning capacity of the network and the partial observability in 3D mazes. In this section, we discuss whether training the local controller with more mazes will help to improve the performance.

\begin{figure}[ht]
\begin{subfigure}{0.5\textwidth}
    \centering
    \includegraphics[scale=0.18]{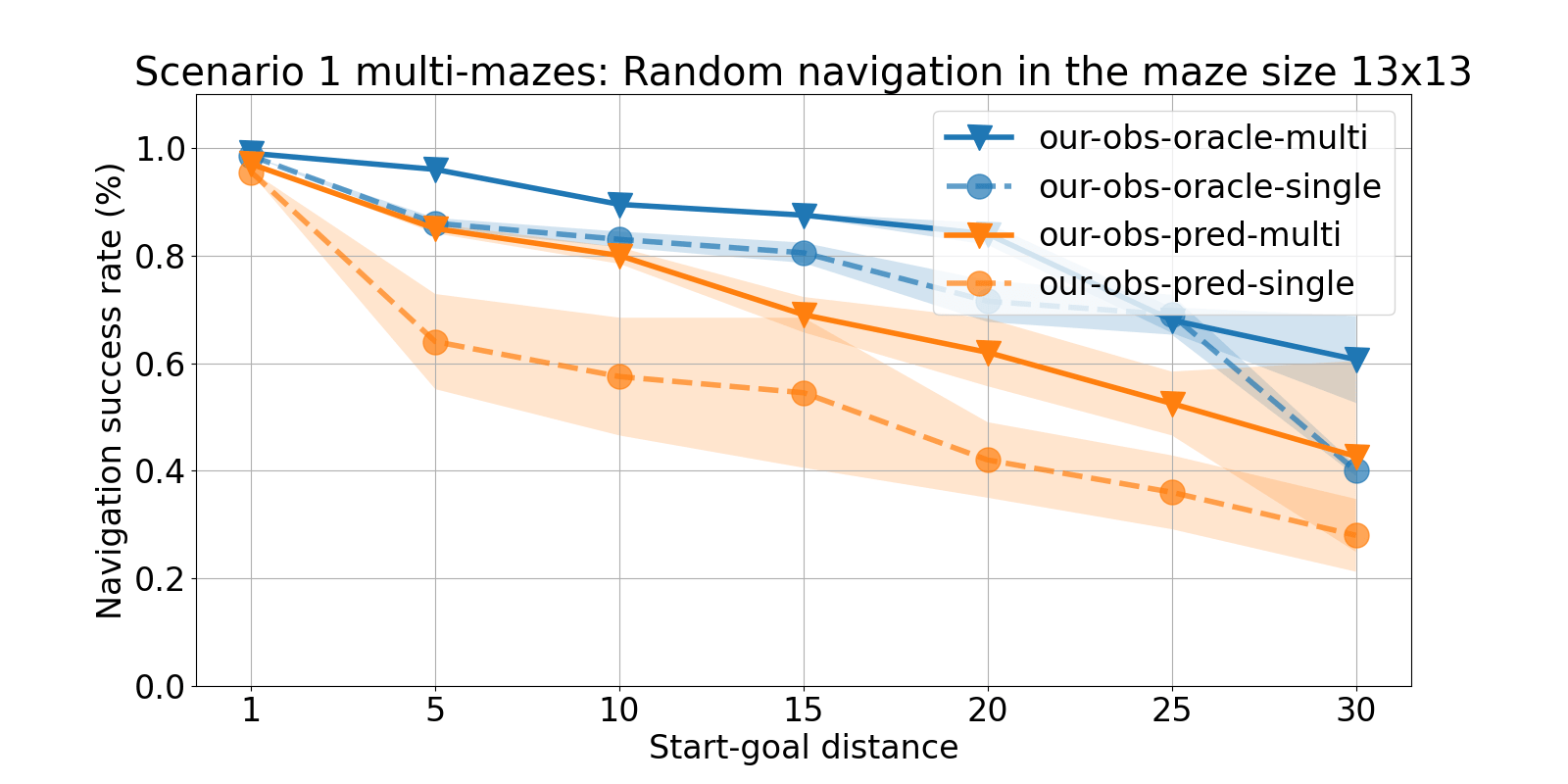}
    \caption{maze 13x13}
    \label{fig:maze_13}
\end{subfigure}
\begin{subfigure}{0.5\textwidth}
    \centering
    \includegraphics[scale=0.18]{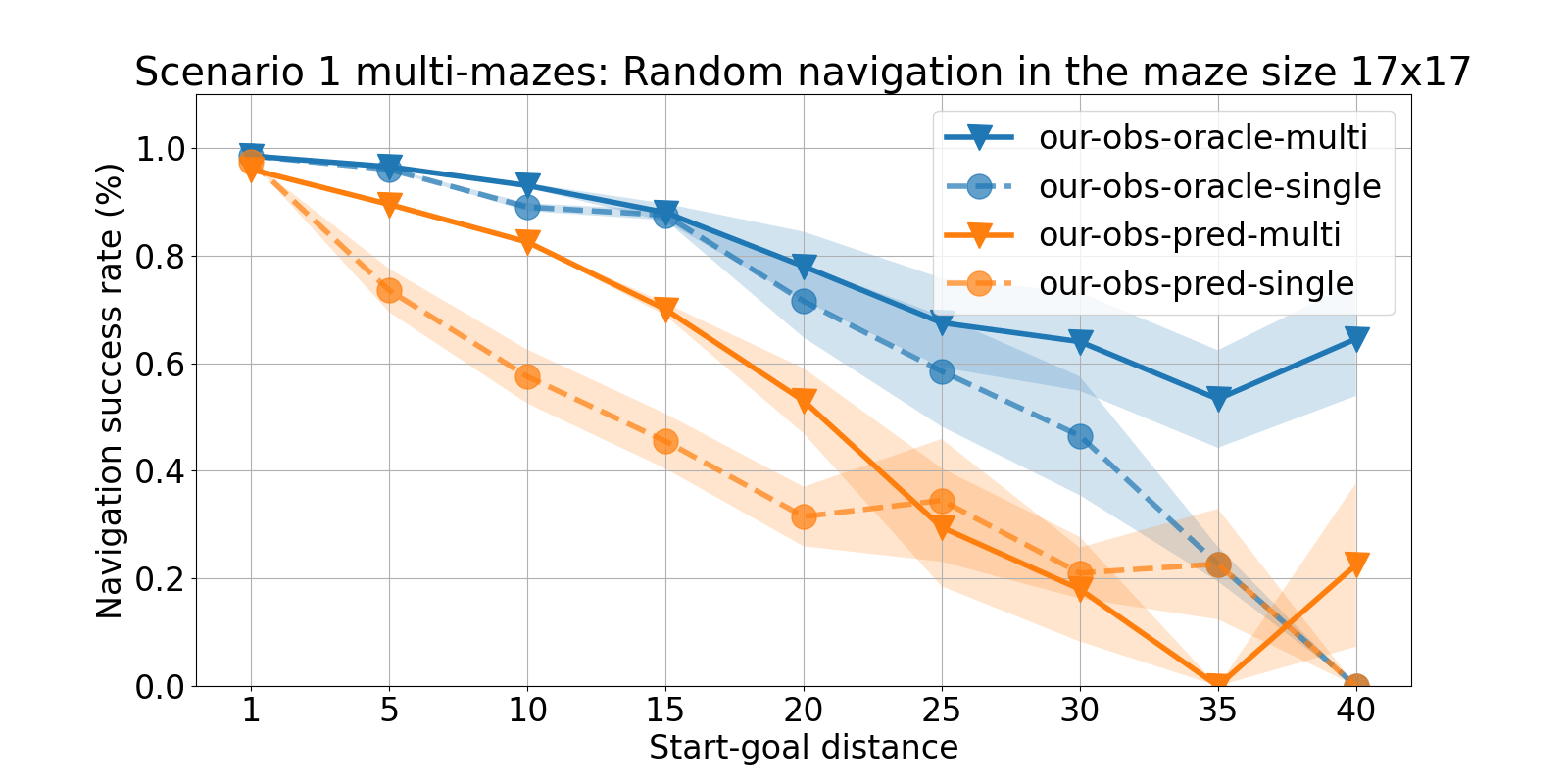}
    \caption{maze 17x17}
    \label{fig:maze_17}
\end{subfigure}
    \caption{Training using multiple mazes vs a single maze.
    }
    \label{fig:train_with_multi_mazes}
\end{figure}

During the training, we randomly select 5 mazes as the training set for each size. For each size, we train the local goal-conditioned controller in the 5 mazes alternately. In particular, we train each maze for 100 episodes before switching to the next maze. Figure \ref{fig:maze_13} shows that performance improves by around $20\%$ for small mazes of size 13x13. However, for larger mazes (e.g. 17x17), the improvement of performance is not obvious for a short distance. But adding more training mazes does 
improve ($30\%$) the performance of longer navigation (i.e. distance $\geq 30$) in Figure \ref{fig:maze_17} in the oracle variant. Overall, adding more mazes for training could be beneficial but the improvement is limited for larger mazes. We hypothesize that the learning capacity of the network and the partial observability might be the main reason. We reserve this to be the future work.

\subsection{Qualitative results of the proposed framework in unseen mazes of 5 sizes}
\begin{figure}[ht]
\begin{subfigure}{0.24\textwidth}
    \centering
    \includegraphics[scale=0.11]{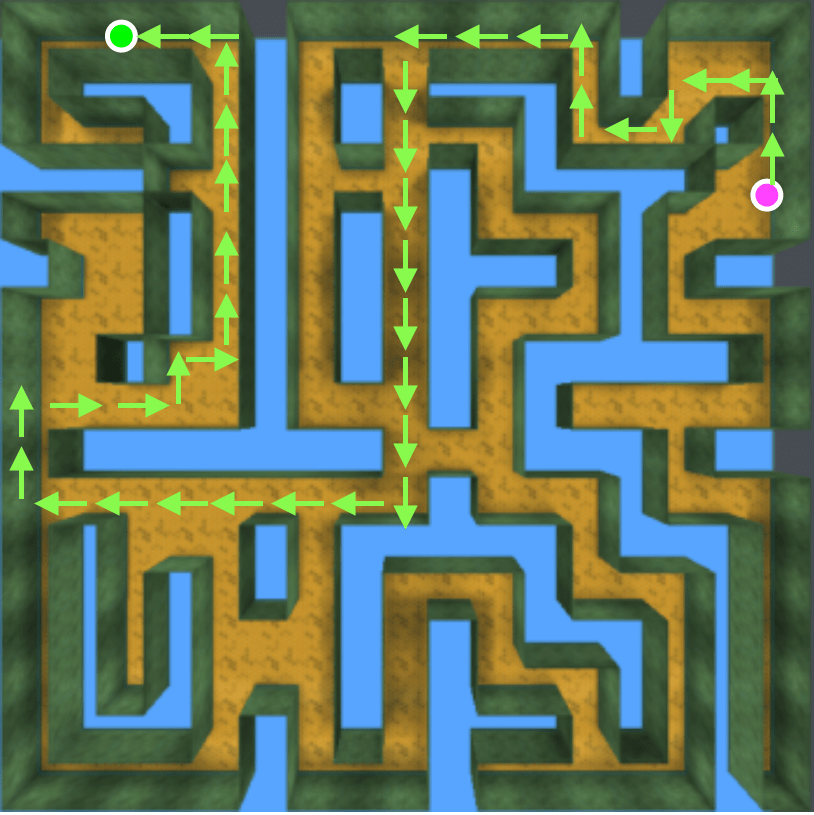}
    \caption{Initial plan}
    \label{fig:init_19}
\end{subfigure}
\begin{subfigure}{0.24\textwidth}
    \centering
    \includegraphics[scale=0.11]{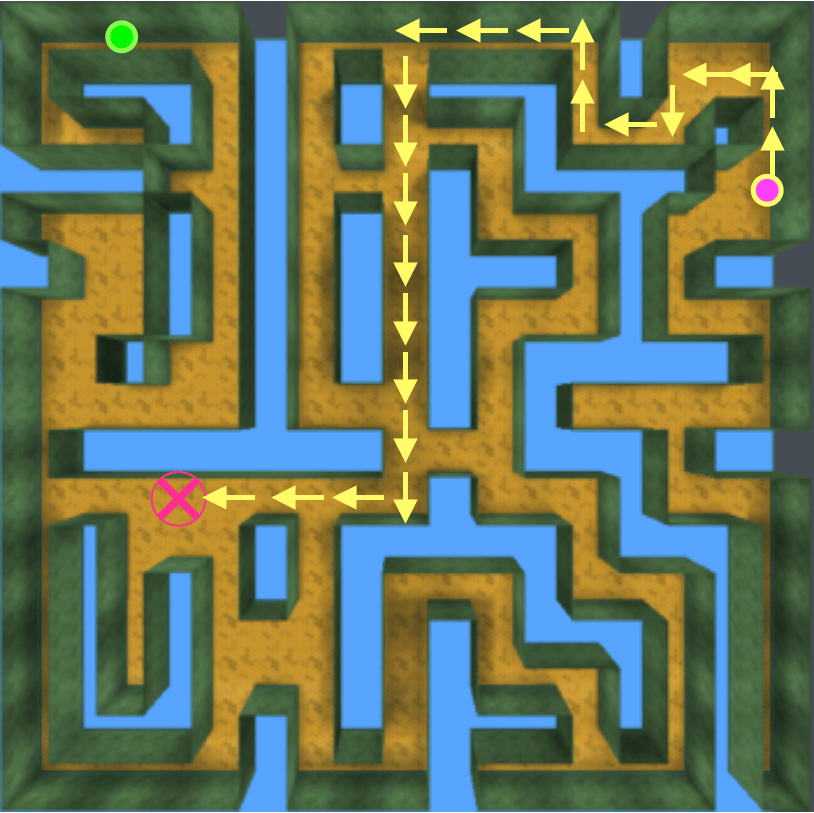}
    \caption{Replan 1}
    \label{fig:replan_1}
\end{subfigure}
\begin{subfigure}{0.24\textwidth}
    \centering
    \includegraphics[scale=0.11]{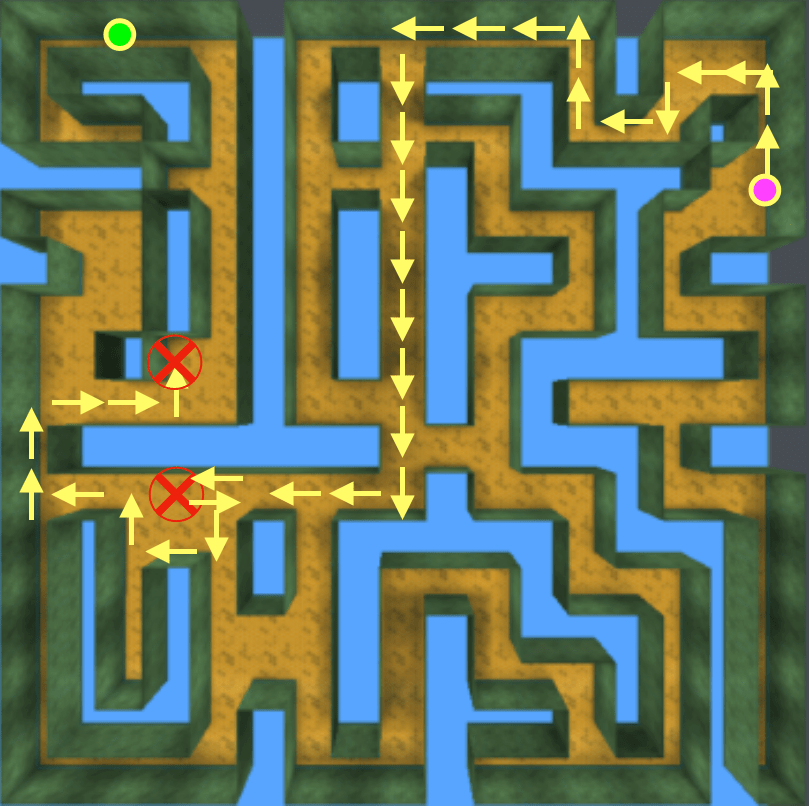}
    \caption{Replan 2}
    \label{fig:replan_2}
\end{subfigure}
\begin{subfigure}{0.24\textwidth}
    \centering
    \includegraphics[scale=0.11]{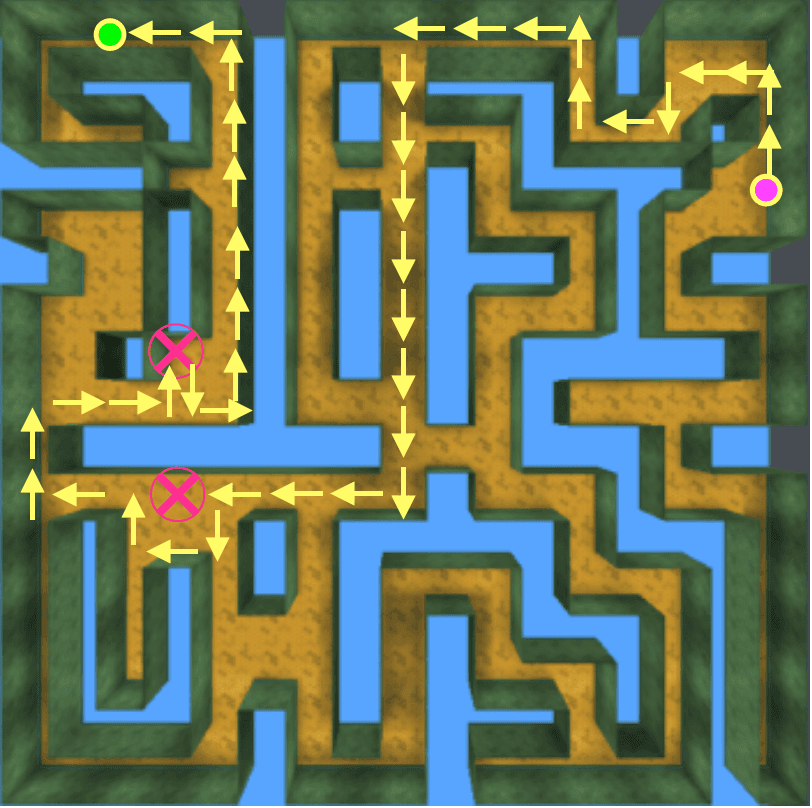}
    \caption{Reach goal}
    \label{fig:reach_goal}
\end{subfigure}
    \caption{Given the start (magenta dot) and goal (green dot) positions, (a) shows the path (i.e. the sequence of green arrows) planned from the initial topological map. Since it is optimistically initialized, the planned path is not guaranteed to be followed. (b) shows the first failure of reaching a landmark (i.e. red cross). (c) shows the dynamic topological map is updated and it replans a new path that avoids the failed landmark. If the failure case happens again, the dynamic topological map keeps updating itself until the robot reaches the goal position in (d). The difference between the actual navigation path (i.e. the sequence of yellow arrows) and the initial navigation path (i.e. the sequence of green arrows) highlights that our framework can 
    adjust the navigation during the online 
    execution. 
    }
    \label{fig:rollout_19}
\end{figure}

\begin{figure}[h]
\begin{subfigure}{0.33\textwidth}
    \centering
    \includegraphics[scale=0.13]{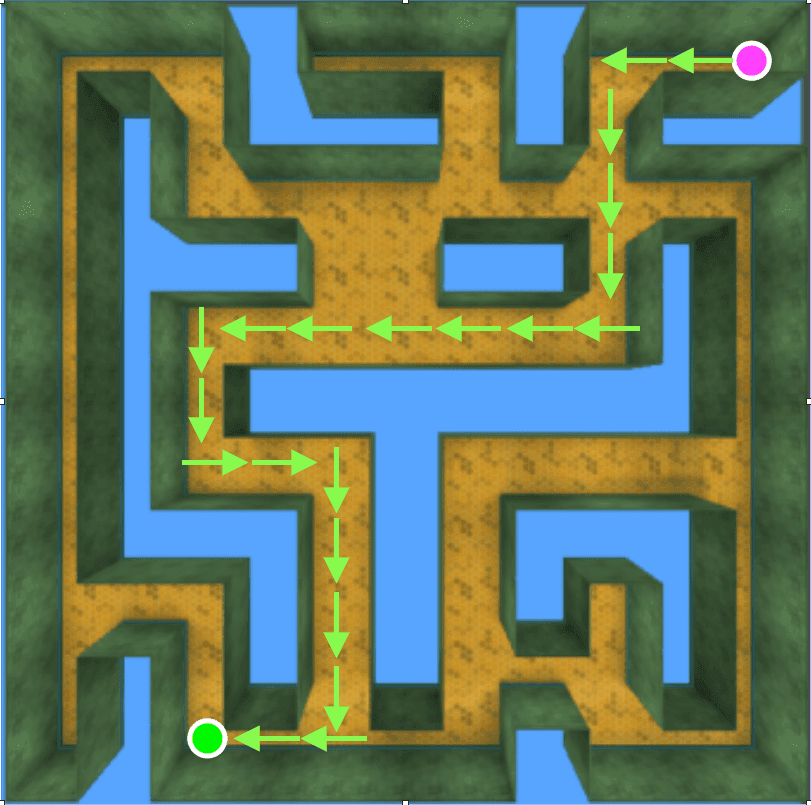}
    \caption{Initial plan}
    \label{fig:init_13}
\end{subfigure}
\begin{subfigure}{0.33\textwidth}
    \centering
    \includegraphics[scale=0.13]{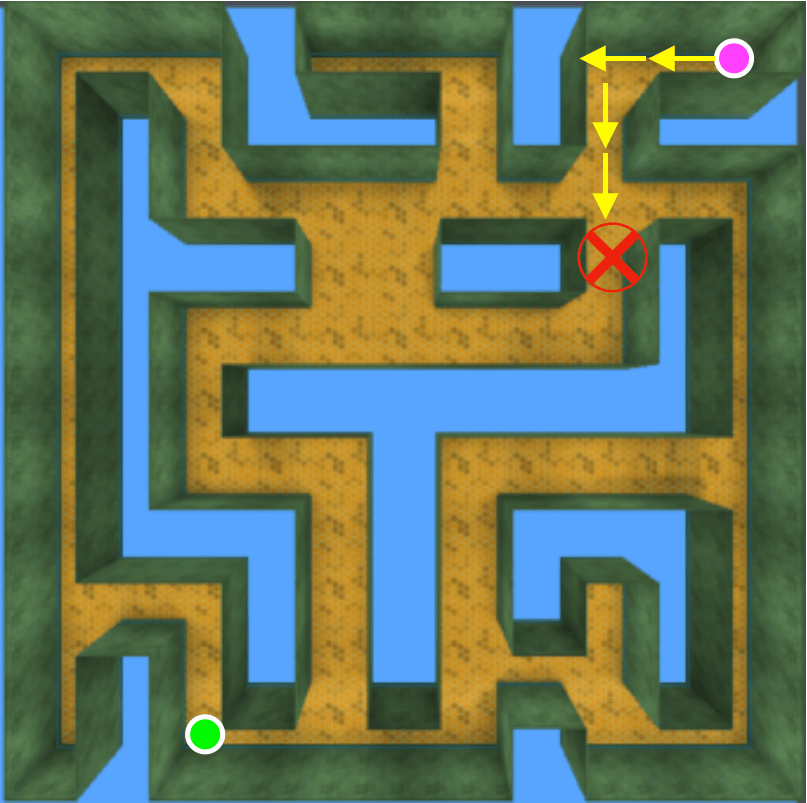}
    \caption{Replan}
    \label{fig:replan_13}
\end{subfigure}
\begin{subfigure}{0.33\textwidth}
    \centering
    \includegraphics[scale=0.13]{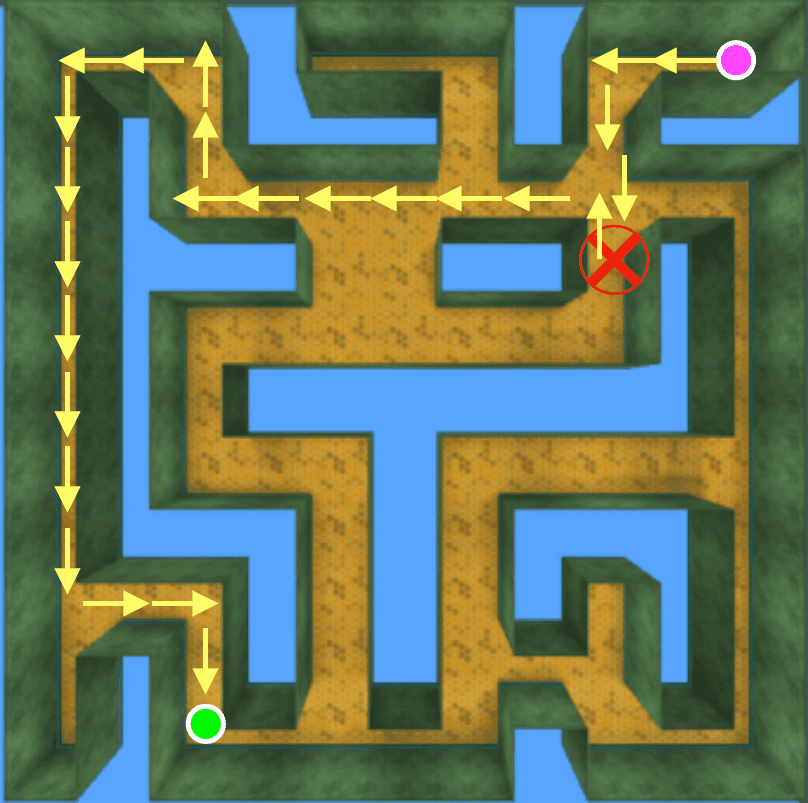}
    \caption{Reach goal}
    \label{fig:reach_goal_13}
\end{subfigure}
    \caption{Visualization in unseen 13x13 maze. Distance between the start and goal positions is 22.}
    \label{fig:rollout_13}
\end{figure}

\begin{figure}[h]
\begin{subfigure}{0.33\textwidth}
    \centering
    \includegraphics[scale=0.13]{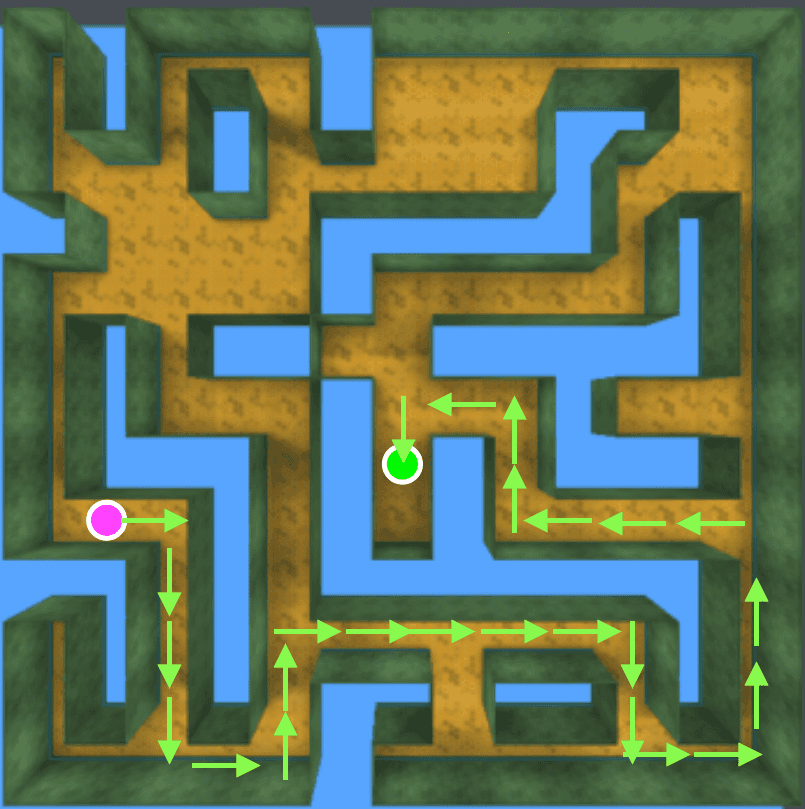}
    \caption{Initial plan}
    \label{fig:init_15}
\end{subfigure}
\begin{subfigure}{0.33\textwidth}
    \centering
    \includegraphics[scale=0.13]{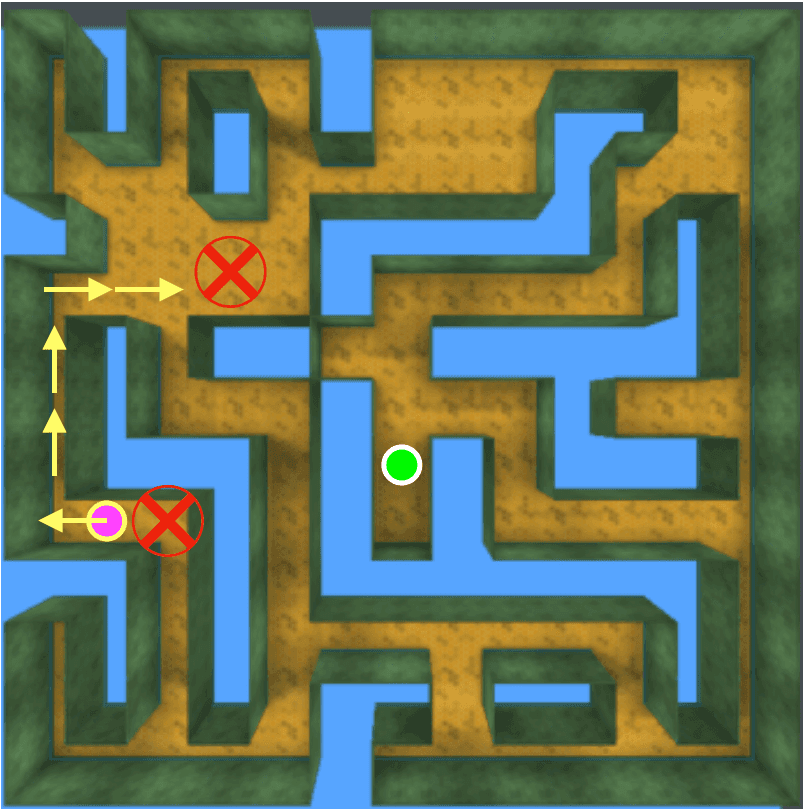}
    \caption{Replan}
    \label{fig:replan_15}
\end{subfigure}
\begin{subfigure}{0.33\textwidth}
    \centering
    \includegraphics[scale=0.13]{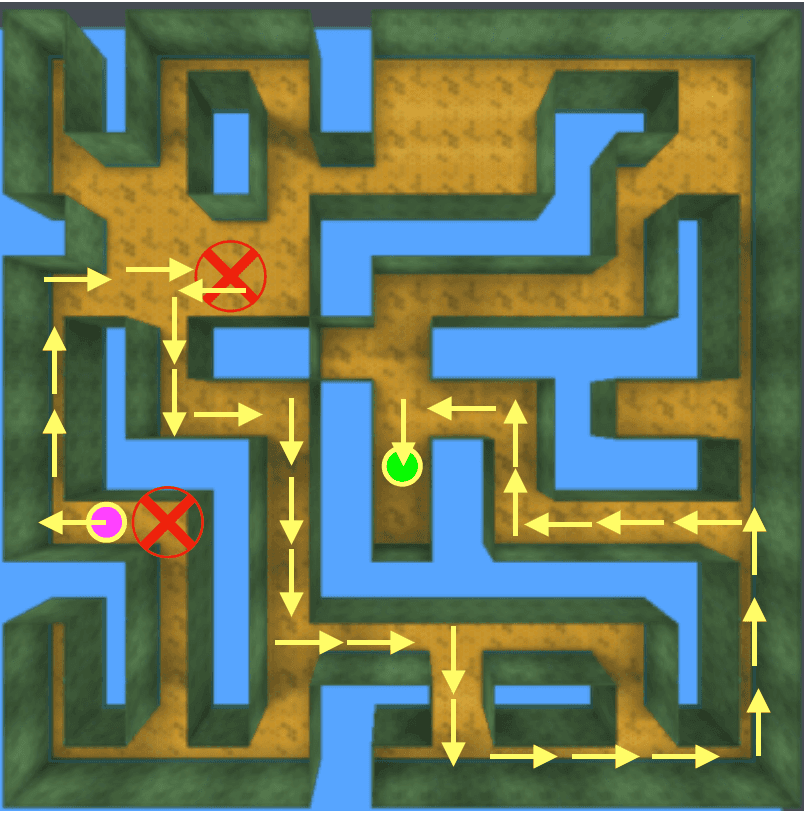}
    \caption{Reach goal}
    \label{fig:reach_goal_15}
\end{subfigure}
    \caption{Visualization in unseen 15x15 maze. Distance between the start and goal positions is 32.}
    \label{fig:rollout_15}
\end{figure}

\begin{figure}[h]
\begin{subfigure}{0.33\textwidth}
    \centering
    \includegraphics[scale=0.13]{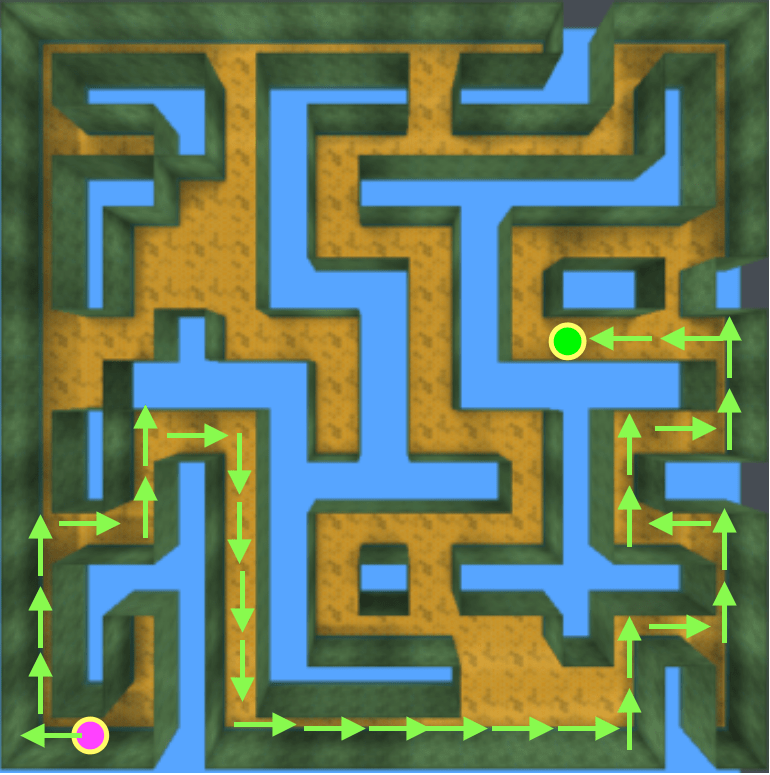}
    \caption{Initial plan}
    \label{fig:init_17}
\end{subfigure}
\begin{subfigure}{0.33\textwidth}
    \centering
    \includegraphics[scale=0.13]{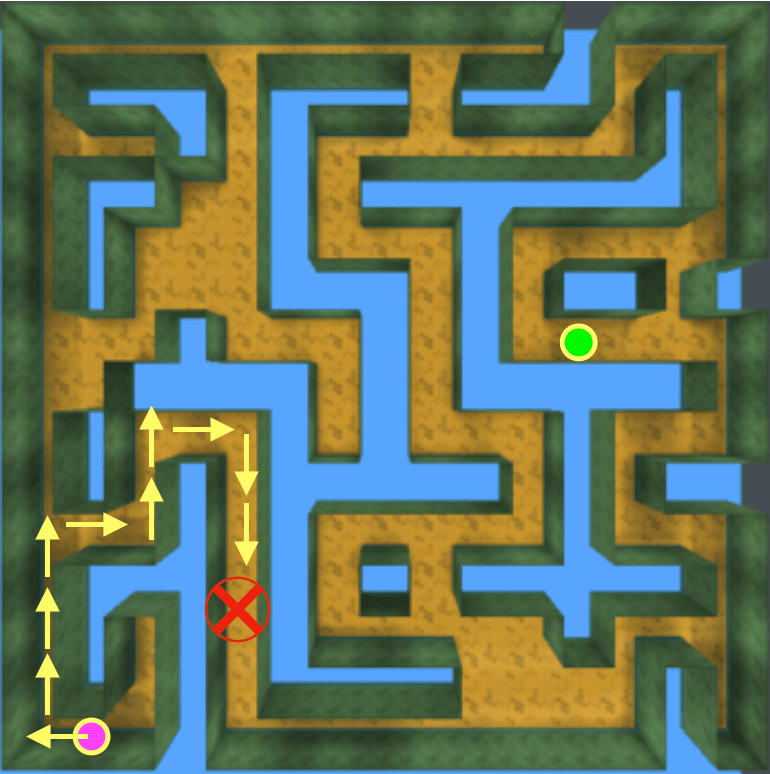}
    \caption{Replan}
    \label{fig:replan_17}
\end{subfigure}
\begin{subfigure}{0.33\textwidth}
    \centering
    \includegraphics[scale=0.13]{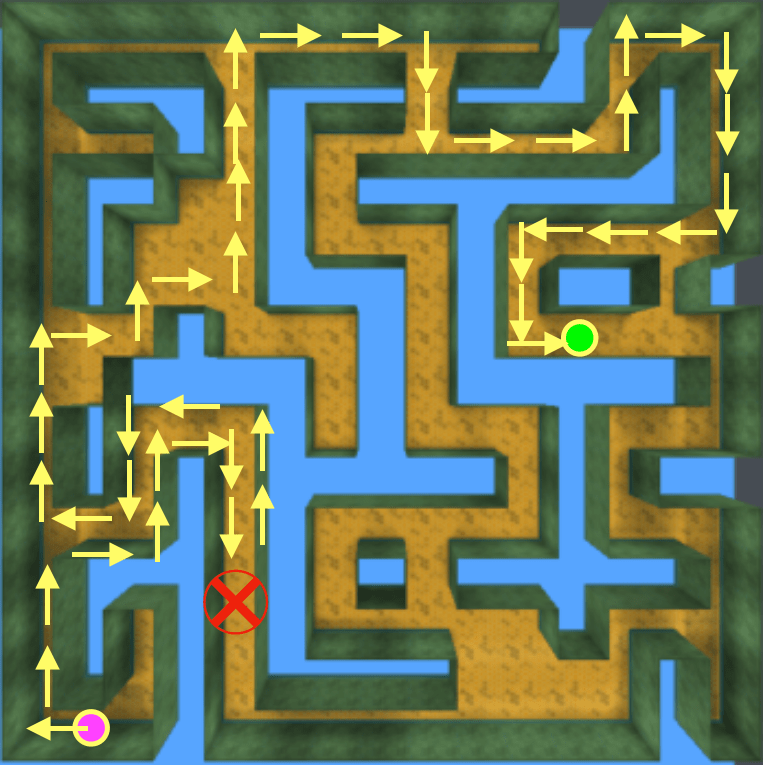}
    \caption{Reach goal}
    \label{fig:reach_goal_17}
\end{subfigure}
    \caption{Visualization in unseen 17x17 maze. Distance between the start and goal positions is 42.}
    \label{fig:rollout_17}
\end{figure}

\begin{figure}[ht]
\begin{subfigure}{0.33\textwidth}
    \centering
    \includegraphics[scale=0.13]{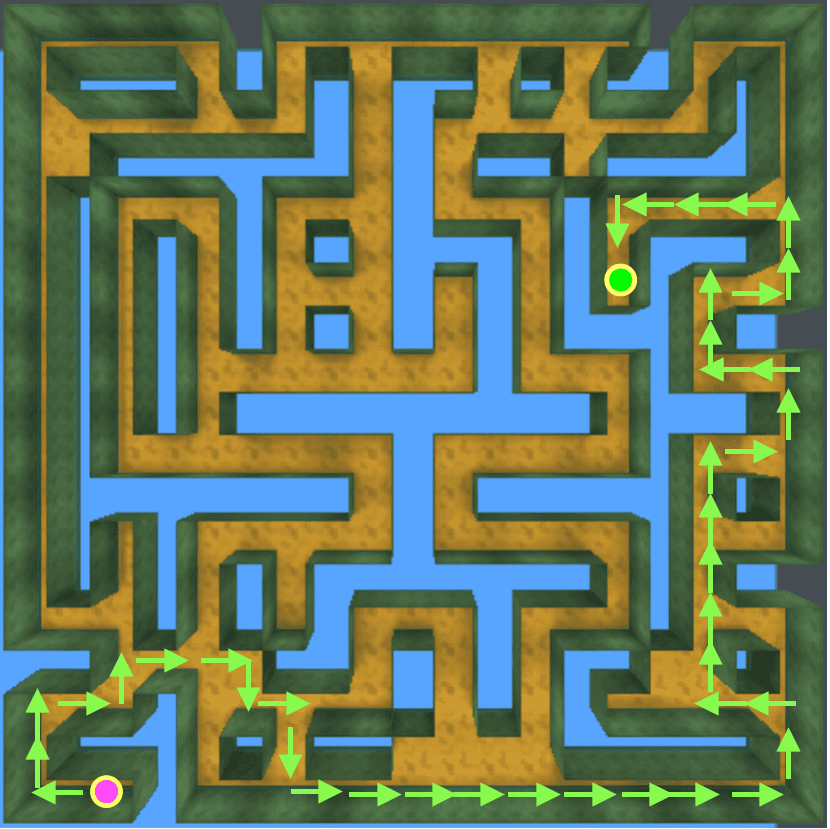}
    \caption{Initial plan}
    \label{fig:init_21}
\end{subfigure}
\begin{subfigure}{0.33\textwidth}
    \centering
    \includegraphics[scale=0.13]{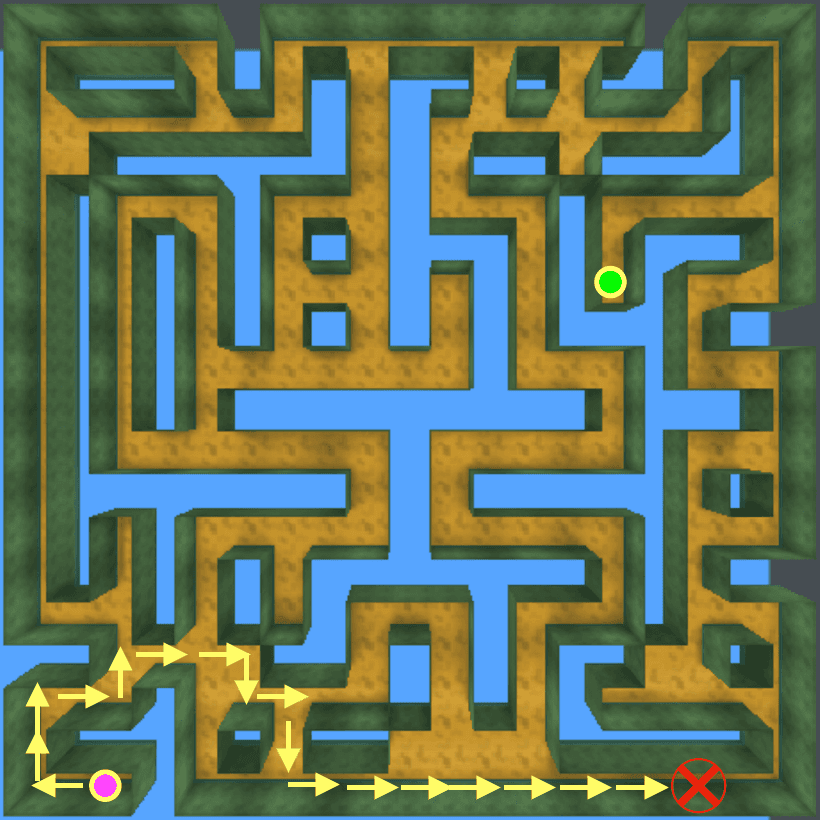}
    \caption{Replan}
    \label{fig:replan_21}
\end{subfigure}
\begin{subfigure}{0.33\textwidth}
    \centering
    \includegraphics[scale=0.13]{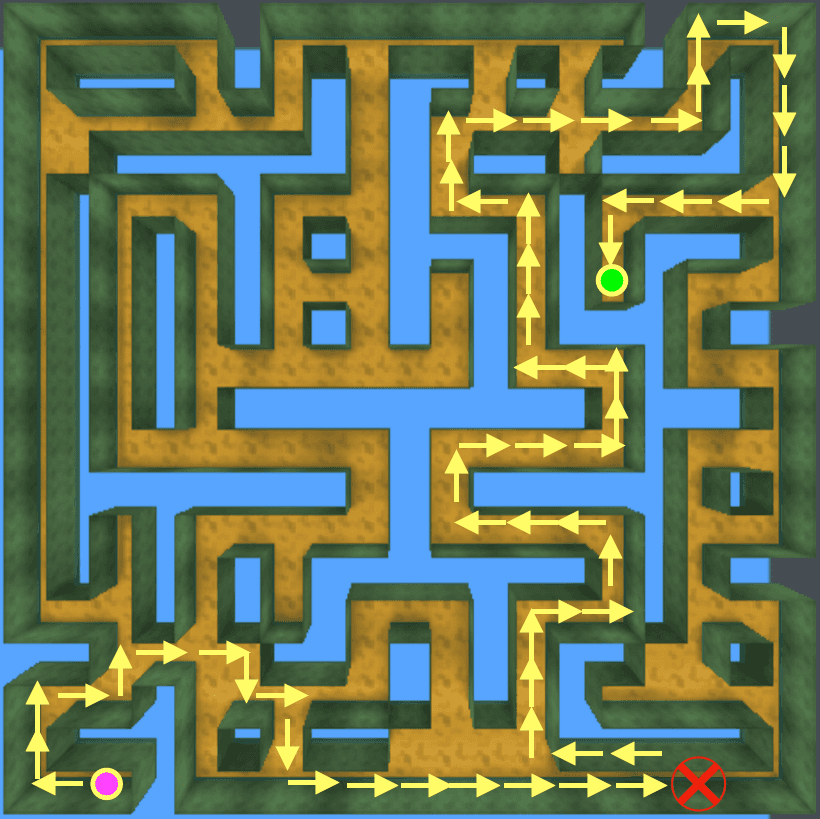}
    \caption{Reach goal}
    \label{fig:reach_goal_21}
\end{subfigure}
    \caption{Visualization in unseen 21x21 maze. Distance between the start and goal positions is 52.}
    \label{fig:rollout_21}
\end{figure}

In this section, we qualitatively show that our method performs robust navigation with the imperfect local goal-conditioned controller. In other words, although some local navigation behaviors are not learned by the controller, the proposed dynamic topological map can update itself and propose reliable landmarks to achieve robust navigation. Figure \ref{fig:rollout_19} gives a detailed example of a real long-distance navigation task in an unseen 19x19 maze, which demonstrates the effectiveness of the proposed framework. More qualitative examples are shown in Figure \ref{fig:rollout_13} to \ref{fig:rollout_21}. 

\begin{figure}[ht]
\begin{subfigure}{0.5\textwidth}
    \centering
    \includegraphics[scale=0.2]{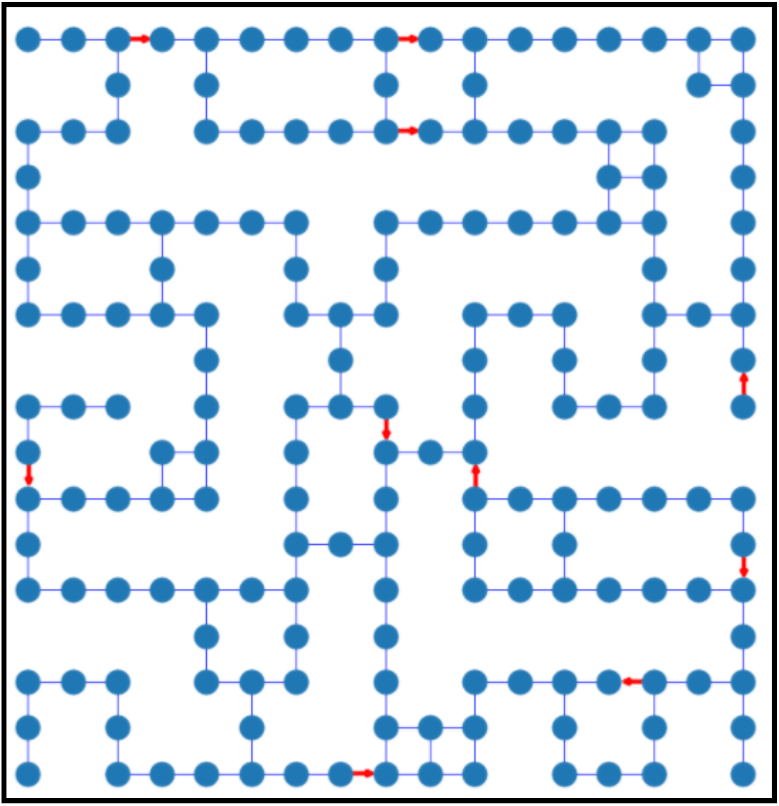}
    \caption{Topological map for ``oracle'' variant}
    \label{fig:DTM_oracle}
\end{subfigure}
\begin{subfigure}{0.5\textwidth}
    \centering
    \includegraphics[scale=0.2]{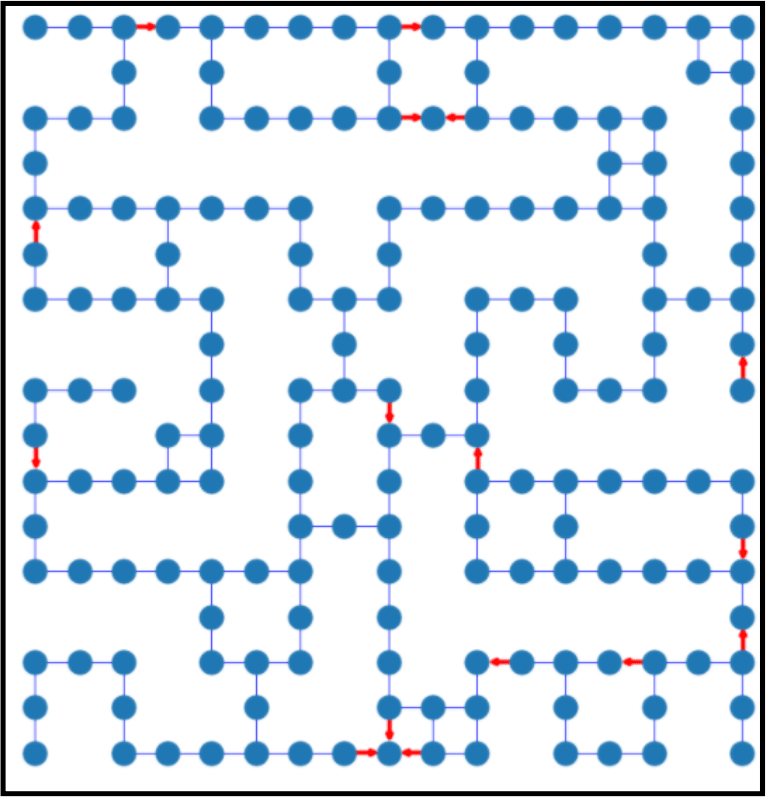}
    \caption{Topological map for ``pred'' variant}
    \label{fig:Pred_pred}
\end{subfigure}
    \caption{Visualization of the topological maps in 19x19 seen maze.}
    \label{fig:dtm_visualization}
\end{figure}

\subsection{Visualization of the dynamic topological map}
Results in Figure \ref{fig:results_main} indicate a variance of performance in different mazes with the same size. Since our method updates the dynamic topological map during the online execution of local behaviors, intuitively, we hypothesize mazes with more learned local behaviors should show better navigation results compared to mazes with less.

To demonstrate it, in this section, we visualize the dynamic topological map by exhaustively looping 
through all the adjacent pairs (i.e. corridors are connected). Note, we connect two nodes with a blue edge if the low-level controller can be used to navigate in any direction. Otherwise, we connect the two nodes with a red arrow whose direction indicates the feasible navigation direction. Figure \ref{fig:dtm_visualization} visualizes 
the topological maps constructed by the ``oracle'' variant and the ``pred'' variant. 

Firstly, it is obvious to see in Figure \ref{fig:dtm_visualization} that some local behaviors are not learned by the low-level controller (i.e. red single-direction edges). Also, the topological map constructed from ``pred'' variant contains more single-direction edges because of the improper prediction of reaching some landmarks. Secondly, from the visualization, we can understand the reason that our method achieves a high navigation success rate ($\geq 90\%$) for short-distance navigation ($\leq 15$) but the lower success rate for extremely long-distance navigation (e.g. $\geq 35$). This is mainly due to the one-direction edges that turn out to be the bottleneck for a long-distance navigation in a particular direction. For example, in Figure \ref{fig:DTM_oracle}, the right arrow on the bottom will only allow the navigation from left to right, but it blocks all paths that try to navigate from right to left if no alternative path exists in the graph.

\clearpage
\section{Appendix B: Details of landmark observation generator}
\begin{figure}[h!]
    \centering
    \includegraphics[height=6cm, width=13cm]{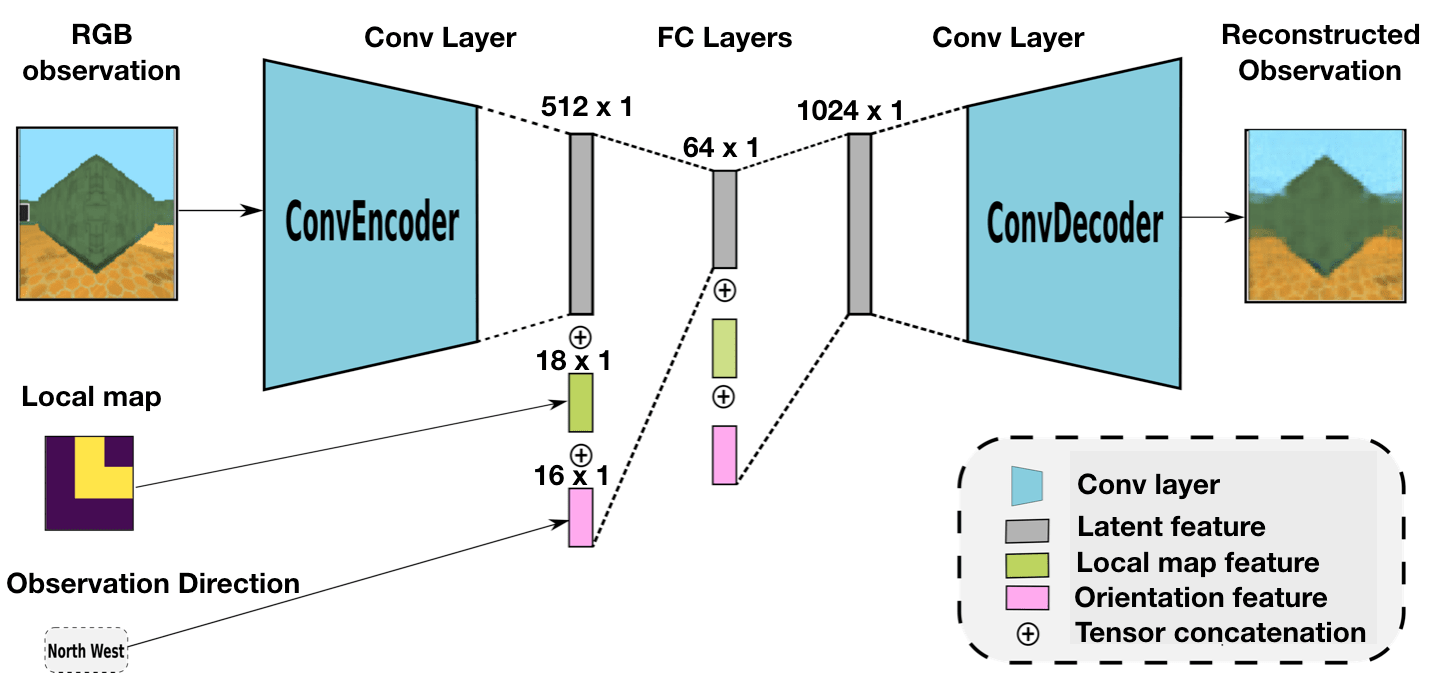}
    \caption{The general architecture of the conditioned VAE. Note, the feature vectors of the local map (9x1) and the direction (8x1) are duplicated in order to enhance to conditional information.}
    \label{fig:cvae_arch}
\end{figure}

In this section, we first explain the data collection for training the landmark observation generator (i.e. CVAE). Then, the details of the architecture are shown in Figure \ref{fig:cvae_arch} and Table \ref{table:cave}.
\subsection{Training data collection and training details} First of all, the random mazes are adopted from \cite{brunner2018MapReader} 
There are 180 mazes in total with 9 different scales. We collect the RGB images (32x32x3) with a random policy in mazes of size $\leq$ 13x13. For each size, we randomly select 10 mazes and collect images from it. We use this data collecting strategy to preserve unseen observations in larger mazes of size $\geq$ 13x13. Each image will be labeled with the corresponding local map and direction. 
Finally, we collect around 22,000 images to train the conditional VAE model. We train the conditioned VAE using stochastic gradient descent (SGD). The optimizer is Adam with a learning rate $1\times10^{-4}$. We train the model with 100 epochs with batch size equals 8. The value of the pixels is normalized between $[0, 1]$. Additionally, the input is the 32x32x3 RGB image, the 9x1 local map feature, and the 8x1 direction feature. The size of the latent representation is 64. The output is a 32x32x3 RGB reconstructed image.  
\subsection{Architecture of the conditioned VAE}

Table \ref{table:cave} below gives the parameters for every layer in the conditioned VAE model. 
\begin{figure}[ht]
\begin{subfigure}{0.5\textwidth}
    \centering
    \includegraphics[scale=0.12]{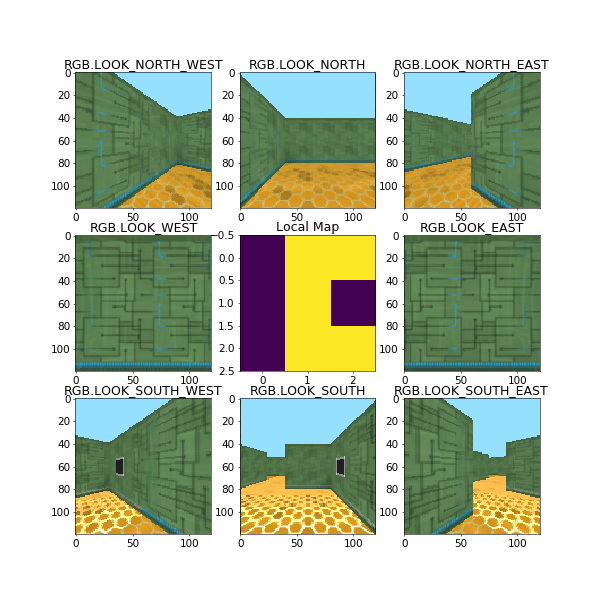}
    \caption{Ground truth observation}
\end{subfigure}
\begin{subfigure}{0.5\textwidth}
    \centering
    \includegraphics[scale=0.12]{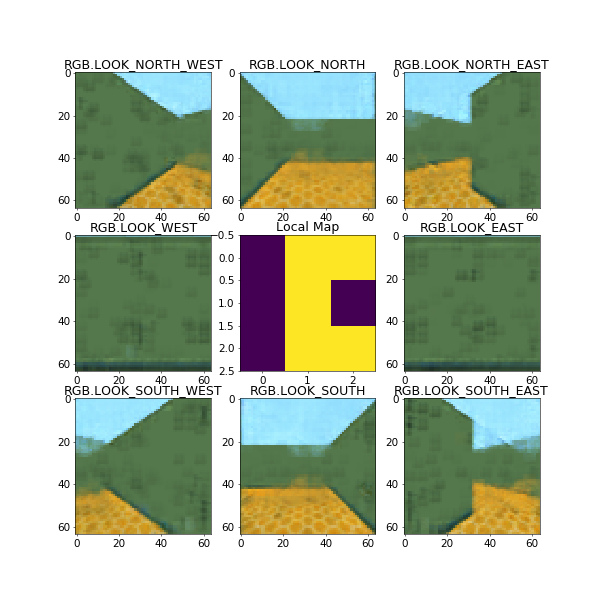}
    \caption{Generated observation}
\end{subfigure}
\begin{subfigure}{0.5\textwidth}
    \centering
    \includegraphics[scale=0.12]{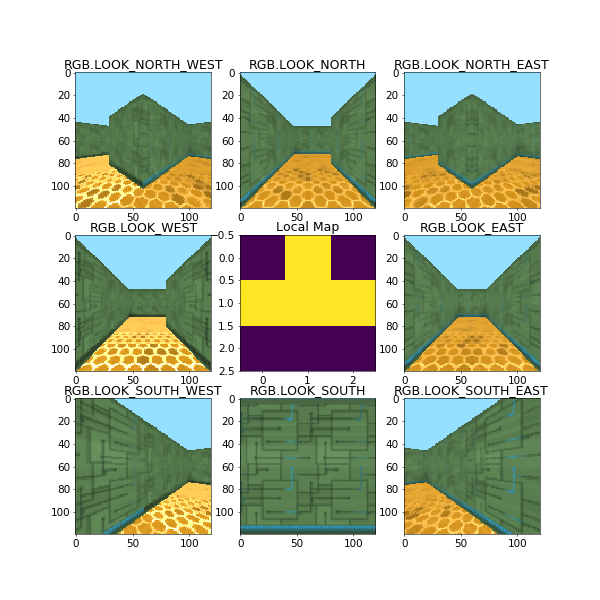}
    \caption{Ground truth observation}
\end{subfigure}
\begin{subfigure}{0.5\textwidth}
    \centering
    \includegraphics[scale=0.12]{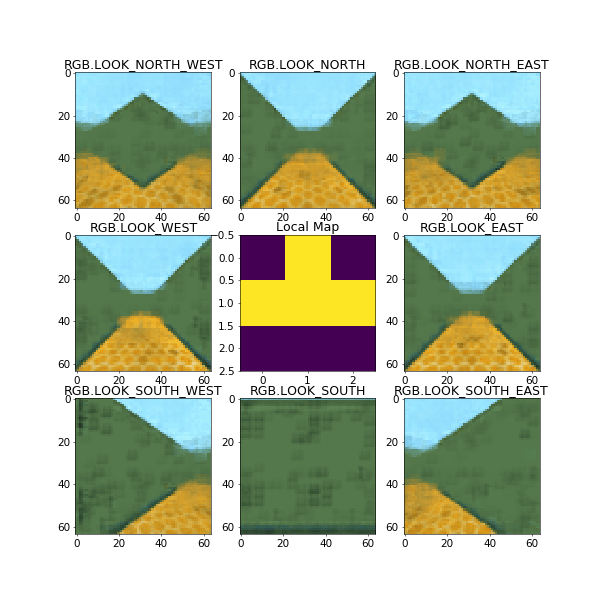}
    \caption{Generated observation}
\end{subfigure}
    \caption{Comparison between the ground truth panoramic observation with the generated observations. The images on the left column are the ground truth observations corresponding to one particular position represented by the local map in the center. The images on the right column is the panoramic observation generated using the same local map and 8 cardinal and ordinal directions. Note, the image in one direction is independently generated from the local map and the related the direction. The results demonstrate that the generator learns to interpret the local map and can infer the observation in one direction from it. }
    \label{fig:compare_gt_vs_gen}
\end{figure}
\begin{table}[h!]
\centering
 \caption{Parameters for each layer in the conditioned VAE. The upper part is the convolutional encoder. The middle part is the fully connected layers (i.e. latent representation). The bottom part is the deconvolutional decoder. Note that $F, P, S$ are abbreviations for the filter, padding, and stride, respectively, in the fourth column.}
 \label{table:cave}
\begin{tabular}{|p{2.4cm}||p{2.4cm}|p{2.4cm}|p{2.4cm}|p{2.4cm}| }
 \hline
 \multicolumn{5}{|c|}{Conditional VAE Architecture} \\
 \hline
 Layer Name & Input Dim & Output Dim & F. / P. / S. & Activation Fn\\
 \hline
 Conv1    & 3 x 32 x 32 & 32 x 28 x 28 & 5 / 0 / 1 & None\\
 MaxPool1 & 32 x 28 x 28 & 32 x 14 x 14 & 2 / 0 / 2 &ReLU\\
 Conv2 &32 x 14 x 14 & 64 x 12 x 12 &3 / 0 / 1 &None\\
 MaxPool2 &64 x 12 x 12 &64 x 6 x 6 &2 / 0 / 2 &ReLU\\
 Conv3 & 64 x 6 x 6 &128 x 4 x 4 & 3 / 0 / 1 &None\\
 MaxPool3 &128 x 4 x 4 & 128 x 2 x 2 & 2 / 0 / 2 &ReLU\\
 \hline
 FC1 & 512 + 18 + 16  & 64 &None &None\\
 FC2 & 64 + 18 + 16  & 1 x 1 x 1024 &None &None\\
 \hline
 DeConv1    & 1 x 1 x 1024 & 128 x 5 x 5 & 5 / 0 / 1 &ReLU\\
 DeConv2 &128 x 5 x 5 & 64 x 13 x 13 & 5 / 0 / 2 &ReLU\\
 DeConv3 & 64 x 13 x 13 &3 x 32 x 32 & 8 / 0 / 2&Sigmoid\\
 \hline
 \end{tabular}
\end{table}

\section{Appendix C: Qualitative results for landmark observation generation}
The observation generator is an essential component of our framework. Because it is used to generate the observations for future landmarks in unseen mazes. In this section, Figure \ref{fig:compare_gt_vs_gen} shows qualitative results to demonstrate two advantages of the generator. 1) It can generate realistic images compared with the ground truth; 2) Given the same local map, it shows a good direction-conditioned performance. In this fashion, by combining the images from 8 cardinal and ordinal directions, we can easily construct a panoramic observation for a particular landmark, which can be used as a sub-goal in the low-level local goal-conditioned controller. 
\clearpage

\section{Appendix D: Details of local goal-conditioned controller}
In this section, we first explain the details of training the local goal-conditioned controller (i.e. goal-conditioned DQN \cite{schaul2015UvFA}). Then, Figure \ref{fig:local_controller} shows the architecture of the two variants used in our experiments.
\subsection{Training details}
Generally, we train the local goal-conditioned controller abiding by the general training paradigm of DQN \cite{mnih2015DQN}. First of all, we use the panoramic observation (i.e. 8 images in cardinal and ordinal directions) as our input rather than a single image to attenuate the issue of partial observability because we don't have recurrent layers in our model. The inputs to our model are the true panoramic observation of the current position and the panoramic observation of the landmark. We train the model for 1M time steps in total. We update the policy network every 10 time steps and update the target network using ``soft'' update with $\tau = 0.05$. The memory size is 100K. Since our model is goal-conditioned, one transition in the memory buffer is $(o, a, o', o_{g}, d)$ where $o, a, o', o_g, d$ 
are the current true panoramic observation, action, next true panoramic observation, landmark panoramic observation, and done flag, respectively. We discount the reward with $\gamma = 0.99$. We train the model using SGD with a batch size of 128. The optimizer is Adam with a learning rate $1\times10^{-4}$.

\subsection{Relabeling the goal observation}
Since the landmark observation is generated rather than received from the environment during navigation, we will relabel the true landmark observation with the generated observation to enhance the utility of the generated observations. For example, during the training, we collect a transition tuple $(o, a, o', o_{g}, d)$ through interaction with the environment. With a particular probability, we relabel $o_g$ with the generated goal observation $\hat{o_g}$ and add the tuple $(o, a, o', \hat{o_{g}}, d)$ to the replay buffer. We test 4 different relabeling proportions $0\%, 25\%, 50\%$, and $100\%$. Empirical results show that $50\%$ performs the best. Our hypothesis is those true observations bring more visual variance compared with the corresponding generated observations, which is good to promote the robustness of the learned deep model. $50\%$ balances the learning to use the generated observations and learning a robust deep model.
\begin{figure}[h]
    \centering
    \includegraphics[scale=0.3]{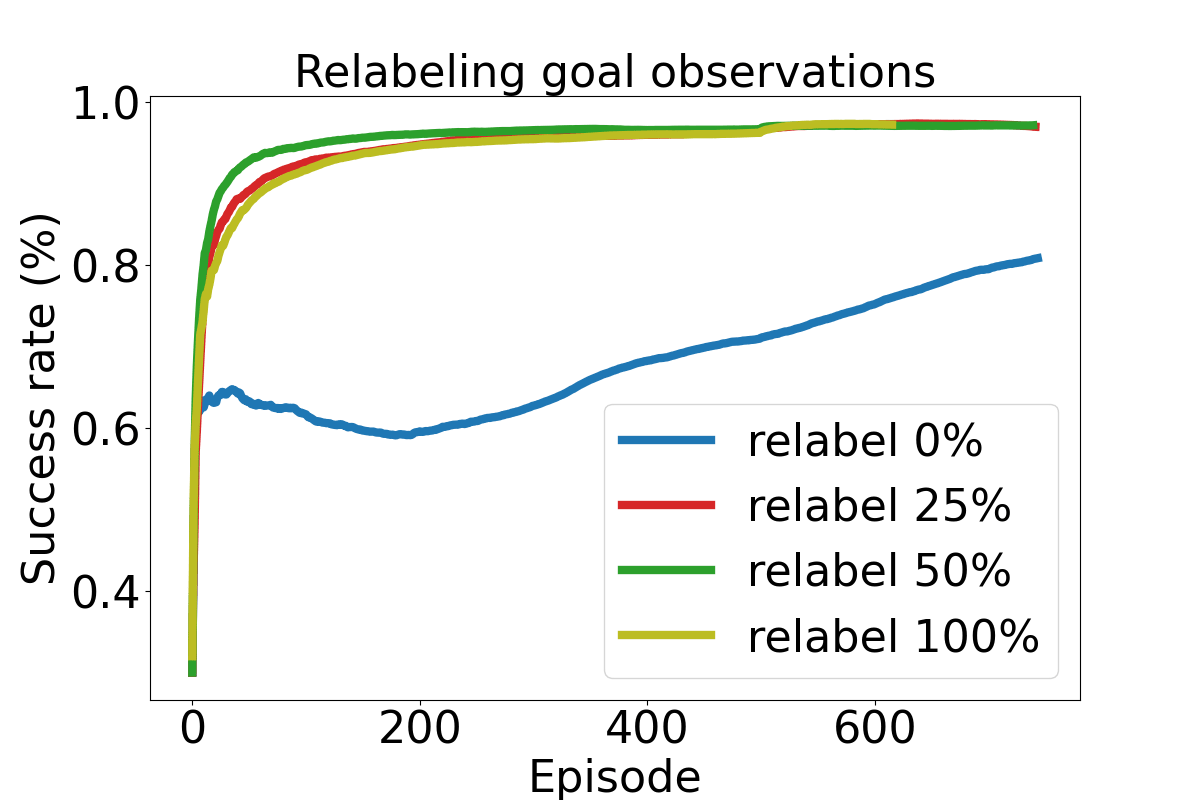}
    \caption{Results for different relabeling proportions. Each line shows the average success rate of 50 randomly sampled one-step navigation tasks during training.}
    \label{fig:relabel_proportion}
\end{figure}

\subsection{Architectures of the two variants}
\begin{figure}[ht]
    \centering
    \includegraphics[height=6cm, width=14cm]{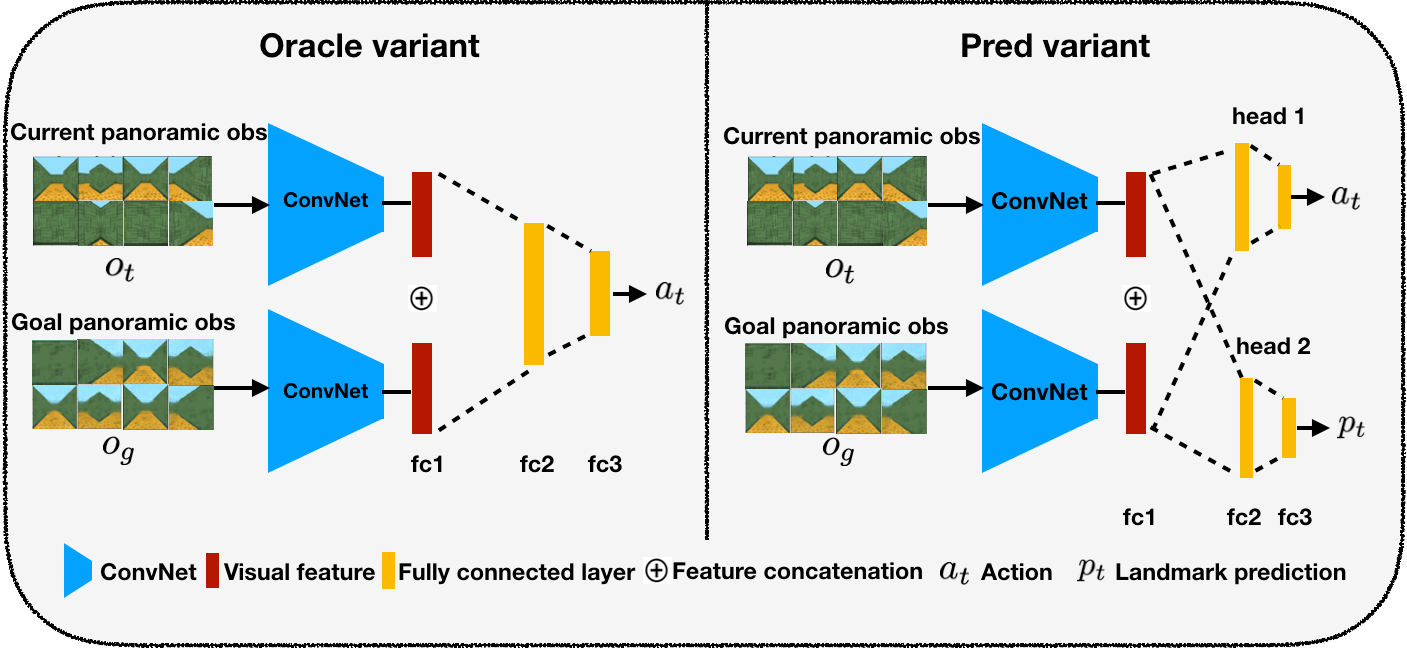}
    \caption{Local goal-conditioned controller architectures. 1) ``Oracle'' variant is the one receiving the landmark reaching signal from the environment. Therefore, it takes the panoramic observations $o_t$ and $o_g$. Then, it outputs an action $a_t$. 2) Besides outputting the action $a_t$, ``pred'' variant is the one that predicts the landmark reaching signal by outputting an extra $p_t$, which predicts the probability of the current $o_t$ being $o_g$}
    \label{fig:local_controller}
\end{figure}

In this section, Figure \ref{fig:local_controller} shows the architectures of the ``oracle'' and ``pred'' variants used in our experiments. Note, in both variants, we share the same visual encoder (i.e. ConvNet) to extract the visual feature of the current observation and the goal observation.

\section{Appendix E: Additional experiment results for using partially wrong 2-D map}
\begin{figure}[ht]
\begin{subfigure}{0.5\textwidth}
    \centering
    \includegraphics[scale=0.15]{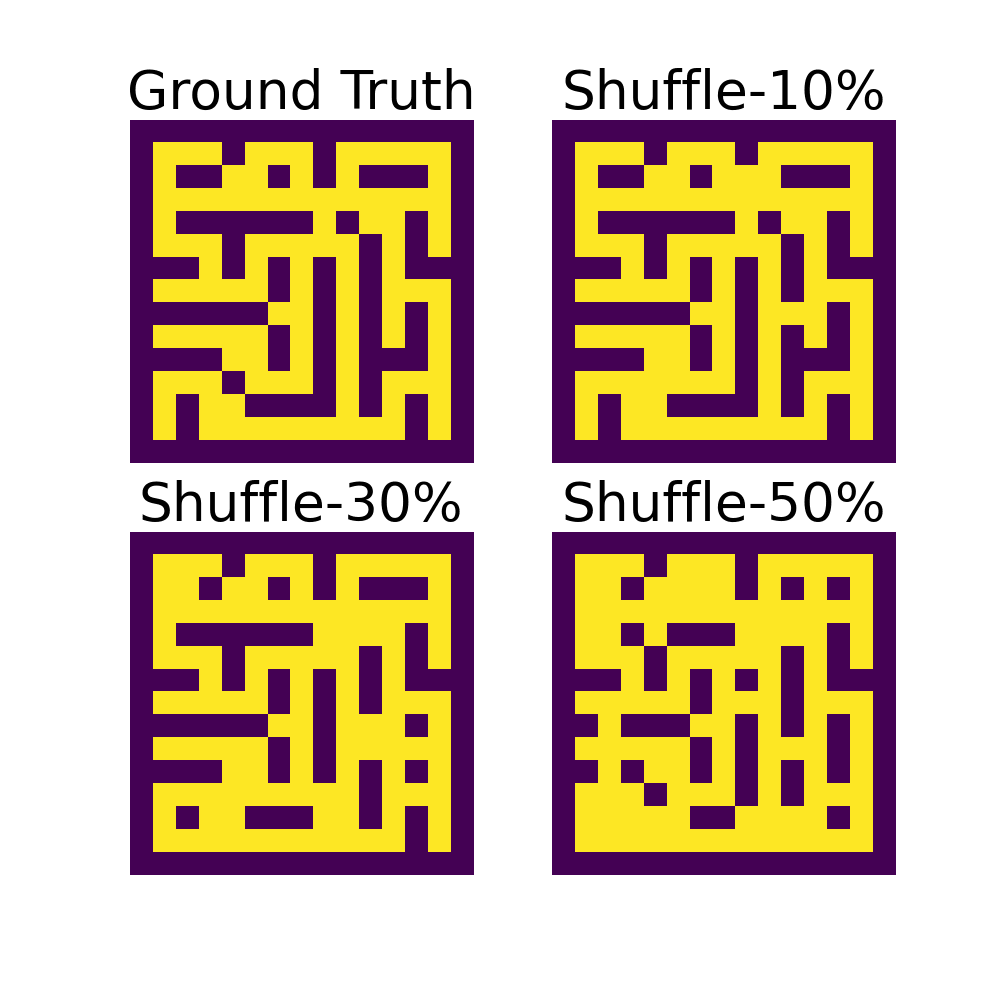}
    \caption{Flip wall to corridor}
    \label{fig:w2c_maps}
\end{subfigure}
\begin{subfigure}{0.5\textwidth}
    \centering
    \includegraphics[scale=0.18]{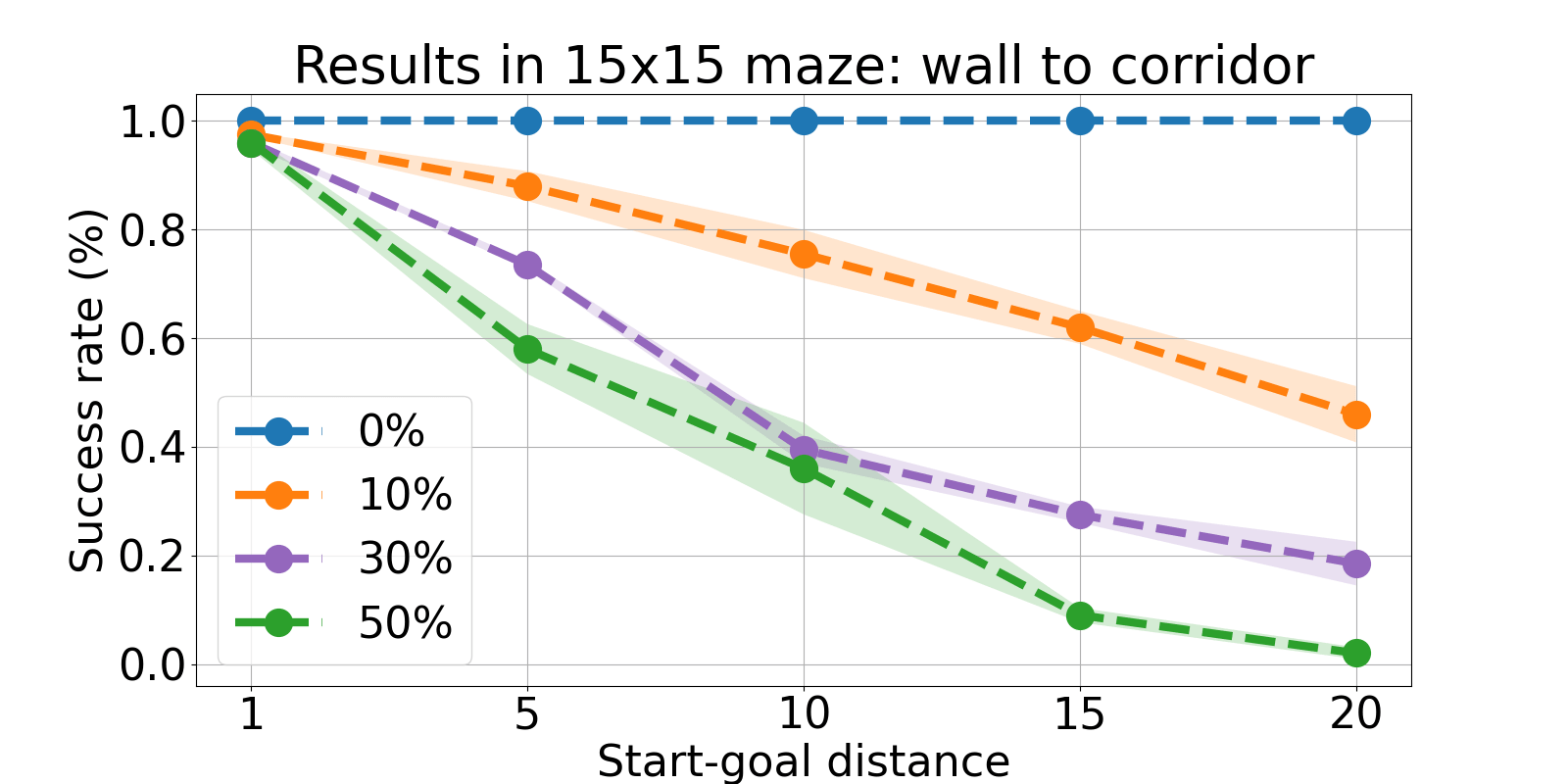}
    \caption{Flip wall to corridor}
    \label{fig:w2c_res}
\end{subfigure}
\begin{subfigure}{0.5\textwidth}
    \centering
    \includegraphics[scale=0.15]{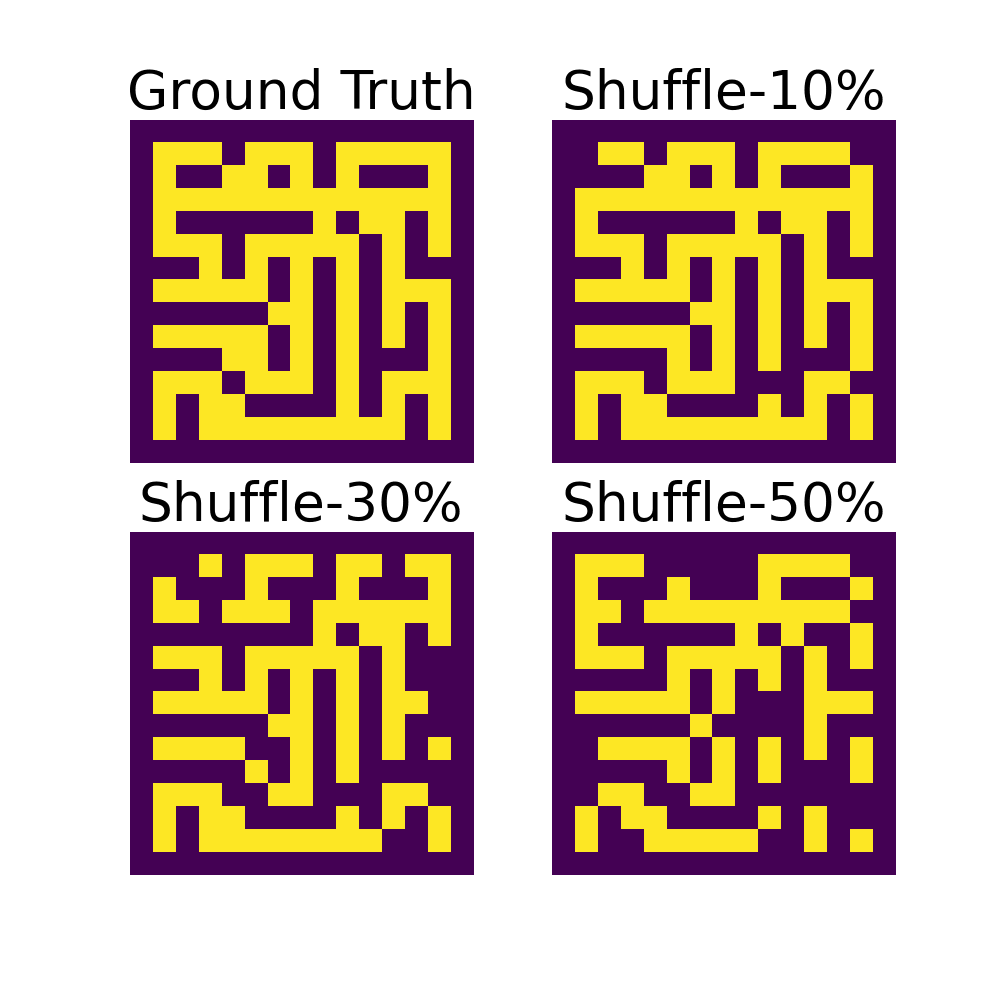}
    \caption{Flip corridor to wall}
    \label{fig:c2w_maps}
\end{subfigure}
\begin{subfigure}{0.5\textwidth}
    \centering
    \includegraphics[scale=0.18]{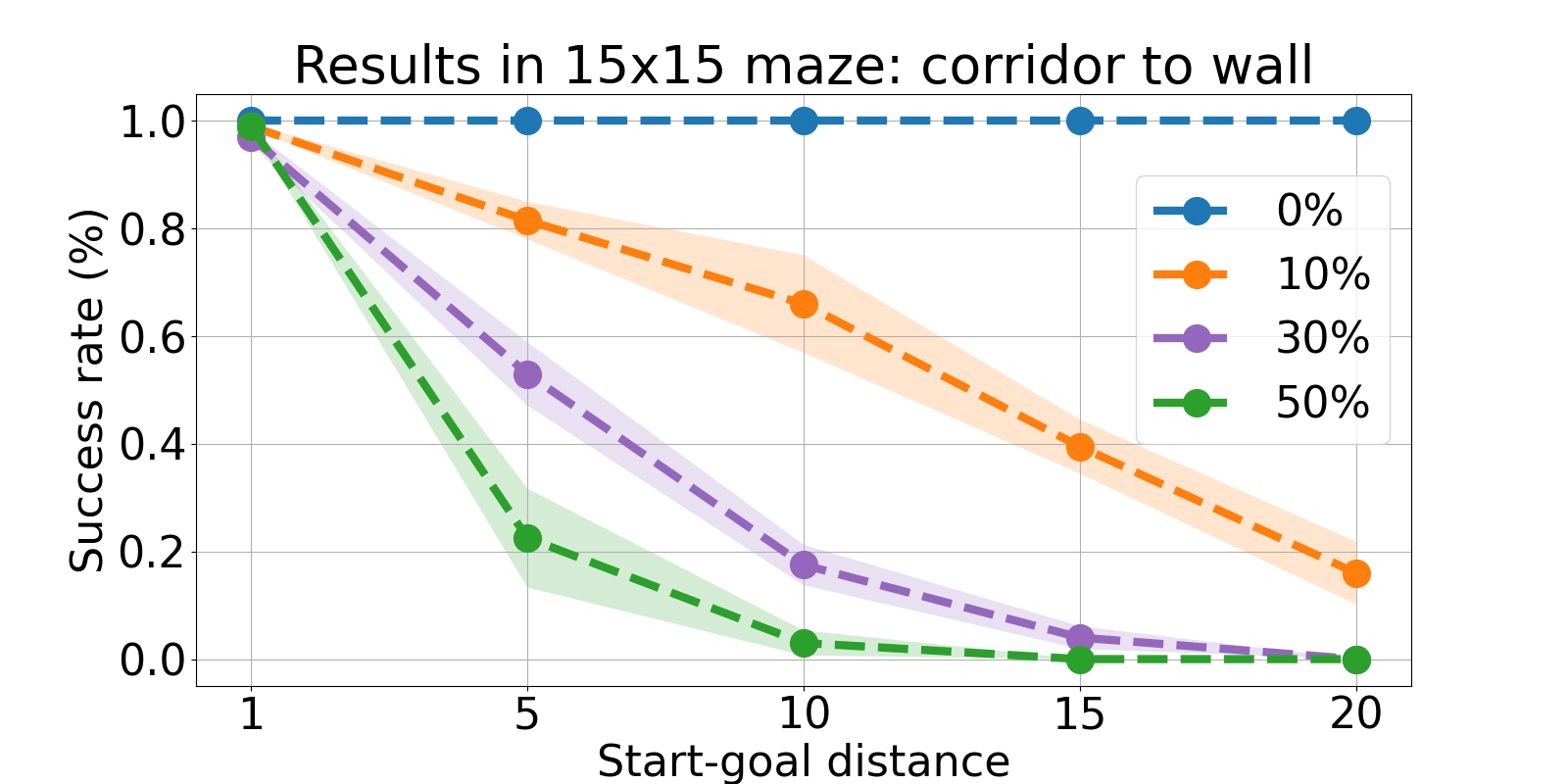}
    \caption{Flip corridor to wall}
    \label{fig:c2w_res}
\end{subfigure}
\begin{subfigure}{0.5\textwidth}
    \centering
    \includegraphics[scale=0.15]{images/mix.png}
    \caption{Flip randomly}
    \label{fig:mix_maps}
\end{subfigure}
\begin{subfigure}{0.5\textwidth}
    \centering
    \includegraphics[scale=0.18]{images/Results_in_15_15_maze_mixed.png}
    \caption{Flip randomly}
    \label{fig:mix_res}
\end{subfigure}
    \caption{Additional results for using partially wrong 2-D maps.}
    \label{fig:append_imprecise_map}
\end{figure}

In this section, we provide additional results for section \ref{subsec:imprecise_map}. Here, we discuss one challenging type of imprecise maps that are partially wrong 2-D maps. In order words, although the rough 2-D maps used in the previous experiments (section \ref{subsec:multi_goal_seen} and section \ref{subsec:generalize_unseen}) can not be directly used to guide navigation, they generally capture the correct 2-D layout. However, in this section, we manually construct some partially wrong 2-D maps by randomly flipping a proportion of wall and corridor positions. (See Figure \ref{fig:w2c_maps}, \ref{fig:c2w_maps} and \ref{fig:mix_maps}). 

To demonstrate the performance, we design three flipping scenarios to construct the partially wrong 2-D maps. In scenario 1, we will randomly select a proportion of wall positions and convert them to be ``fake'' corridors. In scenario 2, we randomly select a proportion of corridor positions and convert them to be ``fake'' wall positions. Finally, in scenario 3, we randomly select a proportion of both wall and corridor positions and convert them to either corridor or wall positions.

Overall, in the three scenarios, the performance decreases when the distance increases. However, with slightly wrong 2-D maps ($\leq 10\%$), our method still preserves an average $75\%$ navigation success rate for distance $\leq 15$ in the three scenarios. This is mainly because of the proposed dynamic topological map that plans another feasible path when the 2-D map is partially wrong. As pointed out in section \ref{subsec:imprecise_map}, we do not expect our method to perform well when most of the positions on the rough 2-D map are wrong. (e.g. shuffling $30\%, 50\%$ in Figure \ref{fig:append_imprecise_map}) On the one hand, a significant change of the local patches can cause the generator to generate wrong landmark observations. On the other hand, rough 2-D maps containing too much error are probably useless because no feasible path can be found on the map. 
\clearpage
\end{document}